\documentclass[]{gtech}
\usepackage{amssymb}
\usepackage{multirow}
\usepackage{bigdelim}
\usepackage{todonotes}
\usepackage{longtable}
\usepackage{tabularray}
\usepackage{wrapfig}
\usepackage[most]{tcolorbox} 
\usepackage{xcolor}
\usepackage{url}
\usepackage{hyperref} 
\usepackage{float}
\usepackage[table,xcdraw]{xcolor}
\usepackage[normalem]{ulem}
\usepackage{booktabs}
\usepackage{graphicx}
\usepackage{colortbl}

\useunder{\uline}{\ul}{}

\renewcommand{\title}[1]{\newcommand{\titlelist}{{\huge\fontfamily{optimistic}\selectfont #1}}}

\definecolor{prompt}{HTML}{5f84e4}
\definecolor{img}{HTML}{820100}

\usepackage{arydshln}

\definecolor{CQColor}{rgb}{0.0,0.0,1.0} 

\definecolor{TSColor}{rgb}{0.5,0.0,0.8} 

\definecolor{CQRColor}{rgb}{1.0,0.0,1.0} 

\usepackage{colortbl}
\usepackage{amssymb}
\usepackage{pifont}
\usepackage{booktabs,multirow}
\usepackage{makecell}
\usepackage{tabulary}
\usepackage{fontawesome5}
\usepackage{bbding}
\usepackage{multicol}

\newlength\savewidth

\title{
\centering
\textbf{\textcolor[HTML]{0369ff}{Venus-DefakerOne2}: A Vision‑Forensics Foundation Model with Broader Coverage}
}

\author{
\textcolor[HTML]{0369ff}{\textbf{GuangJian Team,~Ant Group\\ Technical Committee on Digital Media Forensics and Security, CSIG\\}}
\vspace{-0.8cm}
}
\vspace{-0.8cm}

\abstract{
Real-world forgery is inherently heterogeneous: it manifests not only as digital manipulation but also as physical spoofing, and is carried not only by images but also by videos. Despite substantial progress, a unified anti-forgery foundation model covering such a broad threat spectrum remains absent. In this paper, we present Venus-DefakerOne2, which substantially extends the forensic capability boundary along both the forgery-form and media-type dimensions. Trained on tens of millions of open-source and in-house samples, it incorporates two key designs. Tide-Style-Tuning alternates LoRA and full-parameter fine-tuning to balance computational efficiency, generalization, and hard-sample fitting while mitigating catastrophic forgetting. Entropy-based Test Adaptation determines the optimal prediction modality by analyzing prediction entropy. Venus-DefakerOne2 achieves new state-of-the-art performance across a comprehensive suite of benchmarks, including DeepFake faces (96.7\%), AI-generated images (93.3\%), tampered documents (87.6\%), manipulated natural images (89.6\%), physical spoofs (93.1\%), and AI-generated videos (96.0\%). It significantly outperforms both specialized expert models and its predecessor, DefakerOne, establishing itself as a powerful and versatile foundation model for binary-decision forensic tasks.
\vspace{-0.1cm}
}

\vspace{-0.4cm}

\gtechdata[Project Page]{%
    \textcolor{blue}{\faGlobe} \ \href{https://venus-guangjian.github.io/Venus-DeFakerOne2}{\url{https://venus-guangjian.github.io/Venus-DeFakerOne2}}%
}
\vspace{-0.4cm}
\gtechdata[Code, Model, Data]{%
    \faGithub \ \href{https://github.com/inclusionAI/DefakerOne2}{\url{https://github.com/inclusionAI/DefakerOne2}}%
}
\vspace{-0.2cm}

\begin{document}
\maketitle

\begin{figure*}[h]
\vspace{-0.1cm}
    \centering
    \includegraphics[width=0.85\textwidth]{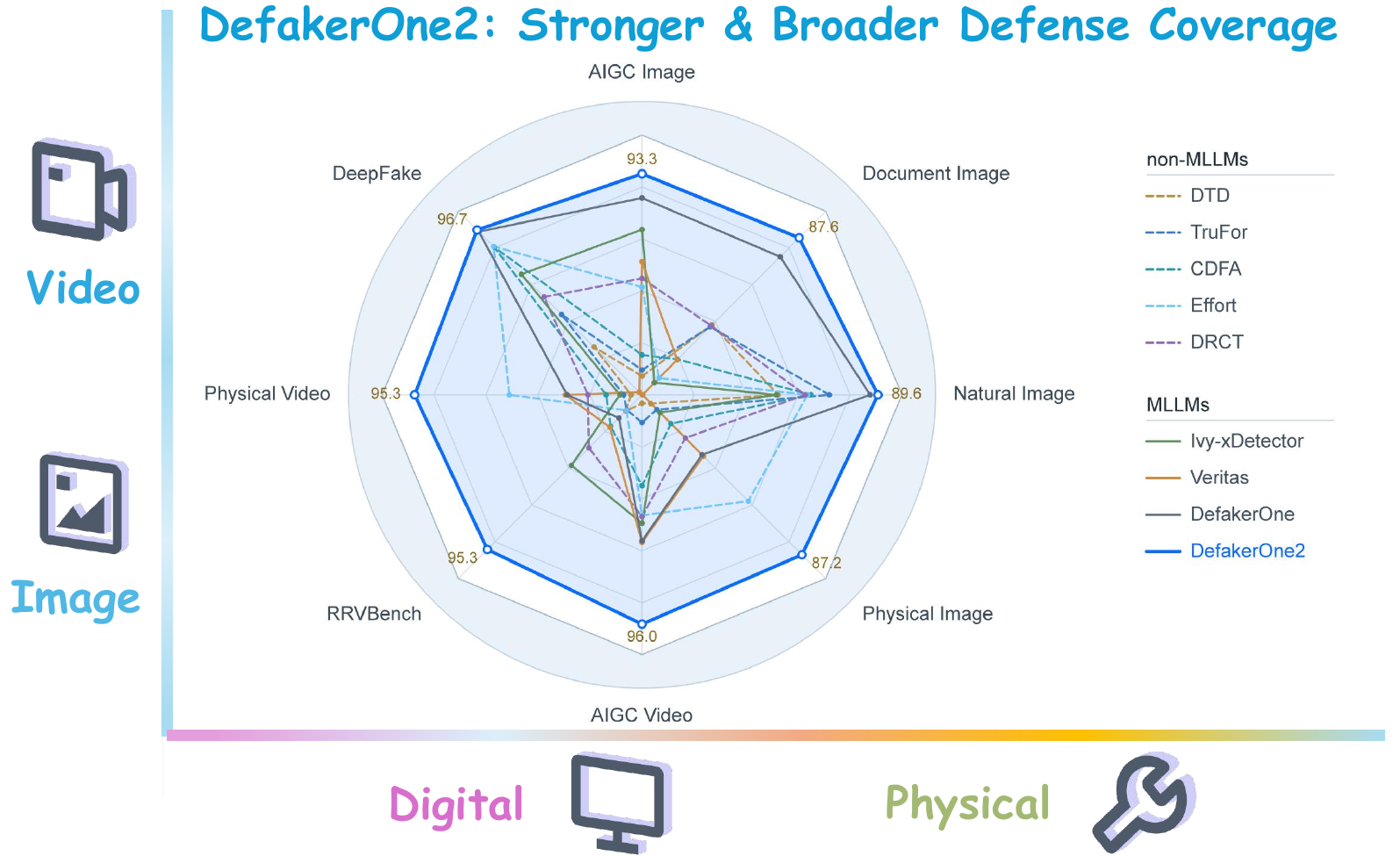}
    \caption{
    DefakerOne2 significantly expands defense coverage along both the forgery-form and media-type dimensions.
    }
    \label{fig:teaser}
\end{figure*}

\vspace{-0.1cm}
\begin{figure*}[h]
    \centering
    \includegraphics[width=1.0\textwidth]{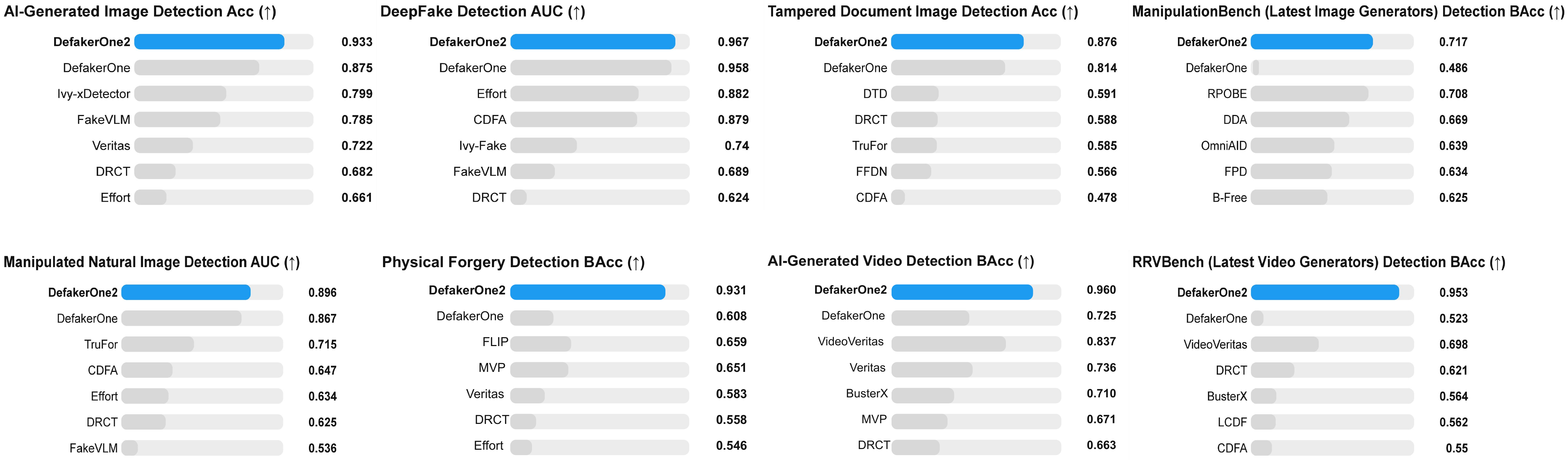}
    \caption{
    DefakerOne2 consistently outperforms its predecessor DeFakerOne and existing state-of-the-art detectors across all image and video forensic domains.
    }
    \label{fig:teaser}
    \vspace{+0.3cm}
\end{figure*}

\tableofcontents

\newpage
\section{Introduction}
\label{sec:intro}

Visual media including images and videos has become the predominant carrier of information in the digital era~\cite{uperhead, Xin2026UniMoMoEM}, underpinning applications from news dissemination, social communication to identity authentication and judicial evidence~\cite{jiang2026think, jiang2025tabdsr}. Meanwhile, the proliferation of sophisticated acquisition devices, accessible editing tools, and generative AI models has dramatically lowered the barrier to fabricating and tampering with visual content. Forged media can bypass biometric verification~\cite{fas1, yu2024unified}, deceive content moderation systems~\cite{ntire_aigc, qu2025webly}, and manipulate public opinion at scale~\cite{ma2023iml, ntire_deepfake}, posing substantial risks to information security and social trust. Consequently, detecting visual media forgery has emerged as a fundamental problem that demands scalable and generalizable solutions.

The landscape of real-world forgery, however, is far more heterogeneous than what most existing solutions assume (Ji-Zhe phenomenon~\cite{du2026buildmonolithicmodelfake}). Along the forgery-form dimension, forgery is not limited to digital manipulation performed in the pixel or latent space (e.g., splicing, inpainting, and AI-generated imagery)~\cite{omniiml, wong2025adcd}; it also takes physical forms  such as mask presentation attacks and document recapture attacks~\cite{juan_pad1, dlc2021}. Along the media-type dimension, forged content is carried not only by still images but also by videos, whose artifacts may reside in spatial textures, temporal dynamics, or cross-frame context. 

A practical vision anti-forgery system must therefore cope with the full matrix of \{digital, physical\} × \{image, video\}.
Existing research, however, remains highly fragmented. The majority of detectors are designed for a single scenario, such as a specific manipulation type~\cite{park2023kid34k}, a fixed media format~\cite{hopf2025practical}, or a narrow acquisition setting~\cite{liu2022pscc}, and generalize poorly beyond their training distribution. Worse still, different forgery forms have traditionally been studied in largely separate communities: digital image forensics~\cite{guozhiqing}, face anti-spoofing~\cite{zhang2020casiasurf}, document fraud detection~\cite{zeng2026docshield}, and video forensics~\cite{chen2026demamba} each maintain their own benchmarks, assumptions, and model designs, even though the underlying goal, distinguishing genuine media from forged ones, is the same. This fragmentation prevents knowledge from being shared across scenarios and makes it costly to deploy and maintain a separate model for every new threat~\cite{qu2024omni}. Our preliminary work, Venus-DeFakerOne~\cite{defakerone}, took a step toward unification by consolidating multi-scenario digital image forgery detection into a single foundation model. Nevertheless, its capability boundary stops at digitally forged images: physical forgeries and video media lie entirely outside its scope. To the best of our knowledge, a unified foundation model that simultaneously covers digital and physical forgery across both images and videos of multiple content types is still lacking.

To build such a comprehensive vision-forensics foundation model, we train a Multi-modal Large Language Model (MLLM) on a large corpus of about 50M anti-forgery training data. Our training data are mainly sourced from three parts: public datasets, self-‌constructed datasets and manually-annotated online-attack samples, covering the full matrix of \{digital, physical\} × \{image, video\}. Despite the high diversity, training such a model faces two key challenges, training cost and uneven data difficulty. This large-scale training corpus entails prohibitive computational cost under naive full-parameter training. Moreover, samples vary drastically in learning difficulty: some exhibit conspicuous artifacts that are fitted within a few steps, while others contain only faint, elusive traces. A static training regime either wastes capacity on trivially easy samples or suffers catastrophic forgetting of the base model's general priors.

Additionally, there exists a modality dilemma in video forensics. Forgery artifacts in videos distribute unevenly across dimensions. For some forgeries, evidence concentrates in temporal information and cross-frame context---flickering, motion inconsistency, and disrupted inter-frame continuity---making the native video modality the natural choice. For others, evidence concentrates in the textural details of individual frames; feeding such a clip as a video brings little temporal gain and instead dilutes the model's attention on spatial artifacts with redundant temporal tokens. Since which modality carries the stronger evidence is sample-dependent and unknown a priori, a one-size-fits-all input modality is inherently suboptimal.

To address the training challenge, we propose \textbf{Tide-Style Tuning}, which reconciles computational efficiency, generalization, and fitting capacity. The training proceeds in three alternating phases, ebbing and flowing like a tide. It begins with Adaptation Ebb, where we adopt the general-purpose pretrained Qwen3.5 model through LoRA-based fine-tuning. This step operates as a warm-up, leveraging the strong priors of the Qwen3.5 backbone to promote generalization while mitigating catastrophic forgetting at low cost. The LoRA weights are then merged into the backbone, and the model undergoes full-parameter fine-tuning, unleashing the entire parameter budget to fit the large-scale diverse forgery corpus. Finally, training reverts to LoRA, which acts as a regularizer that steers the model toward hard samples and prevents overfitting to the abundant easy ones.

To solve the modality dilemma, we propose 	\textbf{Modality Synergism}, which aligns the two modalities on both the optimization side and the inference side. On the optimization side, our joint modality training presents each video to the model in both modalities simultaneously---the native clip and the frame sequence extracted from it with an adaptive frame-rate strategy---and optimizes an independent detection loss for each modality, forcing the shared backbone to internalize both spatial and temporal forensic evidence. On the inference side, our entropy-based test adaptation obtains predictions from both modalities, measures the certainty of each prediction by its Shannon entropy, and adopts the prediction of the more certain modality as the final decision. In this way, the model adaptively routes each sample, without any additional training or learned routing module, to the modality whose evidence is most salient for that sample.

Through our proposed methods, our model, named \textbf{Venus-DefakerOne2}, effectively learns ‌shared fundamental features of vision forgery across all common contents, forgery forms and media types. 

Fake videos in public video forensics dataset are generated by old generators released more than one year ago. To keep up the rapid development of AIGC video generation techniques, we constructed a new video forensics benchmark, which termed as  \textbf{Realistic Rumor Video Benchmark (RRVBench)}. The fake videos in our RRVBench were generated by 7 latest closed-source video generators such as SeeDance 2.5 as well as 4 latest open-source video generators such as Minimax-H3. The video contents in our RRVBench focuses real-world rumor scenarios and minimizes the application gap.

Our DefakerOne2 is evaluated on 53 visual media forgery detection benchmarks and 14 forgery localization benchmarks. With a single set of 2B parameters, it ranks first on every one of the eight domains, achieving SOTA on 40/53 and 11/14 benchmarks, raising the average from 80.4 to 90.9 on images and from 57.6 to 95.7 on videos over its predecessor DefakerOne. The progress is qualitative: physical forgery, entirely outside the scope of DeFakerOne, climbs from 50.7 to 87.2, and on RRVBench DefakerOne2 reaches 95.3 while even 7B-parameter video-forensics MLLMs degenerate to near-chance guessing (best 69.8). Crucially, it is never outperformed by a domain specialist on that specialist's own domain, and its narrow cross-domain spread shows no domain pays for this generality. DefakerOne2 thus collapses a fragmented assembly of scenario-specific detectors into one deployable foundation model.

\begin{figure*}[h]
    \centering
    \includegraphics[width=1.0\textwidth]{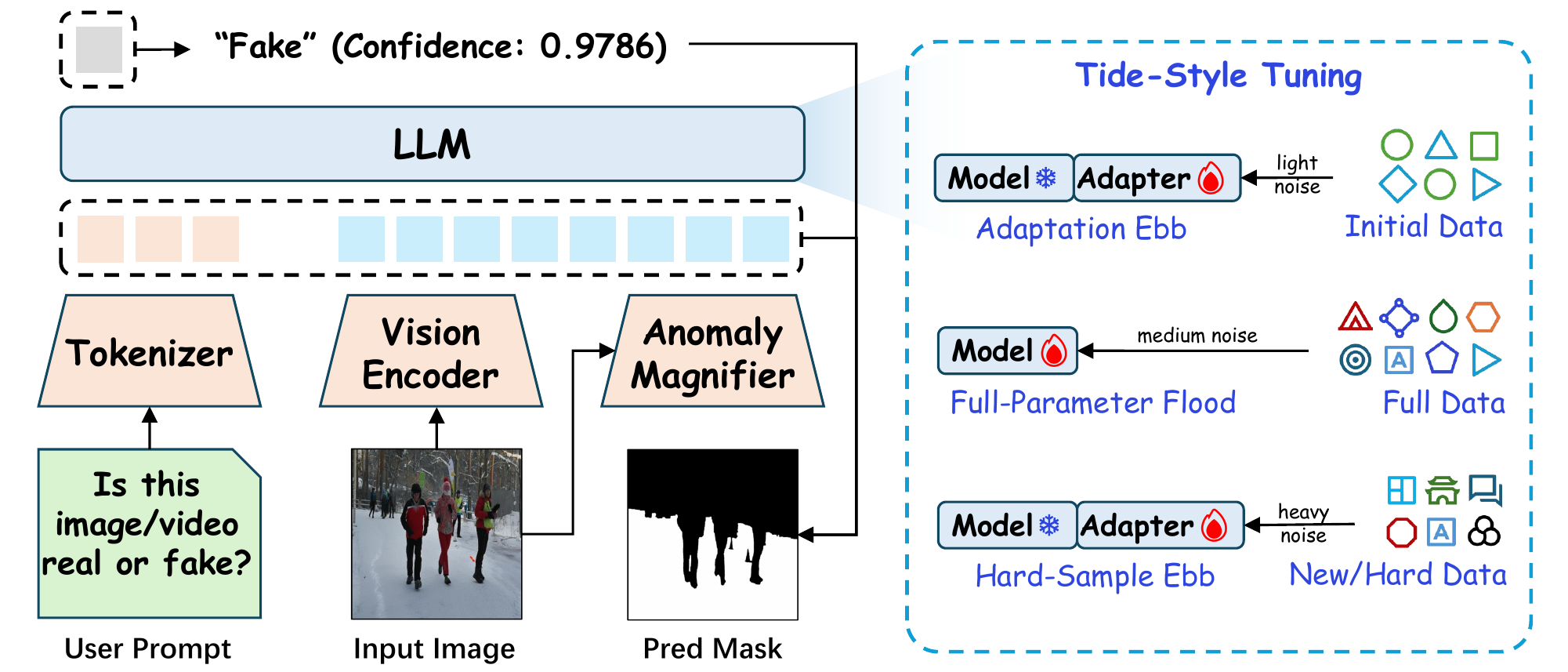}
    \caption{The overall framework of our DefakerOne2.
    }
    \label{fig:model}
\vspace{-0.1cm}
\end{figure*}

\section{Method}
\vspace{-0.1cm}
\subsection{Model Architecture}
\noindent \textbf{Overall Framework.} Figure~\ref{fig:model} illustrates the overall framework of DefakerOne2. Built upon a multimodal large language model (MLLM), DefakerOne2 further introduces an Anomaly Magnifier module as an auxiliary encoder to strengthen the perception of fine-grained forensic cues, enabling a single unified model to produce both image/video-level detection predictions and pixel-level localization masks. Specifically, the MLLM first perceives the input and performs global-level detection: guided by a fixed prompt template tailored for efficient forensics, it analyzes the input, renders a binary authenticity judgment, and generates a corresponding confidence score for the downstream fine-grained segmentation task. The segmentation module then exploits this global forensic prior to conduct pixel-level analysis, yielding a mask that highlights the tampered regions.

\noindent \textbf{Forgery Detection.} DefakerOne2 adopts Qwen3.5-2B as its MLLM backbone. Since Qwen3.5-2B is a native multimodal model, it can better leverage its world knowledge to implicitly reason about potential semantic anomalies, providing complementary high-level evidence for forgery analysis. In addition, the strong video understanding capability of Qwen3.5 makes it well suited to our unified image/video formulation. For detection, we query the MLLM with a fixed and direct template for image/video-level binary classification. Our template query, ``Is this image/video real or fake? Answer only Real or Fake.'', guides the model to commit to its final conclusion in a single output token, which maximizes both inference efficiency and the stability of the obtaining reliable prediction confidence.

\noindent \textbf{Forgery Localization.} MLLMs are adept at global-level reasoning but often fall short in pixel-level analysis and prediction. To overcome this limitation, we introduce the Anomaly Magnifier (AM) module as an auxiliary encoder. This encoder adopts the DINOMAC~\cite{dinomac} backbone, which is a DINOv3-Large~\cite{dinov3} architecture fine-tuned with LoRA~\cite{hu2022lora} adapter and forensic training strategies. Features from the MLLM vision encoder and the AM module are fed into a mask decoder consisting of several convolutional layers for the final mask prediction. Inspired by APSC-Net~\cite{miml}, we further introduce a calibration loop for better localization performance. In this loop, the localization prediction is binarized with a dynamic threshold and integrated with the LLM's global judgment, so that the pixel-level mask is explicitly calibrated by the image/video-level decision. 

\noindent \textbf{Calibration Loop.} A pixel-level decoder that is optimized only
with region-wise supervision treats every input identically, and therefore tends
to over-segment authentic content, where any salient but benign region (texture
boundaries, strong edges, compression artifacts) can be mistaken for tampering.
Our calibration loop removes this granularity gap by making the localization
decision boundary \emph{conditioned on} the global judgment, so that the two
levels of evidence are coupled instead of being produced independently.
Concretely, let $p\in[0,1]$ be the image/video-level forgery probability produced
by the MLLM, and let $\mathbf{M}\in[0,1]^{H\times W}$ be the anomaly map output by
the mask decoder. Rather than binarizing $\mathbf{M}$ with a fixed cut-off, we
derive the cut-off from $p$:
\begin{equation}
    \tau(p) \;=\; 0.75 - \frac{p}{2}
    \;=\; \frac{1}{2} + \frac{1}{2}\left(\frac{1}{2} - p\right),
    \label{eq:dynamic-threshold}
\end{equation}
and the calibrated localization mask is obtained as
\begin{equation}
    \mathbf{M}_{\mathrm{cal}} \;=\; \mathbb{1}\!\left[\,\mathbf{M} > \tau(p)\,\right],
    \label{eq:binarize}
\end{equation}
where $\mathbb{1}[\cdot]$ is the indicator function.

\noindent\emph{Why it works.}
The second form of Eq.~\eqref{eq:dynamic-threshold} shows that $\tau(p)$ is the
neutral boundary $\tfrac{1}{2}$ shifted by \emph{half of the discrepancy between
the global prior and the neutral point}. The shift is therefore signed by the
global evidence and monotonic in $p$: when the MLLM is confident that the content
is forged ($p\!\rightarrow\!1$), the boundary is lowered to $\tau\!=\!0.25$, so the
decoder is granted a larger admissible region and weak but genuine tampering cues
are no longer suppressed --- the global branch effectively \emph{transfers its
confidence} to the pixel branch and raises recall on subtle manipulations.
Conversely, when the MLLM judges the content to be authentic ($p\!\rightarrow\!0$),
the boundary is raised to $\tau\!=\!0.75$, and only highly salient anomalies
survive the cut-off, enforcing a conservative and cautious segmentation policy
that directly attacks the dominant error mode of false positives on benign
images. At $p\!=\!0.5$, where the global branch carries no information,
$\tau\!=\!0.5$ and the loop degenerates to the conventional fixed-threshold
binarization, i.e., the calibration never injects a bias that is not supported by
global evidence.

The behaviour is best characterized in the log-odds space. Writing
$\ell(x)=\log\frac{x}{1-x}$, Eq.~\eqref{eq:binarize} is equivalent to
\begin{equation}
    \ell\!\left(\mathbf{M}\right) \;>\; b(p),
    \qquad
    b(p) \;=\; \ell\!\left(\tau(p)\right)
    \;=\; \log\frac{0.75-\frac{p}{2}}{0.25+\frac{p}{2}},
    \label{eq:logit-boundary}
\end{equation}
so the calibration loop acts as an \emph{additive bias on the pixel-level
log-odds decision boundary}, whose magnitude is driven by the global log-odds.
Two properties make this coupling well-behaved. First, $b(p)$ is antisymmetric
about the neutral point, $b(p)=-b(1-p)$, hence the amount of encouragement given
to forged content is exactly the amount of suppression applied to authentic
content --- the prior is injected in a balanced way rather than as a one-sided
relaxation. Second, $b(p)$ is bounded within $[-\log 3,\,+\log 3]\!\approx\![-1.099,
+1.099]$, so even a saturated global prediction can only shift the boundary by a
limited amount. The mask decoder thus retains the capacity to disagree with the
global branch when the pixel-level evidence is strong enough, and the calibration
functions as a soft, evidence-driven prior rather than a hard override that would
propagate global errors into the localization output.

\noindent \textbf{Training Objective.} Both the detection and localization training are optimized with cross-entropy loss. Let $\mathcal{L}_{det}$ denote the detection loss and $\mathcal{L}_{seg}$ denote the segmentation loss, $I$ denote the input image, $M$ denote the prediction localization mask and $\hat{M}$ denote the ground-truth localization mask, the training losses can be formulated as :

\begin{equation}
    \mathcal{L}_{det} = - \log P(x_i | \mathcal{X}_{<i}, I)
\end{equation}

\begin{equation}
    \mathcal{L}_{seg} = \mathcal{L}_{BCE}(M, \hat{M})
\end{equation}

\subsection{Tide-Style Tuning}

\noindent \textbf{Training challenges.}
Fine-tuning the model directly on our 50M corpus is highly non-trivial, as the data exhibit extreme diversity and a long-tailed difficulty distribution. On the one hand, learning difficulty varies drastically across samples: some forgeries leave conspicuous artifacts that are fitted within a few optimization steps, whereas others expose only faint, elusive traces that cannot be mastered within a single training pass. On the other hand, data abundance varies drastically across subsets: some subsets share strong common patterns and appear in large volumes, while others are highly distinctive yet scarce. These disparities render any static training regime inadequate. A single epoch is sufficient for the easy portion of the data but leaves the hard portion under-trained; simply scaling up the training budget not only incurs prohibitive cost but also increases the risk of overfitting to the easy majority. The choice of fine-tuning parameterization further sharpens the dilemma. Under full-parameter fine-tuning, the easy samples encountered at the beginning of training drive the model to rapidly overfit to dataset-specific cues and catastrophically forget its pretrained knowledge, while hard samples yield large gradients that erode the same knowledge and degrade generalization. Under LoRA fine-tuning, by contrast, the restricted parameter budget leaves the model unable to adequately fit tens of millions of diverse samples. This dilemma is compounded in practice: an anti-forgery foundation model must continually adapt to newly collected training data, and such continual adaptation inherits the same cost and overfitting risks.

\noindent \textbf{Overview.}
To mitigate this dilemma, we propose Tide-Style Tuning, which reconciles computational efficiency, generalization, and fitting capacity. As illustrated in the right panel of Figure~\ref{fig:model}, training proceeds in three alternating phases whose trainable parameter budget ebbs and flows like a tide: an \emph{ebb} phase freezes the backbone and updates only a lightweight LoRA adapter (a ``narrow'' network), whereas a \emph{flood} phase unfreezes the entire model (a ``wide'' network). In parallel, we perturb the training input with DINOMAC adversarial noise~\cite{dinomac} whose intensity is progressively increased from light to medium and finally to relatively heavy, discouraging the memorization of trivially easy samples and improving robustness.

\noindent \textbf{Adaptation Ebb.}
In the first phase, we attach a LoRA adapter to the officially pretrained Qwen3.5 backbone and fine-tune only the adapter on an initial portion (15M) of the data under light noise. This phase acts as a low-cost warm-up that rapidly adapts the general-purpose pretrained model to the forgery detection task. Meanwhile, freezing the backbone parameters effectively prevents catastrophic forgetting, so that the model keeps access to its acquired world knowledge and learns to exploit it for reasoning about the high-level semantic anomalies present in forged images and videos, yielding stronger generalization. Nevertheless, applying LoRA alone to a 2B-parameter model over the full 50M corpus would underfit, since the limited adapter capacity cannot absorb data of such scale and diversity.

\noindent \textbf{Full-Parameter Flood.}
We therefore introduce a second phase that releases the full parameter budget. The LoRA weights obtained in the first phase are first merged into the backbone, and the entire model is then fine-tuned on the full corpus under medium noise. With the greatest degree of freedom, this phase offers the strongest fitting capacity for the large-scale heterogeneous corpus. Importantly, because the first phase has already sufficiently adapted the model to forgery detection, training rarely produces large destructive gradients, and catastrophic forgetting is largely avoided. The strengthened DINOMAC adversarial noise further prevents the model from directly memorizing easy samples and enhances robustness.

\noindent \textbf{Hard-Sample Ebb.}
After the flood phase, the model has mastered the majority of the corpus, yet a small number of hard samples---together with newly collected ones---remain under-fitted. We therefore freeze the backbone once more and attach a fresh LoRA adapter, which is fine-tuned on the new and hard samples under heavy noise. This final ebb phase brings three benefits. First, adapter-only optimization drastically reduces the cost of continued training. Second, the restricted parameter budget acts as an implicit regularizer, preventing the model from overfitting to the easy samples that have already been sufficiently fitted. Third, it steers the model to adapt to the new samples on top of the general features learned by the backbone, rather than overfitting to idiosyncratic visual patterns; the heavy DINOMAC adversarial noise further suppresses overfitting to easy samples. The model obtained after this third phase is taken as our final model.

\subsection{Modality Synergism}
\label{sec: method_ms}

\noindent	\textbf{Modality dilemma.}
Spatial and temporal evidence both exist in video forgery, yet they do not always co-occur strongly within a single sample. From the model's perspective, the two input modalities correspond to two distinct feature-acquisition pathways. Encoding each frame as an independent image lets the vision encoder concentrate on the spatial texture of individual frames, preserving fine-grained intra-frame artifacts such as generation fingerprints and boundary discontinuities, but discards inter-frame dynamics. Encoding the clip natively as a video recovers the temporal context required by evidence such as flickering and motion inconsistency, but at the cost of dispersing attention over redundant temporal tokens and weakening the response to subtle per-frame traces. Since which pathway carries stronger evidence is sample-dependent and unknown a priori, any fixed modality can only be optimal in expectation. We therefore cast the modality choice as a per-sample inference-time decision, which in turn requires a reliable, training-free measure of how confident each modality's prediction is for the given sample. To this end, we propose \textbf{Modality Synergism}, which tackles the dilemma from both the optimization side and the inference side, via joint modality training and entropy-based test adaptation, respectively.

\begin{figure*}[t]
    \centering
    \setlength{\abovecaptionskip}{-0.1cm}
    \includegraphics[width=1.0\textwidth]{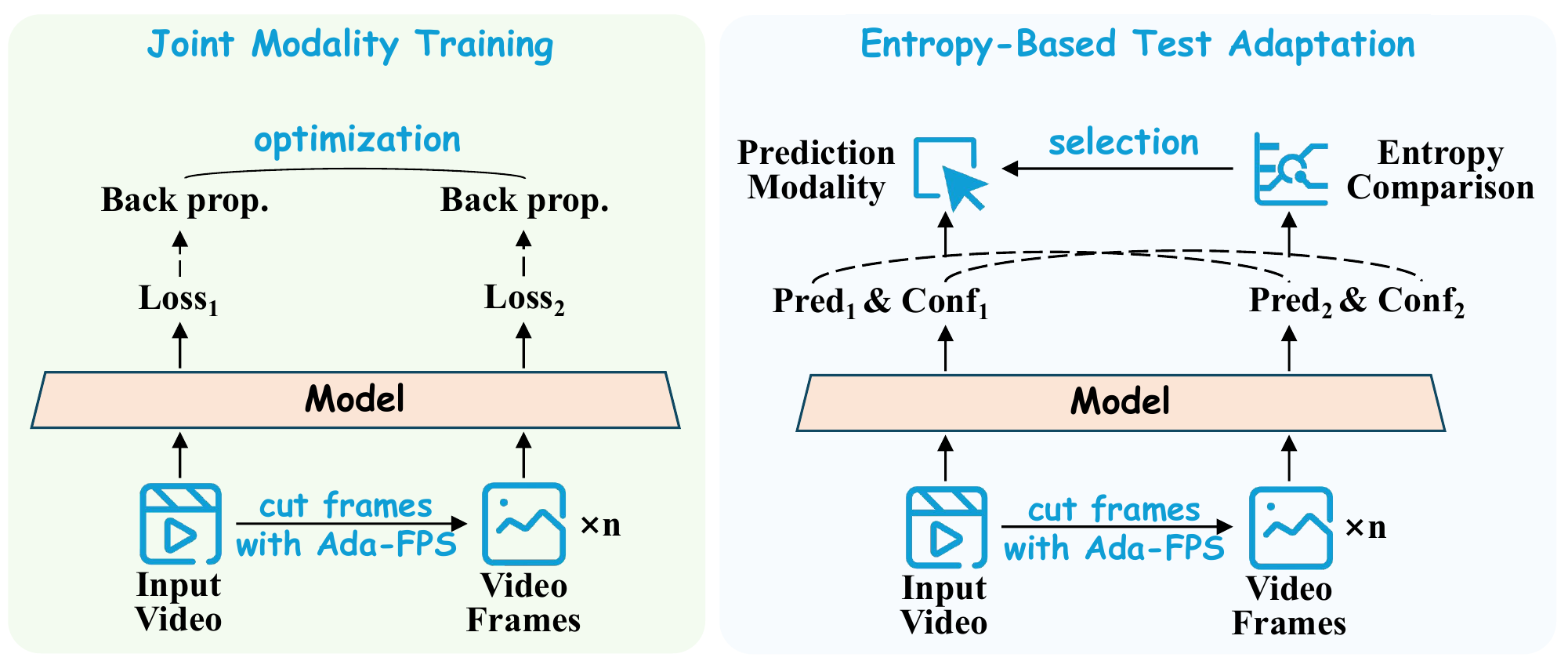}
    \caption{The proposed Entropy-Based Test Adaptation.
    }
    \vspace{-0.1cm}
    \label{fig:eta}
\end{figure*}

\noindent \textbf{Joint modality training.}
During training, each input video is presented to the model in both modalities simultaneously: the native clip, and the frame sequence extracted from it with Ada-FPS (Figure~\ref{fig:model}, left). A detection loss is computed independently for each modality against the same ground-truth label $y$, and both losses are back-propagated to optimize the shared network:
\begin{equation}
\mathcal{L} = \mathcal{L}_1 + \mathcal{L}_2 = \ell\!\big(p^{(\mathrm{vid})}, y\big) + \ell\!\big(p^{(\mathrm{img})}, y\big),
\label{eq:jmt}
\end{equation}
where $p^{(\mathrm{vid})}$ and $p^{(\mathrm{img})}$ are the Real/Fake predictions obtained from the video modality and the frame-image modality, respectively, and $\ell$ is the cross-entropy loss on the single decision token. Optimizing the two pathways jointly forces the shared backbone to internalize both spatial and temporal forensic evidence, so that either pathway can be reliably invoked at test time. (For still-image training samples, only the image-modality loss is applied.)

Because the collected video sources exhibit a strongly long-tailed quantity distribution, sampling every source at a fixed rate would allow a few abundant sources to dominate the overall frame budget. We therefore propose an adaptive frame-rate strategy (Ada-FPS) that equalizes the number of training frames across sources. We set a default rate $f_0 = 1$ FPS and pre-compute, for each source $\mathcal{S}$, the total number of frames that would be extracted at this rate, $F_{\mathcal{S}}(f_0) = \sum_{v \in \mathcal{S}} \lfloor f_0\, d_v \rfloor$, where $d_v$ is the duration of video $v$. If $F_{\mathcal{S}}(f_0)$ exceeds an upper threshold $N_{\max}$, the rate of this source is lowered so that its frame count is capped at $N_{\max}$; if it falls below a lower threshold $N_{\min}$, the rate is raised so that the frame count approaches $N_{\max}$, thereby amplifying the contribution of scarce sources. To avoid redundant sampling of nearly identical consecutive frames, the rate is further capped at $f_{\max} = 10$ FPS. Formally,
\begin{equation}
f_{\mathcal{S}} = \min\!\left( f_{\max},\;
\begin{cases}
f_0, & N_{\min} \le F_{\mathcal{S}}(f_0) \le N_{\max}, \\[4pt]
f_0 \cdot \dfrac{N_{\max}}{F_{\mathcal{S}}(f_0)}, & \text{otherwise},
\end{cases}
\right)
\label{eq:adafps}
\end{equation}
and each video $v \in \mathcal{S}$ is sampled into $\lfloor f_{\mathcal{S}}\, d_v \rfloor$ frames. This per-source rate adaptation effectively mitigates the quantity imbalance across video sources.

\noindent \textbf{Entropy-based test adaptation.}
At inference (Figure~\ref{fig:eta}, right), for each input video we discard the first and last 10\% of its duration to avoid ambiguous lead-in/lead-out content, and sample the remaining middle 80\% into frames with Ada-FPS. The model then evaluates the same video through both modalities, yielding two predictions with associated confidences, $(\hat{y}_1, c_1)$ from the video modality and $(\hat{y}_2, c_2)$ from the frame-image modality. We measure the certainty of each prediction by the Shannon entropy of the induced binary Real/Fake distribution,
\begin{equation}
H_m = -\,c_m \log c_m - (1 - c_m) \log (1 - c_m), \qquad m \in \{1, 2\},
\label{eq:entropy}
\end{equation}
and adopt the prediction of the more certain modality as the final decision:
\begin{equation}
\hat{y} = \hat{y}_{m^\star}, \qquad m^\star = \arg\min_{m \in \{1,2\}} H_m.
\label{eq:eta}
\end{equation}
Since lower entropy corresponds to a more decisive prediction, Eq.~\ref{eq:eta} routes each sample, at test time and without any additional training or learned routing module, to the modality whose evidence is most salient for that particular sample.

\section{Training Data Construction}

\begin{figure*}[t]
    \centering
    \setlength{\abovecaptionskip}{-0.3cm}
    \includegraphics[width=1.0\textwidth]{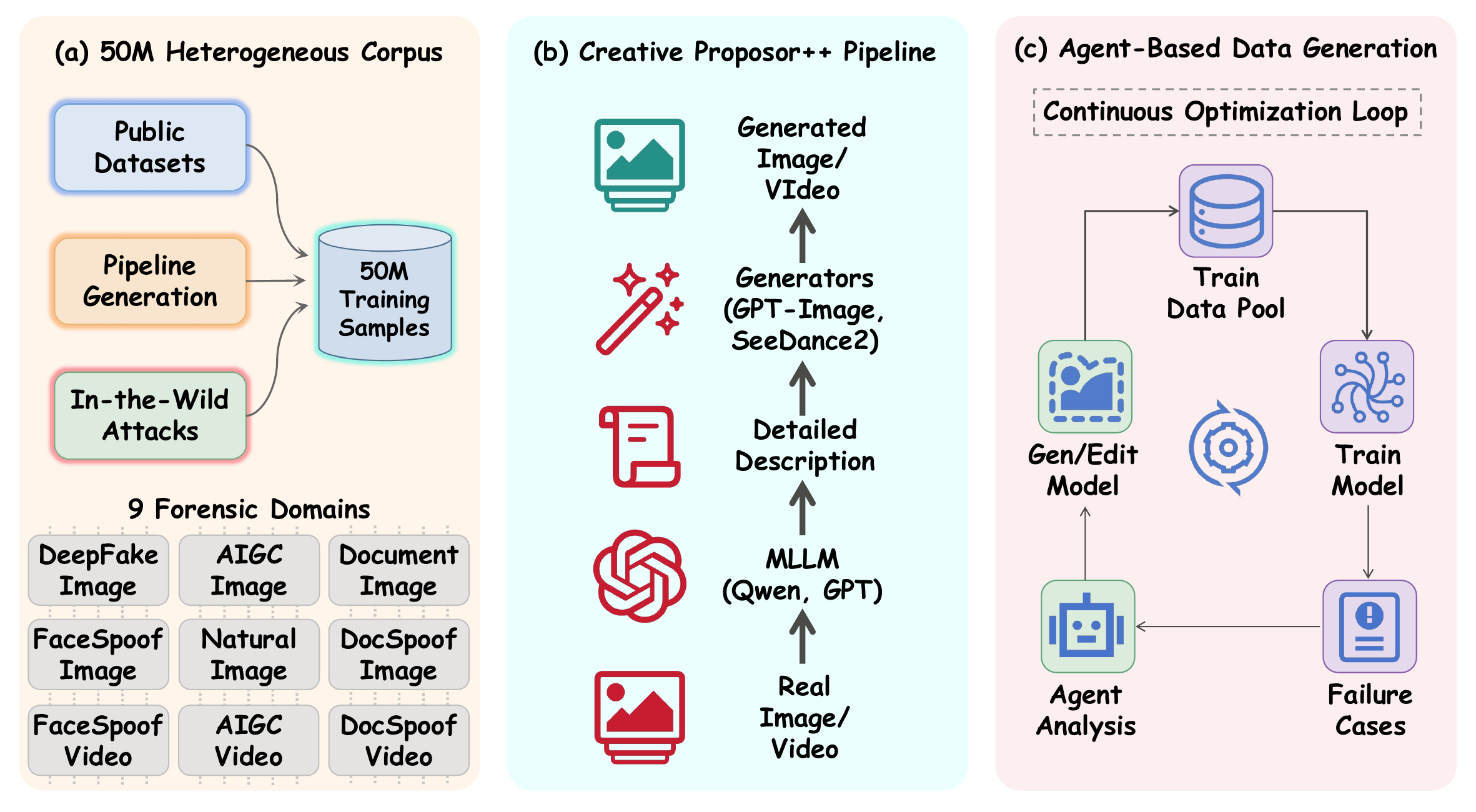}
    \caption{Our data pipeline.}
    \label{fig:data_pipeline}
\end{figure*}

\subsection{Overall}
We train our vision-forensics foundation model on a massive, heterogeneous corpus designed to span a broad spectrum of manipulation techniques, forgery granularities, and content modalities. As depicted in Figure~\ref{fig:data_pipeline}, this $50\text{M}$-sample training dataset is constructed by aggregating heterogeneous data sources and deploying dedicated, automated generation pipelines. Provenance-wise, our corpus integrates three primary tiers: public datasets, engine-synthesized data, and manually curated in-the-wild attack samples. Functionally, the dataset spans nine distinct forensic domains: AI-generated images, DeepFake faces, tampered documents, forged natural images, face anti-spoofing (images and videos), document anti-spoofing (images and videos), and AI-generated videos.

\subsection{Self-Synthesized Data}
Besides manually forged and annotated data, our private self-synthesized data are mainly sourced from Creative Proposer++ and Agent-based Data Generation pipelines:

\textbf{Creative Proposer++ Pipeline.}
A critical step in constructing large-scale datasets for vision forensics is obtaining high-quality, diverse prompts to guide media generators. Existing synthetic-data methods suffer from a significant domain gap: simplistic or random prompts yield unrealistic outputs that lack the semantic plausibility of real-world forgeries. To address this, as illustrated in Figure~\ref{fig:data_pipeline}, we introduce Creative Proposer++, a novel pipeline that automatically generates diverse, high-quality instructions for synthetic media generation. Built upon the Creative Proposer pipeline~\cite{dancetext}, Creative Proposer++ extends it from images to videos. Specifically, we prompt multimodal large language models (MLLMs), such as Qwen3.5-32B and GPT-5.5, to produce a detailed textual description of a real-world image or video with the instruction: ``Describe this image/video in detail so that I can prompt an image/video generation model to generate the very same image/video.'' This prompt elicits rich descriptions that capture subtle visual elements, which are then fed to image or video generators (e.g., GPT-Image-2, SeeDance 2.5) to synthesize realistic counterfeits. Moreover, inherent recognition errors of the MLLMs are leveraged as a natural form of text-content counterfeiting, further enhancing data diversity.

This design brings two key advantages. First, each synthesized sample reproduces the content of its real-world source, so the real and fake counterparts are semantically aligned by construction. Such alignment naturally prevents the forgery detector from overfitting to semantic or content cues irrelevant to the forgery task, and instead compels it to focus on forgery-relevant evidence such as generation traces and artifacts, thereby substantially improving generalization to unseen forgeries. Second, because the prompts are distilled directly from real-world media, the synthesized data inherit a diversity on par with that of the original data, avoiding the limited diversity typically caused by conventional synthesis approaches that rely on handcrafted or template-based prompts.

\textbf{Agent-based Data Generation Pipeline.} Additionally, to support data-centric training, we build an agent-based data generation pipeline, as shown in Figure~\ref{fig:data_pipeline}. After DefakerOne2 is trained on the data pool, its failure cases are collected and sent to an agent-assisted refinement module. The agent analyzes these bad cases to identify missing forgery patterns, difficult manipulation types, and domains, and then selects suitable generation or editing models to synthesize targeted samples. The newly generated data are added back to the training pool for the next round of optimization. Through this loop of training, agent-based bad-case analysis, data synthesis, and re-training, DefakerOne2 can continuously adapt to evolving forgery methods. Although the same agent-assisted refinement framework is applied across multiple domains, its execution process differs according to the characteristics of each domain. We describe these domain-specific pipeline designs in the next subsection.

\subsection{Statistics}

\subsubsection{AI-Generated Images} 
The training corpus for the AI-generated images domain comprises approximately $16\text{M}$ samples. This includes $\sim 8\text{M}$ public samples sourced from the official training splits of DiffusionForensics~\citep{dire}, CommunityForensics~\citep{CommunityForensics}, GenImage~\citep{zhu2023genimage}, LAION~\citep{schuhmann2022laion}, and ForenSynths~\citep{cnnspot}, together with $5.3\text{M}$ video frames extracted via Ada-FPS and $2.7\text{M}$ privately curated samples. In addition, we maintain an extensive pool of commercial APIs and open-source generative/editing models. Upon the release of a new AIGC model, our automated agent dynamically queries the corresponding API or engine, pairs targeted prompts with source images, and synthesizes novel samples in high-throughput batches.

\vspace{-0.1cm}
\subsubsection{DeepFake Images} 
\vspace{-0.1cm}
The DeepFake image domain encompasses roughly $24.4\text{M}$ training samples, incorporating $15.1\text{M}$ public instances from FaceForensics++~\citep{rossler2019faceforensics++}, CelebDF-v2~\citep{li2020celeb}, DFD~\cite{dfd2019}, DFDC~\cite{dfd2019}, ScaleDF~\citep{scaledf}, DF40~\citep{yan2024df40}, WDF~\citep{zi2020wilddeepfake}, and MFFI~\citep{miao2025mffi}, augmented by $9.3\text{M}$ curated private samples. Furthermore, we construct an unconstrained real-face gallery spanning broad variations in identity, head pose, facial expression, illumination, occlusion, background context, and image resolution. Analogous to the AIGC data pipeline, the agent invokes facial manipulation models to generate diverse DeepFake samples that comprehensively account for multi-attribute environmental and biometric perturbations.

\vspace{-0.1cm}
\subsubsection{Tampered Document Images} 
\vspace{-0.1cm}
For the document tampering domain, the training dataset consists of around $3.3\text{M}$ samples. This incorporates $0.95\text{M}$ public samples drawn from the training partitions of DanceText~\cite{dancetext}, DocTamper~\citep{doctamper}, OSTF~\citep{ostf}, T-SROIE~\citep{wang2022tsroie}, RTM~\citep{luo2025rtm}, SACP~\citep{sacp} and RIFLC~\citep{rific}, supplemented by $2.338\text{M}$ private real-world document samples. We establish a private document repository spanning $>4{,}000$ document classes, including credentials, contracts, invoices, and certificates. During synthesis, the agent queries the appropriate document category and executes realistic tampering operations such as text replacement, official seal tampering, layout reorganization, and localized region inpainting.

\vspace{-0.1cm}
\subsubsection{Forged Natural Images} 
\vspace{-0.1cm}
The training set for natural image manipulation contains about $3.7\text{M}$ samples. These are derived from the training partitions of MIML~\cite{qu2025webly}, CASIA-v2~\citep{CASIA_2013}, OpenSDID~\cite{opensdi}, SIDSet~\cite{sida}, MS-COCO~\cite{mscoco_2014}, So-Fake~\cite{huang2025sofakebenchmarkingexplainingsocial}, So-Fake-OOD~\cite{huang2025sofakebenchmarkingexplainingsocial}, and tampCOCO~\cite{CAT-Net2022}. For synthetic generation, the agent exploits pre-segmented semantic regions alongside operation-specific prompts to introduce classical and generative manipulations---including splicing, copy-move, object removal, inpainting, and generative local editing---with corresponding pixel-level ground-truth masks registered into the training pool.

\vspace{-0.1cm}
\subsubsection{Face Anti-Spoofing Images}
\vspace{-0.1cm}
The training data for image-based face presentation attack detection consists of approximately $1.3\text{M}$ samples. This includes public images from the CelebA-Spoof~\cite{celebaspoof} benchmark (both full-frame and face-centered crops), as well as representative video frames sampled from the MSU-MFSD~\cite{msumfsd}, OULU-NPU~\cite{oulunpu} datasets.

\vspace{-0.1cm}
\subsubsection{Face Anti-Spoofing Videos}
\vspace{-0.1cm}
For video-based face presentation attack detection, the training split comprises around $2\text{K}$ videos sourced from the MSU-MFSD~\cite{msumfsd}, OULU-NPU~\cite{oulunpu} datasets.

\vspace{-0.1cm}
\subsubsection{Document Anti-Spoofing Images}
\vspace{-0.1cm}
The document anti-spoofing image domain consists of roughly $0.1\text{M}$ samples, constructed from extracted video frames of the DLC2021~\cite{dlc2021} and SCVD~\cite{scvd} datasets.

\vspace{-0.1cm}
\subsubsection{Document Anti-Spoofing Videos}
\vspace{-0.1cm}
For document anti-spoofing videos domain, the training corpus incorporates approximately $2\text{K}$ complete video sequences collected from the training splits of the DLC2021~\cite{dlc2021} and SCVD~\cite{scvd} datasets.

\vspace{-0.1cm}
\subsubsection{AI-Generated Videos}
\vspace{-0.1cm}
The training set for the AI-generated video domain contains approximately $1.3\text{M}$ samples. This includes $0.6\text{M}$ public samples from GVF~\cite{DeCoF}, GVD~\cite{gvd}, AIGVDBench~\cite{aigcdbench}, MintVid~\cite{tan2026videoveritas}, DF40~\cite{yan2024df40}, TalkingHeadBench~\cite{xiong2026talkingheadbench}, and ComGenVid~\cite{comgenvid}, as well as the training split of GenBuster~\cite{wen2025busterx}, combined with $0.7\text{M}$ private curated video sequences.

\section{Realistic Rumor Video Benchmark (RRVBench)}
\label{sec:rrvbench}

To systematically benchmark forensic detection capabilities against frontier AI-Generated Content (AIGC) video engines under high-stakes, real-world conditions, we establish the \textbf{Realistic Rumor Video Benchmark (RRVBench)}. RRVBench comprises $1,319$ authentic videos sourced from RABench~\cite{rabench} and $11,000$ synthetic videos produced by $11$ modern generative architectures. To reflect the realistic distribution of accessible generative tools, we evaluate both state-of-the-art closed-source and open-source generators:
\begin{itemize}
    \item \textbf{Closed-Source Engines ($2,500$ videos):} SeeDance-2.5~\cite{SeeDance2.5} ($700$ videos), Happyhorse-1.1~\cite{HappyHorse1.1} ($400$ videos), Vidu2-Image~\cite{Vidu2-Image} ($400$ videos), Wan-2.6-Flash~\cite{Wan2.6} ($400$ videos), Wan-2.6~\cite{Wan2.6} ($200$ videos), Kling-V3 Video~\cite{KlingV3} ($200$ videos), and Kling-V3-Omni Video~\cite{KlingV3Omni} ($200$ videos).
    \item \textbf{Open-Source Engines ($8,500$ videos):} JoyAI-Echo~\cite{li2026joyai} ($3,920$ videos), MiniMax-H3~\cite{MiniMaxH3} ($3,336$ videos), HunYuan-Video-1.5~\cite{wu2025hunyuanvideo} ($740$ videos), and LongCatVideo~\cite{LongCatVideo} ($504$ videos).
\end{itemize}

\noindent\textbf{Generation Pipeline and Semantic Bias Mitigation.}
A persistent vulnerability of synthetic benchmarks is the confounding effect of semantic disparity between real and generated distributions. Following the Creative Proposer++ paradigm described above, we mitigate this by using Qwen3.5-32B~\cite{bai2025qwen3} to extract granular semantic descriptions from authentic source videos. These descriptions subsequently serve as conditioning prompts for the target video generators. This semantic-matching strategy ensures that detectors are evaluated purely on synthesis artifacts rather than high-level contextual or thematic cues.

\begin{figure*}[p]
    \centering
    \includegraphics[width=1.0\textwidth]{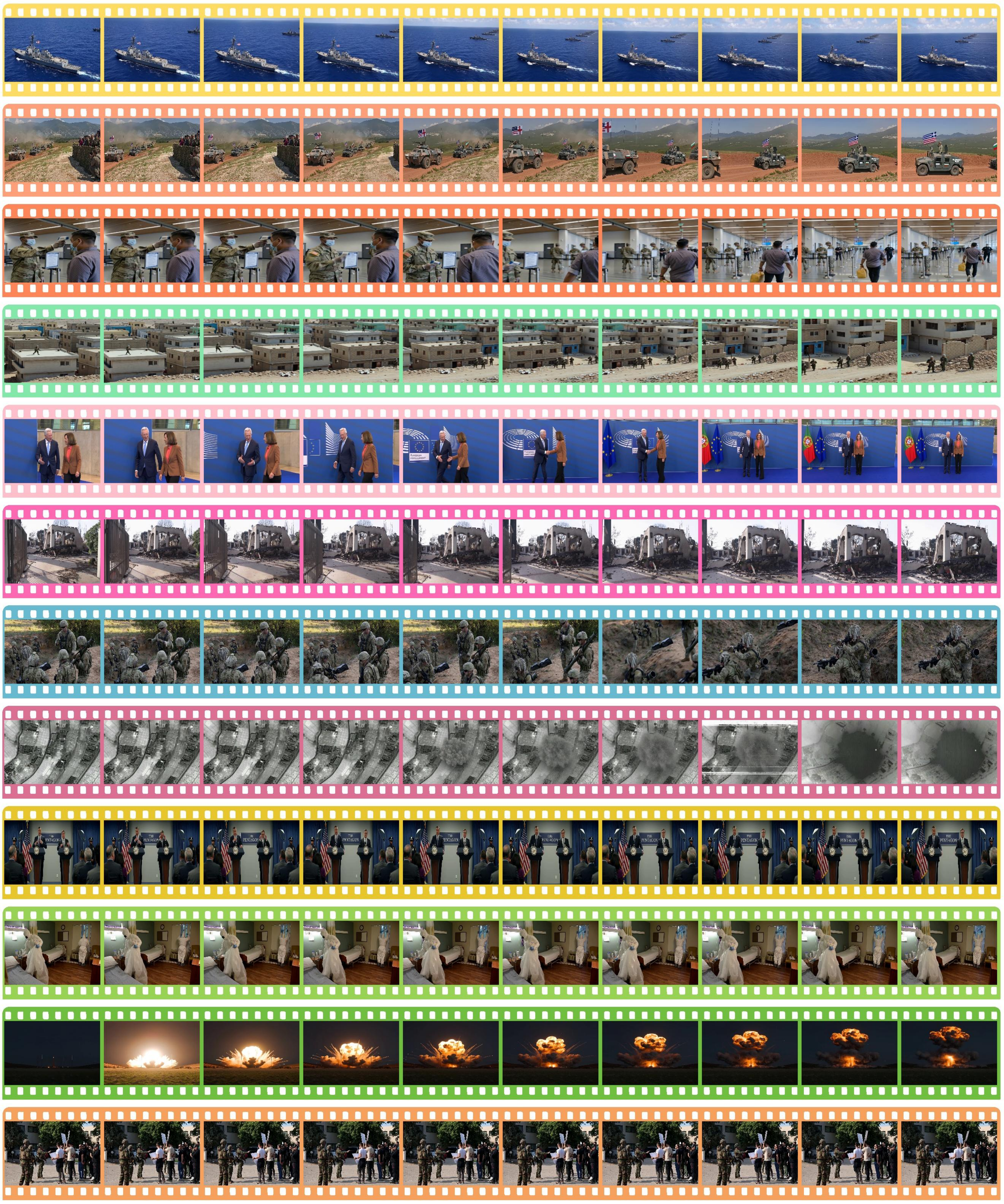}
    \caption{
    Fake video samples from our Realistic Rumor Video Benchmark (RRVBench), which focuses on real-world rumor scenarios such as military actions, political events, public unrest, health emergencies, etc. They are created by the latest video generators.
    }
    \label{fig:rrvbench}
\end{figure*}

\noindent\textbf{Target Domains and Evaluation Protocol.}
RRVBench specifically targets critical, high-impact rumor domains, including military actions, political events, civil unrest, and public health emergencies (representative examples are shown in Figure~\ref{fig:rrvbench}). 
For the evaluation protocol, each generator subset is paired against the complete, unified authentic dataset ($1,319$ videos). We calculate the balanced accuracy ($\text{bACC}$) independently for each generator to mitigate class imbalance, reporting the macro-average across all $11$ generator subsets as the final metric:
\begin{equation}
    \text{bACC}_{\text{avg}} = \frac{1}{|\mathcal{G}|} \sum_{g \in \mathcal{G}} \frac{\text{TPR}_g + \text{TNR}_g}{2},
\end{equation}
where $\mathcal{G}$ denotes the set of evaluated generative models, and $\text{TPR}_g$ and $\text{TNR}_g$ represent the true positive and true negative rates for generator $g$, respectively.

\newpage
\section{Experiments and Analyses}

\subsection{Implementation Details}
We employ the pre-trained Qwen3.5-2B~\cite{bai2025qwen3} as the backbone and initialize the DINOMAC-based~\cite{dinomac} Anomaly Magnifier with official DINOv3-Large~\cite{dinov3} weights. For the Qwen3.5-2B backbone, we adopt LoRA~\cite{hu2022lora} fine-tuning (rank $r=64$, $\alpha=128$) during the Adaptation Ebb (Stage 1) and Hard-Sample Ebb (Stage 3), using a batch size of 1,280 and a learning rate decayed from $1\times10^{-4}$ to 0. In the Full-Parameter Flood stage (Stage 2), we conduct full-model tuning on the layers targeted by LoRA with a batch size of 640 and a learning rate decayed from $1\times10^{-5}$ to 0. Experiments are distributed across 320 NVIDIA A100 GPUs with an $L_2$-norm gradient clipping threshold of 1.0.
The three stages utilize approximately 15M, 40M, and 40M samples, respectively. We apply a random sampling strategy across CommunityForensics~\cite{CommunityForensics}, LAION~\cite{schuhmann2022laion}, DiffusionForensics~\cite{dire}, GenImage~\cite{zhu2023genimage}, ScaleDF~\cite{scaledf}, DF40~\cite{yan2024df40}, MFFI~\cite{miao2025mffi}, and proprietary datasets to balance domain distributions and ensure mutually disjoint splits across stages. This disjoint split constraint applies only to the datasets listed above; all other datasets are shared across stages. A standard threshold of 0.5 is used for binarization. For the Qwen3.5-2B backbone, dynamic image resolutions range from $512\times512$ to $1024\times2352$ pixels, while video inputs are capped at 4 frames and 256 visual tokens. For the Anomaly Magnifier, input resolutions are randomly sampled from $768\times768$ to $1536\times1536$ pixels. Prior to joint end-to-end training in the final stage, the magnifier is continually pre-trained on forgery localization data for 20 epochs with its mask decoder, with a batch size of 768 and a learning rate scheduled from $1\times10^{-4}$ to 0.

\subsection{Image-level Detection Comparison}
\vspace{-0.1cm}
In this subsection, we benchmark our model against forensic-specific detectors across four domains (DeepFake, AIGC, document, and natural images). The comparison includes both non-MLLM detectors and MLLM detectors. The non-MLLM detectors include DTD~\cite{doctamper}, FFDN~\cite{chen2024ffdn}, TruFor~\cite{trufor2023}, CDFA~\cite{cdfa}, Effort~\cite{effort}, DRCT~\cite{drct} and ForensicMoE~\cite{forensic-moe}. The MLLM detectors include FakeVLM~\cite{fakeclue}, Veritas~\cite{veritas}, Ivy-xDetector~\cite{ivyfake} and our predecessor DeFakerOne~\cite{defakerone}. For every domain we report a pair of tables, one per competitor family, so that the two families are never mixed within a single ranking; because the non-MLLM and the MLLM detectors are scored on exactly the same test splits with exactly the same protocol, the two tables of a domain can be read as one comparison, and the per-domain averages are consolidated in Table~\ref{tab: det_summary}.

\subsubsection{DeepFake Detection}
\vspace{-0.1cm}
Quantitative comparisons under the AUC metric are reported in Table~\ref{tab: deepfake_vis} (non-MLLMs) and Table~\ref{tab: deepfake_mllm} (MLLMs). The evaluation covers 15 benchmarks: FF-c23~\cite{rossler2019faceforensics++}, FF-c40~\cite{rossler2019faceforensics++}, FF-DF~\cite{rossler2019faceforensics++}, FF-F2F~\cite{rossler2019faceforensics++}, FF-FS~\cite{rossler2019faceforensics++}, FF-NT~\cite{rossler2019faceforensics++}, FF-CDFv1~\cite{li2020celeb}, CDFv2~\cite{li2020celeb}, DFD~\cite{dolhansky2020dfd}, DFDC~\cite{dolhansky2019deepfake}, DFDCP~\cite{dolhansky2019deepfake}, Fsh~\cite{li2019faceshifter}, WDF~\cite{zi2020wilddeepfake}, ScaleDF~\cite{scaledf}, and MFFI~\cite{miao2025mffi}.

The DeFakerOne family achieves state-of-the-art (SOTA) performance on 14 out of 15 datasets. Specifically, DefakerOne2 attains top results on 11 benchmarks and surpasses version~1 on 12 datasets. In terms of overall average AUC, DefakerOne2 reaches 96.7\%, outperforming version~1 by 0.9\%, the strongest non-MLLM baseline Effort by 8.5\%, and the best forensic MLLM, Ivy-xDetector, by a substantial 22.7\%. This highlights our model's comprehensive facial forgery defense, robustly spotting manipulations stemming from face swapping, reenactment, attribute editing, and entire face synthesis. The result also validates that unified multi-domain training does not dilute face-specific expertise for our foundation model.

\begin{table}[h!]
  \centering
  \caption{Image-level AUC comparison with forensic vision models on DeepFake detection benchmarks. F.MoE: abbreviation of ForensicMoE, Ours V1: abbreviation of DefakerOne, Ours V2: abbreviation of DefakerOne2}
  \vspace{-0.2cm}
\begin{tabular}{lccccccccc}
\toprule
\textbf{Method} & DTD & FFDN & Trufor & CDFA & Effort & DRCT & F.MOE & Ours V1 & Ours V2 \\
\midrule
FF-c23 & 21.2 & 73.4 & 53.9 & {94.2} & 92.5 & 66.7 & 47.7 & 99.4 & \textbf{99.5} \\
FF-c40 & 20.8 & 79.1 & 49.2 & 79.7 & 80.4 & 59.2 & 64.9 & 84.7 & \textbf{88.2} \\
FF-DF & 49.8 & 40.7 & 62.1 & {99.1} & 99.0 & 78.8 & 46.0 & 99.8 & \textbf{99.8} \\
FF-F2F & 50.0 & 43.8 & 53.8 & {93.8} & 92.0 & 59.3 & 36.4 & 99.2 & \textbf{99.3} \\
FF-FS & 50.2 & 42.6 & 47.0 & 96.5 & {97.2} & 76.5 & 43.1 & 99.5 & \textbf{99.6} \\
FF-NT & 49.9 & 49.7 & 52.8 & {87.4} & 81.8 & 52.5 & 34.9 & 98.8 & \textbf{99.1} \\
CDFv1 & 38.1 & 54.8 & 54.8 & 91.5 & {90.7} & 43.1 & 19.9 & 99.9 & \textbf{99.9} \\
CDFv2 & 34.2 & 61.0 & 55.0 & {91.4} & 88.2 & 48.7 & 16.2 & 99.9 & \textbf{99.9} \\
DFD & 12.4 & 74.3 & 63.2 & {93.2} & 92.2 & 77.4 & 81.9 & \textbf{98.7} & 96.0 \\
DFDC & 47.9 & 47.7 & 51.8 & {84.2} & 82.2 & 61.2 & 30.6 & 85.2 & \textbf{90.0} \\
DFDCP & 23.0 & 57.3 & 52.0 & \textbf{93.3} & 90.9 & 65.7 & 16.3 & {91.0} & 91.8 \\
Fsh & 50.0 & 43.1 & 54.0 & 77.7 & {88.9} & 61.6 & 41.2 & 92.5 & \textbf{99.7} \\
WDF & 35.6 & 63.9 & 56.2 & 84.0 & {85.9} & 63.4 & 21.1 & \textbf{93.0}  & 92.5\\
ScaleDF & 22.4 & 73.7 & 44.2 & 76.0 & {83.2} & 74.6 & 55.8 & \textbf{98.6} & 98.0 \\
MFFI & 46.1 & 55.2 & 51.0 & 76.5 & {78.6} & 47.8 & 58.1 & 96.1 & \textbf{97.3} \\
Average & 36.8 & 57.4 & 53.4 & 87.9 & {88.2} & 62.4 & 40.9 & 95.8 & \textbf{96.7} \\
\bottomrule
\end{tabular}
\label{tab: deepfake_vis}
\vspace{+0.6cm}
  \caption{Image-level AUC comparison with forensic multi-modal large language models on DeepFake detection benchmarks}
  \vspace{-0.2cm}
\begin{tabular}{lccccc}
\toprule
\textbf{Method} & FakeVLM & Veritas & Ivy-xDetector & DefakerOne & DefakerOne2 \\
\midrule
FF-c23 & 83.4 & 4.9 & 71.7 & 99.4 & \textbf{99.5} \\
FF-c40 & 68.8 & 4.4 & {83.1} & 84.7 & \textbf{88.2} \\
FF-DF & 63.9 & 3.1 & 73.4 & 99.8 & \textbf{99.8} \\
FF-F2F & 60.1 & 3.9 & 67.3 & 99.2 & \textbf{99.3} \\
FF-FS & 63.3 & 3.3 & 68.7 & 99.5 & \textbf{99.6} \\
FF-NT & 61.7 & 2.8 & 66.6 & 98.8 & \textbf{99.1} \\
CDFv1 & 72.2 & 7.8 & 53.0 & {99.9} & \textbf{99.9} \\
CDFv2 & 81.9 & 7.9 & 53.6 & {99.9} & \textbf{99.9} \\
DFD & 88.9 & 7.4 & 77.9 & \textbf{98.7} & {96.0} \\
DFDC & 60.1 & 4.5 & {84.2} & 85.2 & \textbf{90.0} \\
DFDCP & 84.7 & 5.5 & 90.7 & {91.0} & \textbf{91.8} \\
Fsh & 44.6 & 2.3 & 67.3 & 92.5 & \textbf{99.7} \\
WDF & 63.9 & 6.3 & 82.1 & \textbf{93.0} & 92.5 \\
ScaleDF & 73.5 & 81.0 & 82.9 & \textbf{98.6} & 98.0 \\
MFFI & 62.6 & 59.2 & 87.4 & 96.1 & \textbf{97.3} \\
Avg & 68.9 & 13.6 & 74.0 & 95.8 & \textbf{96.7} \\
\bottomrule
\end{tabular}
\label{tab: deepfake_mllm}
\end{table}

\begin{table}[t!]
  \centering
  \caption{Image-level accuracy comparison with forensic vision models on AI-generated image detection benchmarks.  F.MoE: abbreviation of ForensicMoE, Ours V1: abbreviation of DefakerOne, Ours V2: abbreviation of DefakerOne2}
  \vspace{-0.2cm}
\begin{tabular}{lccccccccc}
\toprule
\textbf{Method} & DTD & FFDN & Trufor & CDFA & Effort & DRCT & F.MOE & Ours V1 & Ours V2 \\
\midrule
ForenSynths & 48.1 & 57.3 & 57.6 & 49.8 & {74.8} & 68.6 & 71.5 & 77.2 & \textbf{79.0} \\
D.Forensics & 21.8 & 55.2 & 51.4 & 59.7 & 71.2 & 72.9 & 70.9 & 99.5 & \textbf{99.8} \\
GenImage & 43.8 & 55.1 & 53.4 & 59.0 & 90.5 & 87.8 & 94.3 & 99.7 & \textbf{99.9} \\
Chameleon & 36.7 & 37.9 & 41.2 & 55.2 & 63.1 & 51.6 & 58.9 & 84.7 & \textbf{86.8} \\
Fakeinversion & 83.8 & 38.4 & 36.7 & 42.3 & {84.6} & 52.4 & 81.2 & 91.8 & \textbf{98.0} \\
BFree-Online & 36.6 & 67.8 & 60.7 & 38.3 & 55.9 & {71.0} & 32.9 & 65.7 & \textbf{91.7} \\
SynthWildx & 51.8 & 43.8 & 45.7 & 41.1 & 57.0 & 69.0 & 30.2 & 80.1 & \textbf{91.4} \\
EvalGEN & 24.7 & 40.2 & 17.6 & 33.2 & 35.8 & 77.7 & 18.9 & 96.4 & \textbf{99.9} \\
HydraFake & 54.6 & 49.6 & 51.1 & 69.3 & 62.3 & 62.7 & 60.4 & 92.8 & \textbf{93.5} \\
Average & 44.7 & 49.5 & 46.2 & 49.8 & 66.1 & 68.2 & 57.7 & 87.5 & \textbf{93.3} \\
\bottomrule
\end{tabular}
\label{tab: aigc_vis}
\vspace{+0.6cm}
  \caption{Image-level accuracy comparison with multi-modal large language models on AI-generated image detection benchmarks}
  \vspace{-0.2cm}
\begin{tabular}{lccccc}
\toprule
\textbf{Method} & FakeVLM & Veritas & Ivy-xDetector & DefakerOne & DefakerOne2 \\
\midrule
ForenSynths & 71.3 & 69.5 & 77.0 & 77.2 & \textbf{79.0} \\
D.Forensics & 74.9 & 66.5 & {93.5} & 99.5 & \textbf{99.8} \\
GenImage & {99.0} & 80.9 & 96.3 & {99.7} & \textbf{99.9} \\
Chameleon & 62.9 & 59.6 & {73.2} & 84.7 & \textbf{86.8} \\
Fakeinversion & 70.0 & 57.6 & 66.7 & 91.8 & \textbf{98.0} \\
BFree-Online & {78.6} & 55.2 & 65.5 & 65.7 & \textbf{91.7} \\
SynthWildx & 80.2 & {83.2} & {81.9} & 80.1 & \textbf{91.4} \\
EvalGEN & 92.1 & 90.4 & {93.2} & 96.4 & \textbf{99.9} \\
HydraFake & 77.8 & {87.2} & 71.7 & 92.8 & \textbf{93.5} \\
Average & 78.5 & 72.2 & {79.9} & 87.5 & \textbf{93.3} \\
\bottomrule
\end{tabular}
\label{tab: aigc_mllm}
\end{table}

\subsubsection{AI-Generated Image Detection}
\vspace{-0.1cm}
Accuracy-based comparisons for AI-generated image detection are summarized in Table~\ref{tab: aigc_vis} (non-MLLMs) and Table~\ref{tab: aigc_mllm} (MLLMs). The benchmarks include ForenSynths~\cite{cnnspot}, Diffusion Forensics~\cite{diffusion_detection}, GenImage~\cite{zhu2023genimage}, Chameleon~\cite{aide}, Fakeinversion~\cite{fakeinversion}, BFree-Online~\cite{bfree}, SynthWildx~\cite{SynthWildx}, EvalGEN~\cite{evalgen}, and HydraFake~\cite{veritas}.

The DeFakerOne series sweeps all 9 benchmarks with SOTA results. Notably, DefakerOne2 leads across all 9 datasets, consistently outperforming version~1. On average accuracy, DefakerOne2 achieves 93.3\%, marking a 5.8\% gain over version~1 and exceeding the runner-up non-series model, Ivy-xDetector, by 13.4\%; against the best non-MLLM baseline, DRCT (68.2), the margin widens to 25.1\%. These results validate our framework's full-spectrum robustness across varied image semantics and disparate generative architectures.

Two observations qualify this sweep. First, the ranking of the baselines is unstable across generators while ours is nearly flat. ForensicMoE swings from 94.3 on GenImage to 18.9 on EvalGEN, Effort from 90.5 to 35.8, and DRCT from 87.8 to 51.6, which is the signature of detectors that latch onto generator-specific fingerprints; DefakerOne2 instead spans only 79.0--99.9 across the same nine suites. Second, the largest version-1$\rightarrow$version-2 improvements fall on the suites whose generator distribution keeps moving---BFree-Online ($65.7\!\rightarrow\!91.7$, $+26.0$), SynthWildx ($+11.3$), Fakeinversion ($+6.2$) and EvalGEN ($+3.5$)---rather than on the static, academic splits, where version~1 was already close to saturation. Stability under generator turnover, not peak score on a single suite, is the property that matters once a new synthesizer is released after training, and it is what these tables show.

\begin{table}[t!]
  \centering
  \caption{Image-level accuracy comparison with forensic vision models on tampered document image detection benchmarks.  F.MoE: abbreviation of ForensicMoE, Ours V1: abbreviation of DefakerOne, Ours V2: abbreviation of DefakerOne2}
  \vspace{-0.2cm}
\begin{tabular}{lccccccccc}
\toprule
\textbf{Method} & DTD & FFDN & Trufor & CDFA & Effort & DRCT & F.MOE & Ours V1 & Ours V2 \\
\midrule
Doc-FCD & {73.2} & 73.1 & 68.4 & 52.3 & 60.1 & 77.0 & 38.9 & 92.8 & \textbf{99.7} \\
Doc-SCD & 69.8 & 58.7 & 67.7 & 39.8 & 26.7 & {71.8} & 24.6 & \textbf{99.8} & 97.6 \\
Doc-Test & 64.3 & 57.8 & 60.9 & 46.9 & 37.1 & {66.8} & 29.8 & \textbf{99.6} & 98.7 \\
TFR & 43.0 & 51.3 & 51.8 & 51.0 & 50.5 & 52.2 & 46.9 & \textbf{99.6} & 81.8 \\
T-IC13 & {75.3} & {61.9} & 58.8 & 48.0 & 19.0 & 48.9 & 30.5 & 54.9 & \textbf{78.5} \\
OSTF & {60.3} & 47.0 & 59.4 & 57.7 & 55.9 & 57.1 & 57.9 & 74.9 & \textbf{91.0} \\
RTM & 45.1 & 51.6 & 59.8 & 47.8 & 38.2 & {63.0} & 33.6 & \textbf{87.4} & 72.7 \\
DanceText & 49.6 & 52.5 & 50.7 & 51.2 & 47.5 & 52.2 & 49.3 & 60.0 & \textbf{88.0} \\
SynCred & 51.6 & 55.7 & 48.8 & 35.5 & 40.6 & 39.8 & 66.1 & 63.7 & \textbf{80.8} \\
Average & 59.1 & 56.6 & 58.5 & 47.8 & 41.7 & 58.8 & 42.0 & 81.4 & \textbf{87.6} \\
\bottomrule
\end{tabular}
\label{tab: doc_vis}
\vspace{+0.6cm}
  \caption{Image-level accuracy comparison with multi-modal large language models on tampered document image detection}
  \vspace{-0.2cm}
\begin{tabular}{lccccc}
\toprule
\textbf{Method} & FakeVLM & Veritas & Ivy-xDetector & DefakerOne & DefakerOne2 \\
\midrule
Doc-FCD & 1.9 & 39.4 & 25.3 & 92.8 & \textbf{99.7} \\
Doc-SCD & 1.0 & 31.2 & 25.1 & \textbf{99.8} & 97.6 \\
Doc-Test & 29.5 & 39.0 & 33.9 & \textbf{99.6} & 98.7 \\
TFR & 52.6 & 52.0 & {54.6} & \textbf{99.6} & 81.8 \\
T-IC13 & 25.3 & 38.2 & 22.8 & 54.9 & \textbf{78.5} \\
OSTF & 57.0 & 56.8 & 58.0 & 74.9 & \textbf{91.0} \\
RTM & 32.5 & 42.3 & 33.7 & \textbf{87.4} & 72.7 \\
DanceText & 53.4 & 59.1 & 50.7 & 60.0 & \textbf{88.0} \\
SynCred & 51.8 & 71.2 & 57.5 & 63.7 & \textbf{80.8} \\
Average & 33.9 & 47.7 & 40.2 & 81.4 & \textbf{87.6} \\
\bottomrule
\end{tabular}
\label{tab: doc_mllm}
\end{table}

\subsubsection{Tampered Document Detection}
Evaluation results for document tampering detection (in Accuracy) are listed in Table~\ref{tab: doc_vis} (non-MLLMs) and Table~\ref{tab: doc_mllm} (MLLMs). Benchmarks include DocTamper-FCD~\cite{doctamper}, DocTamper-SCD~\cite{doctamper}, DocTamper-Test~\cite{doctamper}, TFR~\cite{qu2026textshield}, Tampered-IC13~\cite{wang2022tpic}, OSTF~\cite{ostf}, RTM~\cite{luo2025rtm}, DanceText~\cite{dancetext}, and SynCred-Bench~\cite{yang2026syncredbenchbenchmarkingsyntheticcredibility}.

DTD, the strongest competitor, is a model trained specifically for this task, so the 28.5-point average margin cannot be attributed to a lack of domain-specific supervision on our side; it reflects the breadth of the evidence we aggregate. Every other baseline sits in a narrow 41.7--59.1 band, and the MLLMs are no better (FakeVLM 33.9, Ivy-xDetector 40.2, Veritas 47.7), which shows that strong OCR-level reading ability does not by itself convert into a tampering verdict.
The DeFakerOne family delivers SOTA performance across all 9 benchmarks. Specifically, DefakerOne2 leads in 5 of 9 benchmarks and outperforms version~1 in 5 datasets. Overall, DefakerOne2 achieves an average accuracy of 87.6\%, outperforming version~1 by 6.2\% and exceeding the strongest baseline, DTD, by 28.5\%. This confirms the model's ability to fuse multi-modal spatial cues to expose diverse text manipulation techniques.

\begin{table}[t!]
  \centering
  \caption{Image-level AUC comparison with forensic vision models on manipulated natural image detection benchmarks. F.MoE: abbreviation of ForensicMoE, Ours V1: abbreviation of DefakerOne, Ours V2: abbreviation of DefakerOne2}
  \vspace{-0.2cm}
\begin{tabular}{lccccccccc}
\toprule
\textbf{Method} & DTD & FFDN & Trufor & CDFA & Effort & DRCT & F.MOE & Ours V1 & Ours V2 \\
\midrule
CASIAv1 & 46.8 & 52.5 & 93.9 & 63.9 & 61.9 & 52.7 & 7.7 & \textbf{96.3} & 94.1 \\
COVERAGE & 51.5 & 40.5 & 72.3 & 53.6 & 49.2 & 52.9 & 50.6 & {75.9} & \textbf{80.0} \\
Columbia & 62.3 & 2.20 & \textbf{99.4} & 88.0 & 49.2 & 48.2 & 81.7 & 87.7 & {98.5} \\
NIST16 & 56.0 & 38.7 & 59.4 & 61.6 & 73.4 & {66.7} & 3.9 & 85.1 & \textbf{87.3} \\
CocoGlide & 51.9 & 29.6 & 67.1 & 67.8 & 68.2 & 72.3 & 19.2 & 76.0 & \textbf{83.0} \\
Autosplice & 42.9 & 34.3 & 64.5 & {77.7} & 82.0 & 75.5 & 37.4 & \textbf{99.9} & 98.7 \\
OpenSDI & 35.2 & 42.5 & 57.9 & 52.4 & 77.7 & {84.9} & 5.0 & 98.6 & \textbf{99.4} \\
DSO-1 & {71.0} & 35.0 & 64.5 & 64.1 & 55.9 & 53.6 & 23.2 & 82.4 & \textbf{89.5} \\
DEFACTO & 55.1 & 37.8 & {64.3} & 53.6 & 53.5 & 55.3 & 1.7 & \textbf{78.3} & 76.0 \\
Average & 52.5 & 34.8 & 71.5 & 64.7 & 63.4 & 62.5 & 25.6 & 86.7 & \textbf{89.6} \\
\bottomrule
\end{tabular}
\label{tab: nat_vis}
\vspace{+0.6cm}
  \caption{Image-level AUC comparison with multi-modal large language models on manipulated natural image detection benchmarks}
  \vspace{-0.2cm}
\begin{tabular}{lccccc}
\toprule
\textbf{Method} & FakeVLM & Veritas & Ivy-xDetector & DefakerOne & DefakerOne2 \\
\midrule
CASIAv1 & 4.2 & 3.2 & 45.8 & \textbf{96.3} & 94.1 \\
COVERAGE & 50.0 & 0.4 & 55.5 & {75.9} & \textbf{80.0} \\
Columbia & 52.6 & 2.5 & 87.6 & 87.7 & \textbf{98.5} \\
NIST16 & 58.7 & 1.2 & 55.4 & 85.1 & \textbf{87.3} \\
CocoGlide & 66.0 & 3.3 & 65.3 & 76.0 & \textbf{83.0} \\
Autosplice & 67.2 & 4.0 & 40.8 & \textbf{99.9} & 98.7 \\
OpenSDI & 73.5 & 1.9 & 69.5 & 98.6 & \textbf{99.4} \\
DSO-1 & 60.5 & 0.5 & 15.1 & 82.4 & \textbf{89.5} \\
DEFACTO & 50.0 & 0.9 & 30.9 & 78.3 & \textbf{76.0} \\
Average & 53.6 & 2.0 & 51.8 & 86.7 & \textbf{89.6} \\
\bottomrule
\end{tabular}
\label{tab: nat_mllm}
\end{table}

\subsubsection{Manipulated Natural Image Detection}
Natural image manipulation detection results under AUC are presented in Table~\ref{tab: nat_vis} (non-MLLMs) and Table~\ref{tab: nat_mllm} (MLLMs), evaluated on CASIAv1~\cite{CASIA_2013}, COVERAGE~\cite{Coverage_2016}, Columbia~\cite{Columbia_2006}, NIST16~\cite{NIST16_2019}, CocoGlide~\cite{trufor2023}, AutoSplice~\cite{jia2023autosplice}, OpenSDI~\cite{opensdi}, DSO-1~\cite{dso-1}, and DEFACTO~\cite{defacto_2019}.

The DeFakerOne series establishes SOTA records on 8 of the 9 benchmarks. DefakerOne2 achieves the highest score on 5 benchmarks and improves upon version~1 on 6 datasets, yielding a mean AUC of 89.6\% (+2.9\% over version~1; +18.1\% over the leading competitor, TruFor).

Furthermore, Table~\ref{tab:mau_compare} presents balanced accuracy on ManipulationBench~\cite{jiang2026dailybenchunifiedbenchmarkaigenerated}, a contemporary suite featuring the latest AI generators. We compare against 17 methods, including UniFD~\cite{univfd}, LOTA~\cite{wang2025lota}, DRCT~\cite{drct}, FatFormer~\cite{fatformer}, NPR~\cite{npr}, SAFE~\cite{safe}, AIDE~\cite{aide}, Effort~\cite{effort}, ForgeLens~\cite{chen2025forgelens}, LTD~\cite{ltd}, Aligned~\cite{sundara2025aligned}, B-Free~\cite{bfree}, RPOBE~\cite{rpobe}, OmniAID~\cite{guo2025omniaid}, DDA~\cite{evalgen}, DINO-Det-7B~\cite{dinodet}, and FPD~\cite{jiang2026dailybenchunifiedbenchmarkaigenerated}. Our model ranks first on average in both the pre-trained and the officially fine-tuned settings, with 71.7 against 70.8 for the best pre-trained competitor (RPOBE) and 63.2 for the best fine-tuned one (AIDE). DefakerOne2 surpasses version~1 by 23.1 points in average balanced accuracy ($48.6\!\rightarrow\!71.7$), proving its superior generalization to unseen, realistic synthetic manipulations regardless of style or content.

\begin{table}[t!]
    \centering
     \caption{Average / Balanced accuracy~(\%) comparison on ManipulationBench. F.$_{\textrm{Random}}$ and F.$_{\textrm{Object}}$ denote FLUX-Fill using random and object masks, respectively.
    F.2-Klein denotes FLUX.2-Klein.
    Q.-Edit and S.-Edit are abbreviations for Qwen-Image-Edit and Step1X-Edit-v1p2, respectively. Avg. denotes average performance of official pre-trained models. Avg.* denotes average performance of models officially fine-tuned on ManipulationBench's training split. Unreported results are denoted by ``-''}
    \resizebox{1.0\textwidth}{!}{
    \begin{tabular}{lcccccccccc}
    \hline
        \textbf{Methods}& \textbf{F.$_\textrm{Random}$} & \textbf{F.$_\textrm{Object}$}  & \textbf{F.2-klein} & \textbf{Q.-Edit} 
        & \textbf{S.-Edit} & \textbf{N.B. 2} & \textbf{GPT-I. 2}
        &\textbf{Avg.}&\textbf{Avg.*}
        \\
        \hline
            
        UniFD &50.2 &50.8 &50.7 &50.4 &50.7 &50.3 &50.2 &50.5 &53.0 \\
        LOTA &49.9 &49.9 &49.9 &50.0 &49.9 &50.0 &\textbf{100.0} &57.1 &53.2 \\
        DRCT &51.0 &50.9 &50.6 &52.2 &50.1 &57.2 &64.0 &53.7 &60.8 \\
        FatFormer &49.9 &50.0 &50.1 &50.1 &50.3 &50.0 &57.0 &51.0 &- \\
        NPR &45.1 &46.0 &49.8 &55.2 &48.8 &55.5 &80.3 &54.4 &61.0 \\
        SAFE &52.2 &51.9 &50.0 &49.0 &48.3 &49.5 &86.7 &55.4 &- \\
        AIDE &51.1 &50.7 &49.6 &58.6 &57.5 &54.4 &60.4 &54.6 &63.2 \\
        Effort &45.6 &45.7 &50.7 &57.0 &53.5 &60.9 &69.0 &54.6 &55.2 \\
        Forgelens &50.0 &50.0 &50.1 &51.9 &51.2 &53.2 &93.5 &57.1 &63.1 \\
        LTD &49.9 &49.9 &50.0 &50.0 &50.0 &50.0 &73.4 &53.3 &60.6 \\
        Aligned &50.5 &50.4 &50.4 &73.2 &50.4 &64.1 &57.0 &56.6 &60.5 \\
        B-Free &62.4 &70.0 &63.6 &80.0 &64.6 &45.4 &51.9 &62.5 &62.2 \\
        RPOBE &61.7 &62.9 & \textbf{65.5} &\textbf{91.5} &61.8 &69.6 &82.5 &70.8 &- \\
        OmniAID &52.7 &53.4 &56.4 &88.5 &61.5 &\textbf{71.8} &63.1 &63.9 &52.4 \\
        DDA &54.5 &54.8 &62.3 &89.2 &62.9 &62.5 &82.0 &66.9 &62.4 \\
        DINO-Det-7B &52.2 &53.0 &54.0 &69.0 &55.5 &52.0 &71.6 &58.2 &58.8 \\
        FPD &53.7 &54.1 &58.0 &80.5 &60.7 &53.2 &83.7 &63.4 &62.2 \\ \hline
        Ours V1 &53.2 & 52.9 & 51.5 & 52.4 & 52.3 & 31.8 & 45.9 & 48.6 & 48.6 \\
        Ours V2 & \textbf{77.8} & \textbf{76.1} & 63.8 & 78.1 & \textbf{70.3} & 57.2 & 78.7 & \textbf{71.7} & \textbf{71.7}
         \\
         \hline
    \end{tabular}
    }
    \label{tab:mau_compare}
    \vspace{-2mm}
\end{table}

\subsubsection{Physical Face and Document Image Forgery Detection}
Physical face and document spoofing attacks (print, replay, and mask presentations) are not tabulated separately here, as the physical benchmarks CASIA-FASD, DLC2021, and SCVD overlap directly with the video-level evaluation (where models process decomposed frame-level inputs). Reporting them redundantly would generate duplicate tables and fragmented discussions of identical experimental phenomena. Consequently, we consolidate their results in Section~\ref{sec: physical}, where per-benchmark Accuracy and HTER metrics are jointly presented alongside the digital video forgery suites; we refer readers to that section for comprehensive analyses. Furthermore, the aggregated per-domain averages are included in the \emph{Physical} rows of both the image- and video-level sections in Table~\ref{tab: det_summary}, enabling direct cross-domain comparisons within a single unified view.

\begin{table}[]
  \centering
  \small
  \caption{Video-level accuracy comparison with forensic vision models on video forgery detection benchmarks.  F.MoE: abbreviation of ForensicMoE, Ours V1: abbreviation of DefakerOne, Ours V2: abbreviation of DefakerOne2}
  \vspace{-0.2cm}
\setlength{\tabcolsep}{5pt}
\begin{tabular}{ccccccccccccc}
\hline
dataset & FLIP & MVP & LCDF & DTD & FFDN & Trufor & CDFA & Effort & DRCT & F.MoE & Ours V1 & Ours V2 \\ \hline
CASIA-FASD & \textbf{98.5} & 93.8 & 54.7 & 25.0 & 56.2 & 28.5 & 45.3 & 79.0 & 49.0 & 25.0 & 64.3 & 90.3 \\
DLC2021 & 55.5 & 55.5 & 56.9 & 44.5 & 55.3 & 39.4 & 34.8 & 54.7 & 49.6 & 44.2 & 59.6 & \textbf{99.9} \\
GenBuster++ & 50.5 & 71.6 & 49.8 & 50.0 & 47.0 & 51.0 & 48.8 & 59.4 & 49.6 & 57.8 & 81.4 & \textbf{88.6} \\
GenBuster-ID & 49.1 & 68.9 & 51.9 & 50.0 & 51.6 & 50.4 & 58.0 & 58.8 & 71.4 & 69.7 & 80.7 & \textbf{99.9} \\
GenBuster-OOD & 49.5 & 77.9 & 53.3 & 50.0 & 44.7 & 51.7 & 59.6 & 54.9 & 67.7 & 60.2 & 83.5 & \textbf{99.6} \\
LOKI & 52.5 & 55.5 & 55.2 & 33.6 & 63.1 & 35.8 & 51.1 & 57.2 & 63.2 & 33.6 & 64.9 & \textbf{92.9} \\
SCVD & 73.5 & 73.5 & 51.3 & 26.5 & 68.6 & 34.6 & 37.8 & 69.2 & 35.1 & 26.5 & 28.1 & \textbf{97.3} \\
VideoPhy2 & 54.4 & 61.4 & 70.0 & 0.0 & 63.3 & 8.2 & 43.1 & 67.5 & 58.6 & 7.6 & 38.9 & \textbf{99.4} \\ \hline
Average & 60.4 & 69.8 & 55.4 & 35.0 & 56.2 & 37.5 & 47.3 & 62.6 & 55.5 & 40.6 & 62.7 & \textbf{96.0} \\ \hline
\end{tabular}
\label{tab: videos_vis}

\vspace{+0.6cm}
  \caption{Video-level accuracy comparison with multi-modal large language models on video forgery detection benchmarks}
  \vspace{-0.2cm}
\begin{tabular}{ccccccc}
\hline
dataset & BusterX & VideoVeritas & Ivy-xDetector & Veritas & Ours V1 & Ours V2 \\ \hline
CASIA-FASD & 25.8 & 34.2 & 33.7 & 63.7 & 64.3 & \textbf{90.3} \\
DLC2021 & 46.1 & 45.1 & 46.9 & 61.2 & 59.6 & \textbf{99.9} \\
GenBuster++ & 74.8 & 85.8 & 68.5 & 81.7 & 81.4 & \textbf{88.6} \\
GenBuster-ID & 80.0 & 93.0 & 52.6 & 81.2 & 80.7 & \textbf{99.9} \\
GenBuster-OOD & 87.6 & 90.2 & 52.5 & 87.4 & 83.5 & \textbf{99.6} \\
LOKI & 62.6 & 75.8 & 70.3 & 63.5 & 64.9 & \textbf{92.9} \\
SCVD & 28.7 & 30.8 & 26.0 & 28.6 & 28.1 & \textbf{97.3} \\
VideoPhy2 & 14.5 & 91.1 & 77.9 & 38.1 & 38.9 & \textbf{99.4} \\ \hline
Average & 52.5 & 68.3 & 53.5 & 63.2 & 62.7 & \textbf{96.0} \\ \hline
\end{tabular}
\label{tab: videos_mllm}

\vspace{+0.6cm}
  \caption{Video-level Half Total Error Rate (HTER, lower is better) comparison with forensic vision models on video forgery detection benchmarks.  F.MoE: ForensicMoE, Ours V1: DefakerOne, Ours V2: DefakerOne2}
  \vspace{-0.2cm}
\setlength{\tabcolsep}{5pt}
\begin{tabular}{ccccccccccccc}
\hline
dataset & FLIP & MVP & LCDF & DTD & FFDN & Trufor & CDFA & Effort & DRCT & F.MoE & Ours V1 & Ours V2 \\ \hline
CASIA-FASD & \textbf{2.3} & 4.6 & 41.8 & 50.0 & 35.2 & 49.0 & 50.9 & 41.3 & 40.4 & 50.0 & 28.7 & 18.9 \\
DLC2021 & 50.0 & 50.0 & 43.5 & 50.0 & 50.2 & 56.8 & 63.3 & 50.4 & 47.3 & 50.5 & 39.5 & \textbf{0.1} \\
GenBuster++ & 49.5 & 28.4 & 50.2 & 50.0 & 53.0 & 49.0 & 51.2 & 40.6 & 50.3 & 42.2 & 18.6 & \textbf{11.4} \\
GenBuster-ID & 50.8 & 31.1 & 48.0 & 50.0 & 48.4 & 49.5 & 42.0 & 41.2 & 28.6 & 30.3 & 19.4 & \textbf{0.1} \\
GenBuster-OOD & 50.4 & 22.1 & 46.7 & 50.0 & 55.3 & 48.3 & 40.4 & 45.1 & 32.3 & 39.9 & 16.5 & \textbf{0.4} \\
LOKI & 49.5 & 47.1 & 50.6 & 50.0 & 44.1 & 50.3 & 46.5 & 49.3 & 32.7 & 50.8 & 32.4 & \textbf{6.9} \\
SCVD & 50.0 & 50.0 & 48.8 & 50.0 & 52.7 & 58.8 & 61.2 & 44.5 & 44.8 & 50.7 & 49.5 & \textbf{1.8} \\
VideoPhy2 & 22.8 & 19.3 & 15.0 & 50.0 & 18.4 & 45.9 & 28.4 & 16.2 & 20.7 & 46.2 & 30.5 & \textbf{0.3} \\ \hline
Average & 40.7 & 31.6 & 43.1 & 50.0 & 44.7 & 51.0 & 48.0 & 41.1 & 37.1 & 45.1 & 29.4 & \textbf{5.0} \\ \hline
\end{tabular}
\label{tab: videos_vis_hter}

\vspace{+0.6cm}
  \caption{Video-level Half Total Error Rate (HTER, lower is better) comparison with multi-modal large language models on video forgery detection benchmarks}
  \vspace{-0.2cm}
\begin{tabular}{ccccccc}
\hline
dataset & BusterX & VideoVeritas & Ivy-xDetector & veritas & Ours V1 & Ours V2 \\ \hline
CASIA-FASD & 49.4 & 51.6 & 53.6 & 32.4 & 28.7 & 18.9 \\
DLC2021 & 49.4 & 49.6 & 55.0 & 40.9 & 39.5 & 0.1 \\
GenBuster++ & 25.2 & 14.2 & 31.5 & 18.4 & 18.6 & 11.4 \\
GenBuster-ID & 20.0 & 7.0 & 47.4 & 18.9 & 19.4 & 0.1 \\
GenBuster-OOD & 12.5 & 9.8 & 47.5 & 12.6 & 16.5 & 0.4 \\
LOKI & 30.3 & 32.1 & 38.9 & 33.0 & 32.4 & 6.9 \\
SCVD & 49.2 & 54.2 & 55.6 & 51.8 & 49.5 & 1.8 \\
VideoPhy2 & 42.7 & 4.5 & 11.1 & 31.0 & 30.5 & 0.3 \\ \hline
Average & 34.8 & 27.9 & 42.6 & 29.9 & 29.4 & 5.0 \\ \hline
\end{tabular}
\label{tab: videos_mllm_hter}
\end{table}

\subsection{Video-level Detection Comparison}
In this subsection, we compare our model with models for the physical anti-forgery domain, including FLIP~\cite{FLIP}, MVP~\cite{MVPFAS} and LCDF~\cite{LCDF}, as well as models for the video anti-forgery domain, including BusterX~\cite{wen2025busterx}, VideoVeritas~\cite{tan2026videoveritas}, Ivy-xDetector~\cite{ivyfake} and Veritas~\cite{veritas}. We further include the forensic-specific models from the four image detection domains (DeepFake, AIGC, document and natural images), namely DTD~\cite{doctamper}, FFDN~\cite{chen2024ffdn}, TruFor~\cite{trufor2023}, CDFA~\cite{cdfa}, Effort~\cite{effort}, DRCT~\cite{drct} and ForensicMoE~\cite{forensic-moe}, as well as our previous version DeFakerOne~\cite{defakerone}. For models incapable of processing video inputs directly, each video clip is decomposed into individual frames and evaluated at the image level under a unified protocol, ensuring a fair and direct comparison across detectors from all three domains: anti-spoofing, video forensics, and still-image forensics.

\subsubsection{Comparison on Classic Physical and Digital Forgery Benchmarks}
\label{sec: physical}

Image-level accuracy comparisons are reported in Table~\ref{tab: videos_vis} (non-MLLMs) and Table~\ref{tab: videos_mllm} (MLLMs), and the corresponding Half Total Error Rate (HTER) comparisons, where lower is better, in Table~\ref{tab: videos_vis_hter} (non-MLLMs) and Table~\ref{tab: videos_mllm_hter} (MLLMs). The eight benchmarks jointly span classic physical spoofing and modern digital video forgery: CASIA-FASD, DLC2021, GenBuster++, GenBuster-ID, GenBuster-OOD, LOKI, SCVD and VideoPhy2.

\noindent\textbf{Accuracy.}
DefakerOne2 attains the best result on 7 of the 8 benchmarks, the single exception being CASIA-FASD, where the dedicated face anti-spoofing detector FLIP remains ahead ($98.5$ vs.\ $90.3$). It reaches an average accuracy of $96.0\%$, outperforming version~1 by $33.3\%$ and surpassing the strongest external vision baseline, MVP ($69.8$), by $26.2\%$; against the best forensic MLLM, VideoVeritas ($68.3$), the margin is $27.7\%$. The improvement over version~1 is not marginal but structural: it exceeds $25$ points on five benchmarks and reaches $+69.2$ on SCVD ($28.1\!\rightarrow\!97.3$) and $+60.5$ on VideoPhy2 ($38.9\!\rightarrow\!99.4$), the two suites on which version~1 performs at or below the chance level. Version~2 additionally saturates DLC2021 ($99.9$), GenBuster-ID ($99.9$) and GenBuster-OOD ($99.6$), showing that the gain does not come from over-fitting the in-distribution generators: the OOD split, on which every competing method degrades to between $44.7$ and $90.2$, is handled almost perfectly, and the residual difficulty concentrates on GenBuster++ ($88.6$), the in-the-wild split that mixes compression and post-processing perturbations with the forgery itself.

The baseline families fail for different reasons. Most image-domain forensic models simply do not transfer to video frames: DTD ($35.0$), TruFor ($37.5$), ForensicMoE ($40.6$), CDFA ($47.3$), LCDF ($55.4$) and DRCT ($55.5$) all remain within a few points of random guessing, and DTD collapses to $0.0$ on VideoPhy2, indicating that the tampering cues they rely on---splicing boundaries, copy--move residuals, text-region inconsistencies---are absent from a frame whose entire content is synthesized rather than locally edited. The physical anti-spoofing models transfer better on the benchmarks that match their training domain (FLIP on CASIA-FASD, MVP on GenBuster-OOD) but do not generalize beyond them, with FLIP fluctuating between $49.1$ and $98.5$ across the suite. The forensic MLLMs, despite their reasoning capability, are equally limited: BusterX ($52.5$), Ivy-xDetector ($53.5$) and Veritas ($63.2$) stay close to chance on SCVD ($28.7/26.0/28.6$) and CASIA-FASD ($25.8/33.7/63.7$), and only VideoVeritas exhibits a usable level of robustness, yet with pronounced per-generator variance ($91.1$ on VideoPhy2 but $30.8$ on SCVD).

\noindent\textbf{HTER.}
The HTER tables sharpen these observations, because the metric averages the error on the real and on the fake class and therefore cannot be inflated by a biased decision threshold. DefakerOne2 again leads on 7 of the 8 benchmarks with an average HTER of $5.0$, against $29.4$ for version~1, $31.6$ for the best external vision baseline (MVP) and $27.9$ for the best forensic MLLM (VideoVeritas). In relative terms this is an $83.0\%$ reduction of the error with respect to version~1 and an $84.2\%$ reduction with respect to MVP. Five benchmarks are driven to an essentially error-free regime---DLC2021 ($0.1$), GenBuster-ID ($0.1$), VideoPhy2 ($0.3$), GenBuster-OOD ($0.4$) and SCVD ($1.8$)---and even the two hardest ones, GenBuster++ ($11.4$) and CASIA-FASD ($18.9$), remain below every competitor, with the sole exceptions of FLIP ($2.3$) and MVP ($4.6$) on CASIA-FASD.

More importantly, HTER exposes a systematic bias that the accuracy tables conceal. Several baselines obtain a moderate accuracy while their average HTER sits at roughly $35$ or higher, i.e.\ they classify one of the two classes much better than the other: DTD ($35.0$ accuracy / $50.0$ HTER), TruFor ($37.5$ / $51.0$), ForensicMoE ($40.6$ / $45.1$), CDFA ($47.3$ / $48.0$ HTER) and LCDF ($55.4$ / $43.1$) belong to this group, as do the MLLMs Ivy-xDetector ($53.5$ / $42.6$) and BusterX ($52.5$ / $34.8$). DefakerOne2 is also the only method whose two metrics agree with each other across the suite: the mean discrepancy between its accuracy and its balanced accuracy stays within one point on seven of the eight benchmarks (average $1.3$ points, with CASIA-FASD the single exception at $9.2$), whereas every competitor deviates by $3.1$ to $15.1$ points on average. Its verdicts are therefore balanced between real and forged clips rather than obtained by collapsing onto the majority class, a property that accuracy alone would not reveal and that is indispensable for deployment, where the real/fake proportion of the incoming traffic is unknown.

\noindent\textbf{Summary.}
Across both metrics and both competitor families, the results support two conclusions. First, neither physical anti-spoofing models nor AI-generated-video reasoning models, and neither still-image forensic detectors nor forensic MLLMs, cover the full spectrum of classic physical and digital video forgeries on their own; each family is competent only on the subset that resembles its training distribution. Second, DefakerOne2 closes this gap with a single set of weights, improving the average accuracy by $33.3$ points and cutting the average HTER by $24.4$ points over version~1, which shows that the unified multi-domain training together with the modality-synergistic inference transfers to the video level rather than being confined to the image domains.

\subsubsection{Comparison on RABench}

\begin{table}[t!]
\vspace{-0.6cm}
\caption{Video-level AUC comparison of traditional forensic detectors across nine RA-Bench generative subsets}
\setlength{\tabcolsep}{5pt}
\begin{tabular}{ccccccccccc}
\hline
Detector & \begin{tabular}[c]{@{}c@{}}Wan2.2\\ dyn.\end{tabular} & \begin{tabular}[c]{@{}c@{}}Wan2.2\\ Light.\end{tabular} & LTX & \begin{tabular}[c]{@{}c@{}}Omni\\ Weav.\end{tabular} & \begin{tabular}[c]{@{}c@{}}Happy\\ Horse\end{tabular} & \begin{tabular}[c]{@{}c@{}}Run\\ way\end{tabular} & Kling & \begin{tabular}[c]{@{}c@{}}SeeDance\\ 2.0\end{tabular} & \begin{tabular}[c]{@{}c@{}}Hai\\ luo\end{tabular} & \begin{tabular}[c]{@{}c@{}}All\\ avg.\end{tabular} \\ \hline
\multicolumn{11}{c}{Image / Frame-level Forensic Detectors} \\ \hline
CNNSpot & 35.5 & 43.8 & 64.0 & 45.8 & 39.2 & 31.9 & 45.4 & 40.4 & 53.9 & 44.4 \\
NPR & 50.6 & 55.2 & 58.6 & 50.9 & 54.6 & 52.3 & 44.3 & 48.7 & 54.6 & 52.2 \\
UnivFD & 32.7 & 43.0 & 51.2 & 39.9 & 43.0 & 38.4 & 39.1 & 35.0 & 44.0 & 40.7 \\
ForgeLens & 60.0 & 69.6 & 64.1 & 63.0 & 64.9 & 59.5 & 59.2 & 48.1 & 61.7 & 61.1 \\ \hline
\multicolumn{11}{c}{Video / Temporal-level Forensic Detectors} \\ \hline
DeCoF & 62.5 & 56.7 & 62.5 & 72.1 & 58.5 & 55.2 & 57.2 & 60.6 & 63.2 & 60.9 \\
D3 & 55.4 & 49.1 & 58.0 & 52.9 & 28.0 & 24.4 & 28.9 & 50.4 & 44.0 & 43.5 \\
ReStraV & 59.3 & 62.4 & 42.5 & 68.3 & 55.7 & 45.8 & 67.4 & 49.3 & 56.4 & 56.3 \\ \hline
\begin{tabular}[c]{@{}c@{}}DefakerOne2\\ (Ours V2)\end{tabular} & \textbf{98.9} & \textbf{99.7} & \textbf{99.2} & \textbf{99.5} & \textbf{99.6} & \textbf{99.9} & \textbf{99.1} & \textbf{98.4} & \textbf{99.8} & \textbf{99.3} \\ \hline
\label{tab: rabench_vis}
\end{tabular}

\vspace{+0.4cm}
\caption{Video-level balanced accuracy comparison of multi-modal large language models across nine RA-Bench generative subsets}
\setlength{\tabcolsep}{5pt}
\begin{tabular}{ccccccccccc}
\hline
Model & \begin{tabular}[c]{@{}c@{}}Wan2.2\\ dyn.\end{tabular} & \begin{tabular}[c]{@{}c@{}}Wan2.2\\ Light.\end{tabular} & LTX & \begin{tabular}[c]{@{}c@{}}Omni\\ Weav.\end{tabular} & \begin{tabular}[c]{@{}c@{}}Happy\\ Horse\end{tabular} & \begin{tabular}[c]{@{}c@{}}Run\\ way\end{tabular} & Kling & \begin{tabular}[c]{@{}c@{}}SeeDance\\ 2.0\end{tabular} & \begin{tabular}[c]{@{}c@{}}Hai\\ luo\end{tabular} & \begin{tabular}[c]{@{}c@{}}All\\ Avg.\end{tabular} \\ \hline
\multicolumn{11}{c}{General-purpose MLLMs} \\ \hline
Qwen3.5-27B & 51.3 & 55.3 & 59.7 & 54.8 & 51.6 & 55.3 & 50.0 & 49.8 & 51.4 & 53.2 \\ \cline{1-1}
\begin{tabular}[c]{@{}c@{}}Qwen3.5-\\ 122B-A10B\end{tabular} & 51.6 & 52.7 & 58.3 & 54.1 & 51.8 & 50.2 & 53.4 & 52.4 & 52.3 & 53.0 \\ \cline{1-1}
\begin{tabular}[c]{@{}c@{}}Qwen3.7-Plus\\ (thinking)\end{tabular} & 53.7 & 55.5 & 64.6 & 54.8 & 53.7 & 54.9 & 53.9 & 51.2 & 52.9 & 55.0 \\ \cline{1-1}
\begin{tabular}[c]{@{}c@{}}Gemini-3.1-\\ Pro-Preview\end{tabular} & 63.4 & 61.6 & 74.5 & 66.6 & 62.5 & 67.8 & 60.1 & 54.3 & 59.9 & 63.4 \\ \cline{1-1}
GPT-5.5 & 50.8 & 51.5 & 56.9 & 50.9 & 51.7 & 51.6 & 51.2 & 50.6 & 51.1 & 51.8 \\ \hline
\multicolumn{11}{c}{Forensic-specific MLLMs} \\ \hline
Skyra-SFT & 60.2 & 58.7 & 31.1 & 50.9 & 59.4 & 61.0 & 59.1 & 52.8 & 56.0 & 54.4 \\
Skyra-RL & 60.0 & 58.6 & 29.7 & 53.0 & 59.7 & 61.3 & 60.6 & 54.5 & 57.1 & 54.9 \\
BusterX++ & 49.8 & 51.4 & 48.9 & 49.9 & 50.1 & 50.2 & 49.7 & 49.7 & 49.9 & 50.0 \\
DefakerOne & 56.2 & 47.8 & 53.8 & 49.4 & 46.3 & 48.0 & 47.6 & 46.7 & 48.5 & 49.3 \\ \hline
\begin{tabular}[c]{@{}c@{}}DefakerOne2\\ (Ours V2)\end{tabular} & \textbf{96.3} & \textbf{96.4} & \textbf{96.3} & \textbf{96.5} & \textbf{96.4} & \textbf{96.5} & \textbf{96.4} & \textbf{95.3} & \textbf{96.5} & \textbf{96.3} \\ \hline
\end{tabular}
\label{tab: rabench_mllm}
\end{table}

Table~\ref{tab: rabench_vis} reports zero-shot AUC comparisons on RA-Bench against specialized image and video forensic detectors. Evaluated image-based methods include CNNSpot~\cite{cnnspot}, NPR~\cite{npr}, UnivFD~\cite{univfd}, and ForgeLens~\cite{chen2025forgelens}, while video-based baselines include DeCoF~\cite{DeCoF}, D3~\cite{comgenvid}, and ReStraV~\cite{ReStraV}. DefakerOne2 significantly outperforms all baselines across every generator family, attaining the highest AUC on all nine generators. Its average AUC of $99.3$ exceeds the leading image detector ForgeLens ($61.1$) by $38.2$ points and the top video model DeCoF ($60.9$) by $38.4$ points, with a per-generator margin over the best competitor that never falls below $27.4$; the temporal-modeling baselines do not help either, as D3---which reasons over clips---collapses to $28.0$ on Happy Horse and $24.4$ on Runway.

Table~\ref{tab: rabench_mllm} presents zero-shot balanced accuracy against general MLLMs and domain-specific video forensic MLLMs, including Skyra~\cite{skyra}, BusterX++~\cite{buster++}, and our predecessor DeFakerOne~\cite{defakerone}. DefakerOne2 surpasses the state-of-the-art generalist MLLM Gemini-3.1-Pro by $32.9$ points, the top-performing specialized model Skyra-RL by $41.4$ points, and nearly doubles the score of DeFakerOne with an absolute lead of $47.0$ points. These empirical gains validate our model's robust detection capability against diverse synthetic rumor videos.

Two properties of the RA-Bench results deserve emphasis, because they are not visible from the average columns alone. First, the baselines are not merely weaker but \emph{inconsistent}: the spread between their best and worst generator is typically $15$--$30$ points (ForgeLens $48.1$--$69.6$, DeCoF $55.2$--$72.1$, ReStraV $42.5$--$68.3$), and Skyra-RL, the strongest forensic MLLM, falls to $29.7$ on LTX---below chance---while reaching $61.3$ on Runway. DefakerOne2 spans $98.4$--$99.9$ in AUC and $95.3$--$96.5$ in balanced accuracy, i.e.\ its worst generator is better than the best generator of every competitor. Second, the two metrics rank our model identically, which means the AUC advantage is not obtained by a favorable threshold but reflects genuinely better-separated scores; this matters on RA-Bench, where all nine generators are unseen during training and a detector that exploits one generator family's fingerprint is rewarded on some columns and punished on others.

\begin{table}[h!]
  \centering
  \small
  \caption{Video-level balanced accuracy comparison of traditional forensic detectors across eleven RRVBench generative subsets}
  \vspace{-0.2cm}
\setlength{\tabcolsep}{6pt}
\begin{tabular}{ccccccccccccc}
\hline
Generator & FLIP & MVP & LCDF & DTD & FFDN & TruFor & CDFA & Effort & DRCT & F.MoE & Ours V1 & Ours V2 \\ \hline
\begin{tabular}[c]{@{}c@{}}Happy\\ Horse-1.1\end{tabular} & 40.0 & 46.9 & 55.5 & 50.0 & 55.9 & 49.9 & 53.6 & 43.9 & 71.5 & 50.2 & 53.9 & \textbf{96.3} \\ \cline{1-1}
\begin{tabular}[c]{@{}c@{}}Hunyuan\\ Video-1.5\end{tabular} & 59.6 & 44.8 & 55.1 & 50.0 & 49.6 & 49.6 & 54.6 & 48.0 & 64.1 & 51.5 & 50.3 & \textbf{95.9} \\ \cline{1-1}
\begin{tabular}[c]{@{}c@{}}JoyAI-\\ Echo\end{tabular} & 39.4 & 33.8 & 50.2 & 50.0 & 53.1 & 50.7 & 52.1 & 50.2 & 57.2 & 51.9 & 51.9 & \textbf{96.5} \\ \cline{1-1}
\begin{tabular}[c]{@{}c@{}}KlingV3\\ Video\end{tabular} & 51.6 & 55.7 & 63.1 & 50.0 & 42.6 & 49.3 & 60.9 & 54.3 & 48.4 & 51.6 & 56.2 & \textbf{96.1} \\ \cline{1-1}
\begin{tabular}[c]{@{}c@{}}KlingV3\\ Omni\end{tabular} & 46.9 & 66.2 & 63.6 & 50.0 & 43.8 & 49.1 & 61.7 & 55.0 & 47.1 & 52.1 & 56.2 & \textbf{96.3} \\ \cline{1-1}
\begin{tabular}[c]{@{}c@{}}LongCat\\ Video\end{tabular} & 50.6 & 58.0 & 57.1 & 50.0 & 57.2 & 49.4 & 55.2 & 49.6 & 62.2 & 58.1 & 58.7 & \textbf{95.6} \\ \cline{1-1}
\begin{tabular}[c]{@{}c@{}}MiniMax\\ H3\end{tabular} & 44.8 & 47.7 & 56.5 & 50.0 & 47.2 & 49.9 & 52.1 & 49.8 & 49.8 & 51.2 & 50.6 & \textbf{93.6} \\ \cline{1-1}
\begin{tabular}[c]{@{}c@{}}SeeDance\\ 2.5\end{tabular} & 46.3 & 55.3 & 56.5 & 50.0 & 46.9 & 49.7 & 49.9 & 49.6 & 44.7 & 52.3 & 47.7 & \textbf{88.0} \\ \cline{1-1}
\begin{tabular}[c]{@{}c@{}}Vidu2-\\ Image\end{tabular} & 51.9 & 43.9 & 42.6 & 50.0 & 49.3 & 49.8 & 56.7 & 46.1 & 69.0 & 51.9 & 45.6 & \textbf{96.5} \\ \cline{1-1}
\begin{tabular}[c]{@{}c@{}}Wan2.6\\ Video\end{tabular} & 49.4 & 51.7 & 60.4 & 50.0 & 57.6 & 49.6 & 55.4 & 49.8 & 86.4 & 50.9 & 56.2 & \textbf{96.6} \\ \cline{1-1}
\begin{tabular}[c]{@{}c@{}}Wan2.6\\ Flash\end{tabular} & 49.9 & 49.6 & 57.6 & 50.0 & 55.9 & 50.1 & 52.3 & 51.5 & 82.3 & 54.2 & 47.8 & \textbf{96.5} \\ \hline
Average & 48.2 & 50.3 & 56.2 & 50.0 & 50.8 & 49.7 & 55.0 & 49.8 & 62.1 & 52.4 & 52.3 & \textbf{95.3} \\ \hline
\end{tabular}
\label{tab: rrvbench_small}

\vspace{+0.4cm}
  \caption{Video-level balanced accuracy of multi-modal large language models across eleven RRVBench generative subsets}
  \vspace{-0.2cm}
\begin{tabular}{ccccccc}
\hline
Generator & BusterX & Ivy-xDetector & VideoVeritas & Veritas & Ours V1 & Ours V2 \\ \hline
\begin{tabular}[c]{@{}c@{}}Happy\\ Horse-1.1\end{tabular} & 60.1 & 61.9 & 70.1 & 57.0 & 53.9 & \textbf{96.3} \\ \cline{1-1}
\begin{tabular}[c]{@{}c@{}}Hunyuan\\ Video-1.5\end{tabular} & 54.3 & 52.4 & 68.4 & 51.0 & 50.3 & \textbf{95.9} \\ \cline{1-1}
\begin{tabular}[c]{@{}c@{}}JoyAI-\\ Echo\end{tabular} & 54.2 & 73.0 & 70.2 & 59.2 & 51.9 & \textbf{96.5} \\ \cline{1-1}
\begin{tabular}[c]{@{}c@{}}KlingV3\\ Video\end{tabular} & 61.5 & 72.0 & 72.2 & 55.5 & 56.2 & \textbf{96.1} \\ \cline{1-1}
\begin{tabular}[c]{@{}c@{}}KlingV3\\ Omni\end{tabular} & 61.5 & 76.5 & 76.5 & 57.8 & 56.2 & \textbf{96.3} \\ \cline{1-1}
\begin{tabular}[c]{@{}c@{}}LongCat\\ Video\end{tabular} & 57.0 & 75.2 & 76.8 & 62.5 & 58.7 & \textbf{95.6} \\ \cline{1-1}
\begin{tabular}[c]{@{}c@{}}MiniMax\\ H3\end{tabular} & 53.2 & 69.8 & 66.7 & 53.0 & 50.6 & \textbf{93.6} \\ \cline{1-1}
\begin{tabular}[c]{@{}c@{}}SeeDance\\ 2.5\end{tabular} & 51.4 & 59.3 & 62.5 & 48.9 & 47.7 & \textbf{88.0} \\ \cline{1-1}
\begin{tabular}[c]{@{}c@{}}Vidu2-\\ Image\end{tabular} & 50.2 & 57.8 & 49.0 & 50.1 & 45.6 & \textbf{96.5} \\ \cline{1-1}
\begin{tabular}[c]{@{}c@{}}Wan2.6\\ Video\end{tabular} & 64.2 & 74.0 & 79.2 & 60.8 & 56.2 & \textbf{96.6} \\ \cline{1-1}
\begin{tabular}[c]{@{}c@{}}Wan2.6\\ Flash\end{tabular} & 52.6 & 73.8 & 76.6 & 52.9 & 47.8 & \textbf{96.5} \\ \hline
Average & 56.4 & 67.8 & 69.8 & 55.3 & 52.3 & \textbf{95.3} \\ \hline
\end{tabular}
\label{tab: rrvbench_mllm}
\end{table}

\subsubsection{Comparison on RRVBench}

Zero-shot video detection results on RRVBench under the balanced accuracy metric are listed in Table~\ref{tab: rrvbench_small} (non-MLLMs) and Table~\ref{tab: rrvbench_mllm} (MLLMs). The non-MLLM baselines encompass cross-domain forensic models: FLIP~\cite{FLIP}, MVP~\cite{MVPFAS}, LCDF~\cite{LCDF}, DTD~\cite{doctamper}, FFDN~\cite{chen2024ffdn}, TruFor~\cite{trufor2023}, CDFA~\cite{cdfa}, Effort~\cite{effort}, DRCT~\cite{drct}, and ForensicMoE~\cite{forensic-moe}. Compared MLLMs comprise BusterX~\cite{wen2025busterx}, VideoVeritas~\cite{tan2026videoveritas}, Ivy-xDetector~\cite{ivyfake}, Veritas~\cite{veritas}, and DeFakerOne~\cite{defakerone}.

Consistently, DefakerOne2 sets a superior benchmark across all 11 sub-benchmarks. Our 2B-parameter model outperforms the substantially larger 7B-parameter video forensic models BusterX, Ivy-xDetector, and VideoVeritas by $38.9\%$, $27.5\%$, and $25.5\%$, respectively, while exceeding DeFakerOne by a massive $43.0\%$ margin and the best non-MLLM baseline DRCT ($62.1$) by $33.2\%$. This exceptional generalization confirms the model's zero-shot robustness against unseen, contemporary video synthesizers, underscoring its real-world utility in countering viral synthetic rumors.

RRVBench is, of the three video evaluations, the harshest test of generalization, and the baselines behave accordingly. Nine of the ten non-MLLM detectors score within $48.2$--$56.2$, statistically indistinguishable from the $50$ of a random classifier; DTD is exactly $50.0$ on every generator and TruFor varies by less than two points ($49.1$--$50.7$), which is the signature of a detector that has no usable evidence at all on this data rather than one that is merely inaccurate. Only DRCT departs from this plateau, and its profile shows why: it reaches $86.4$ on Wan2.6 Video and $82.3$ on Wan2.6 Flash but drops to $44.7$ on SeeDance 2.5 and $48.4$ on KlingV3 Video, i.e.\ it has locked onto a specific synthesizer's artifacts and inverts its decision elsewhere. The MLLMs improve the average (VideoVeritas $69.8$, Ivy-xDetector $67.8$) without changing the pattern, since their per-generator spread remains wide (VideoVeritas $49.0$--$79.2$). DefakerOne2 is the only method whose eleven scores form a narrow high band, $88.0$--$96.6$, and whose weakest entry, SeeDance~2.5, still exceeds the strongest entry of every competitor. The version-1$\rightarrow$version-2 jump of $43.0$ points ($52.3\!\rightarrow\!95.3$), which is larger than any gain observed on the image benchmarks, indicates that the video-level generalization gap is closed by the unified multi-domain representation rather than by additional video-specific supervision.

\begin{table*}[t]
\centering
\setlength{\tabcolsep}{6.5pt}
\caption{Pixel-level forgery localization performance comparison under binary F1 metric. The upper block reports document benchmarks with document-oriented baselines, and the lower block reports nature benchmarks with nature-oriented baselines}
\label{tab:2_seg}
\begin{tabular}{lccccc}
\toprule
\multicolumn{6}{c}{\textbf{Document Image Manipulation Localization Performance (Pixel-F1)}} \\
\midrule
Method & DTD & CAFTB & TIFDM & DefakerOne & DefakerOne2 \\
\midrule
DocTamperFCD  & 68.6 & 29.2 & 9.0  & 80.8 & \textbf{86.9} \\
DocTamperSCD  & 73.9 & 37.7 & 25.7 & 74.9 & \textbf{83.5} \\
DocTamperTest & \textbf{80.3} & 32.8 & 25.9 & 73.3 & 79.6 \\
Tampered IC13 & 83.4 & 83.9 & 79.7 & 90.5 & \textbf{92.1} \\
T-SROIE       & \textbf{92.1} & 91.7 & 89.4 & 81.7 & 84.6 \\
RTM           & 17.2 & 24.9 & 5.9  & \textbf{59.1} & 56.5 \\
OSTF          & 56.3 & 64.8 & 54.1 & 90.6 & \textbf{91.4} \\
DanceText     & 9.0  & 7.1  & 4.2  & 38.4 & \textbf{61.7} \\
\midrule
Average & 60.1 & 46.5 & 36.7 & 73.7 & \textbf{79.5} \\
\midrule
\multicolumn{6}{c}{\textbf{Natural Image Manipulation Localization Performance (Pixel-F1)}} \\
\midrule
Method & TruFor & APSC-Net & Mesorch & DefakerOne & DefakerOne2 \\
\midrule
CASIAv1    & 69.2 & 84.8 & 84.0 & 78.5 & \textbf{93.5} \\
COVERAGE   & 52.2 & 56.8 & 58.6 & 71.7 & \textbf{78.7} \\
Columbia   & 85.9 & 90.1 & 89.0 & 80.9 & \textbf{91.0} \\
NIST16     & 34.8 & 52.5 & 39.2 & 65.6 & \textbf{72.4} \\
CocoGlide  & 20.5 & 35.5 & 16.2 & 62.0 & \textbf{65.6} \\
AutoSplice & 39.3 & 32.4 & 24.9 & 45.7 & \textbf{67.4} \\
\midrule
Average & 50.3 & 58.7 & 52.0 & 67.4 & \textbf{78.1} \\
\bottomrule
\end{tabular}
\label{tab: loc_comp}
\vspace{+0.5cm}
\end{table*}

\subsection{Pixel-level Localization Comparison}

Table~\ref{tab: loc_comp} reports pixel-level F1 scores for tamper localization. For document forensics (top section), we benchmark against DTD~\cite{doctamper}, CAFTB~\cite{song2025caftb}, TIFDM~\cite{dong2024tifdm}, and DeFakerOne across eight datasets: DocTamper-FCD~\cite{doctamper}, DocTamper-SCD~\cite{doctamper}, DocTamper-Test~\cite{doctamper}, Tampered-IC13~\cite{wang2022tpic}, T-SROIE~\cite{wang2022tsroie}, RTM~\cite{luo2025rtm}, OSTF~\cite{ostf}, and DanceText~\cite{dancetext}. These suites span diverse modalities (e.g., receipts, contracts, IDs, scene text) and manipulation types (e.g., copy-paste, AI inpainting). For natural images (bottom section), we evaluate against TruFor~\cite{trufor2023}, APSC-Net~\cite{miml}, Mesorch~\cite{zhu2025mesoscopic}, and DeFakerOne across CASIAv1~\cite{CASIA_2013}, COVERAGE~\cite{Coverage_2016}, Columbia~\cite{Columbia_2006}, NIST16~\cite{NIST16_2019}, CocoGlide~\cite{trufor2023}, and AutoSplice~\cite{jia2023autosplice}.

Localization is a stricter criterion than detection, since it requires the model to name the pixels responsible for its verdict rather than merely to agree with the ground-truth label, and it is here that the generality of our representation is most clearly visible. The specialized document detectors degrade sharply off their own layout priors---TIFDM averages $36.7$ and collapses to $4.2$ on DanceText, CAFTB to $7.1$ on the same suite, and even DTD, the strongest of them, reaches only $9.0$ on DanceText and $17.2$ on RTM---whereas DefakerOne2 attains $61.7$ and $56.5$, i.e.\ $52.7$ and $39.3$ points above the DTD model on the two suites that break the scanned-document assumption. On natural images the same holds against segmentation-oriented baselines: the largest margins appear exactly where the manipulation is produced by a modern generator rather than by classic splicing, CocoGlide ($65.6$ vs.\ APSC-Net $35.5$) and AutoSplice ($67.4$ vs.\ TruFor $39.3$), which confirms that the anomaly magnification and the modality-synergistic inference learned for detection also transfer to dense prediction. Version~2 is not uniformly better than version~1 on the document side---it concedes $2.6$ points on RTM, a suite of rotated and perspective-distorted receipts on which version~1 was already the best method---but it never loses a benchmark to an external competitor on the natural-image side and improves the document average by $5.8$ points overall. Read together with the detection tables, the localization results show that our accuracy is not obtained from a holistic shortcut: the model that wins the image-level verdict also localizes the tampered region, which is the property required for evidence presentation in a real moderation pipeline.

\begin{table}[]
\setlength{\tabcolsep}{6pt}
\caption{Domain-wise summary of full-spectrum forgery detection. `Physical' denotes average accuracy on CASIA-FASD, DLC2021 and SCVD. `Digital' denotes average accuracy on GenBuster-ID, GenBuster-OOD, GenBuster++, LOKI, VideoPhy2.}
\small
\begin{tabular}{cccccccccccc}
\hline
Domain & DTD & FFDN & Trufor & CDFA & Effort & DRCT & F.MoE & Ivy-xDetector & Veritas & Ours V1 & Ours V2 \\ \hline
\multicolumn{12}{c}{Image Domain} \\ \hline
DeepFake & 36.8 & 57.4 & 53.4 & 87.9 & 88.2 & 62.4 & 40.9 & 74.0 & 13.6 & 95.8 & \textbf{96.7} \\
AIGC & 44.7 & 49.5 & 46.2 & 49.8 & 66.1 & 68.2 & 57.7 & 79.9 & 72.2 & 87.5 & \textbf{93.3} \\
Document & 59.1 & 56.6 & 58.5 & 47.8 & 41.7 & 58.8 & 42.0 & 40.2 & 47.7 & 81.4 & \textbf{87.6} \\
Nature & 52.5 & 34.8 & 71.5 & 64.7 & 63.4 & 62.5 & 25.6 & 51.8 & 2.0 & 86.7 & \textbf{89.6} \\
Physical & 32.0 & 60.0 & 34.2 & 39.3 & 67.6 & 44.6 & 31.9 & 35.5 & 51.2 & 50.7 & \textbf{87.2} \\ \hline
Average & 45.0 & 51.7 & 52.8 & 57.9 & 65.4 & 59.3 & 39.6 & 56.3 & 37.3 & 80.4 & \textbf{90.9} \\ \hline
\multicolumn{12}{c}{Video Domain} \\ \hline
Digital & 36.7 & 53.9 & 39.4 & 52.1 & 59.6 & 62.1 & 45.8 & 64.4 & 70.4 & 69.9 & \textbf{96.1} \\
Physical & 32.0 & 60.0 & 34.2 & 39.3 & 67.6 & 44.6 & 31.9 & 35.3 & 51.2 & 50.7 & \textbf{95.8} \\
RRVBench & 50.0 & 50.8 & 49.7 & 55.0 & 49.8 & 62.1 & 52.4 & 67.8 & 55.3 & 52.3 & \textbf{95.3} \\ \hline
Average & 39.6 & 54.9 & 41.1 & 48.8 & 59.0 & 56.3 & 43.4 & 55.8 & 59.0 & 57.6 & \textbf{95.7} \\ \hline
\end{tabular}
\label{tab: det_summary}
\end{table}

\subsection{Comparison Summary}

Table~\ref{tab: det_summary} condenses the preceding comparisons into a domain-wise view, reporting for each domain the score of every competitor and of our two versions under the metric of that domain---AUC for DeepFake and natural images, accuracy for AIGC, document and physical image forgery, and accuracy for the three video rows. Two averages are given, one per modality block, so that the image and the video results are never mixed into a single number.

DefakerOne2 attains the best score on all eight rows of the table, and the margin over the strongest external competitor in each row is never smaller than $8.5$ points (DeepFake, against Effort) and reaches $28.5$ points on document images (against DTD). Even its weakest row, physical image forgery at $87.2$, clears the best external entry of that row by $19.6$ points, so no competitor is better than our model on \emph{any} domain---not even on the domain it was designed for. On average over the five image domains, version~2 reaches $90.9$ against $80.4$ for version~1 and $65.4$ for the best external detector, Effort; over the three video domains the contrast is starker, $95.7$ against $57.6$ for version~1 and $59.0$ for the best external detectors, Effort and Veritas, which are tied on that average.

The table also makes explicit where the gain of version~2 comes from. Version~1 is already ahead of every competitor on the four image domains that its predecessor was trained on (DeepFake $95.8$, AIGC $87.5$, document $81.4$, natural $86.7$), yet it falls behind Effort by $16.9$ points on physical image forgery ($50.7$ vs.\ $67.6$) and fails to lead a single video row, where its scores of $69.9$, $50.7$ and $52.3$ are matched or exceeded by Veritas, Effort and Ivy-xDetector respectively. Version~2 keeps the image-domain advantage and repairs both weaknesses: it lifts physical image forgery from $50.7$ to $87.2$ ($+36.5$), physical video forgery from $50.7$ to $95.8$ ($+45.1$), and the RRVBench row from $52.3$ to $95.3$ ($+43.0$). The progression is therefore not a uniform scaling of the previous model's competence but a closure of the specific gaps that separated it from a full-spectrum detector.

Two further regularities are worth recording. First, the competitor averages are almost flat across domains, while their per-domain rankings rotate: a model that is best on one row can be near-chance on another. Effort, for example, leads the image block outright ($65.4$) and shares the top of the video block with Veritas ($59.0$ each), yet falls from $67.6$ on physical image forgery to $41.7$ on document images; the two methods that tie on the video average do so for opposite reasons, since Effort's $59.6$ on digital video forgery is surpassed by Veritas's $70.4$ while Veritas collapses to $2.0$ on natural images, where Effort still reaches $63.4$. No external method therefore offers a stable operating point across the eight domains. Second, the spread within our own rows is small ($87.2$--$96.7$ in the image block, $95.3$--$96.1$ in the video block), which means the unified training does not trade one domain against another. Together these observations support the claim of this work: a single 2B-parameter model, trained once and inferred with the modality-synergistic protocol, provides a detection capability that is simultaneously state-of-the-art in every domain examined and balanced across them, whereas the existing landscape requires one specialized detector per domain and still leaves physical forgery and unseen video synthesizers unsolved.

\newpage
\subsection{Ablation Study for Forgery Detection Task}

\subsubsection{Ablation on Fine-tuning Approach} 

\begin{table*}[t]
\centering
\setlength{\tabcolsep}{4pt}
\caption{Ablation of the proposed Tide-Style Tuning. Image/video-level detection task}
\label{tab:2_seg}
\begin{tabular}{ccccccccccc}
\hline
Settings &  & All-LoRA &  & All-Full &  & Ours (Stage 1) &  & Ours (Stage 2) &  & Ours (Stage 3) \\ \cline{1-1} \cline{3-3} \cline{5-5} \cline{7-7} \cline{9-9} \cline{11-11} 
Image Average &  & 87.3 &  & 87.8 &  & 84.6 &  & 88.4 &  & \textbf{90.9} \\
Video Average &  & 93.6 &  & 93.0 &  & 91.8 &  & 94.2 &  & \textbf{95.7} \\ \hline
\end{tabular}
\vspace{+0.4cm}
\label{tab: det_abl_tst}
\end{table*}

Table~\ref{tab: det_abl_tst} compares our Tide-Style Tuning against two static
fine-tuning parameterizations: \emph{All-LoRA}, which keeps the Qwen3.5 backbone frozen throughout and trains only the adapter on the full corpus, and
\emph{All-Full}, which releases all parameters from the very beginning. The three \emph{Ours} columns on the right are the cumulative checkpoints produced after the
adaptation ebb, the full-parameter flood and the hard-sample ebb, respectively,
so that consecutive columns isolate the marginal contribution of each phase.

\noindent\textbf{Neither static parameterization dominates.}
The first observation concerns the two baselines themselves. All-Full improves the image average over All-LoRA by only $+0.5$ points ($87.3\!\rightarrow\!87.8$)
while \emph{losing} $0.6$ points on video ($93.6\!\rightarrow\!93.0$), leaving
the two regimes statistically indistinguishable overall (means of $90.45$ and
$90.40$). This trade-off is exactly the dilemma analyzed in
Sec.~\ref{sec: method_ms}: enlarging the parameter budget buys fitting capacity for
the heterogeneous majority at the price of eroding the pretrained knowledge
that the temporal branch relies on, whereas restricting the budget preserves
that knowledge but cannot absorb a 50M-scale training corpus. The fact that the two
extremes land on the same overall level, but through complementary
image/video strengths, indicates that the bottleneck is not the amount of trainable parameters but the \emph{schedule} by which they are released.

\noindent\textbf{Stage 1 quantifies the cost of the constrained warm-up.}
After the adaptation ebb the model reaches $84.6$ on images and $91.8$ on
video, i.e.\ $2.7$ and $1.8$ points \emph{below} All-LoRA. This gap is expected
rather than contradictory: Stage~1 consumes only the initial 15M portion of the
corpus under light DINOMAC noise, whereas All-LoRA is trained on all 50M
samples, and its adapter is therefore asked to cover three times the data with
the same capacity. The comparison is nonetheless informative, as it confirms
empirically the underfitting argument that motivates the flood phase---a purely
adapter-based recipe, however long it is trained, saturates well below the level
that the full parameter budget can reach. Stage~1 should consequently be read as
an initialization rather than as a candidate final model.

\noindent\textbf{Stage 2 isolates the benefit of the ebb initialization.}
Merging the Stage-1 adapter into the backbone and unfreezing all parameters
under medium noise lifts the model to $88.4$ / $94.2$, the first setting that
surpasses \emph{both} static baselines on \emph{both} modalities. Since
Stage~2 and All-Full share the same trainable parameterization and the same
full corpus, the $+0.6$ (image) and $+1.2$ (video) margin is attributable to the
preceding adapter warm-up alone: having already been steered towards the
forgery-detection objective, the backbone enters the high-capacity phase in a
region of the loss landscape where destructive gradients are rare, so full
fine-tuning adds capacity without triggering the catastrophic forgetting that
degrades All-Full. The jump from Stage~1 ($+3.8$ image, $+2.4$ video) further
shows that the released parameter budget is indeed converted into fitting power
for the hard portion of the corpus.

\noindent\textbf{Stage 3 isolates the benefit of the tide schedule at equal
parameter budget.}
The final hard-sample ebb re-freezes the backbone, attaches a fresh adapter and
trains it on the new and hard samples under heavy noise, yielding $90.9$ /
$95.7$ and the best result in every column of the table. The most stringent
comparison is against All-LoRA, because both settings end with adapter-only
optimization over a comparable amount of data: the tide schedule nevertheless
delivers $+3.6$ points on images and $+2.1$ points on video, which in relative
error terms corresponds to a $28.3\%$ reduction of the image error
($12.7\!\rightarrow\!9.1$) and a $32.8\%$ reduction of the video error
($6.4\!\rightarrow\!4.3$). Against All-Full the margins are $+3.1$ and $+2.7$,
i.e.\ a $25.4\%$ and $38.6\%$ error reduction. Identical parameter budgets and
identical data therefore produce markedly different detectors, and the
difference can only come from the order in which capacity is granted: the narrow
adapter of Stage~3 operates on top of features already generalized by the flood
phase, which is what prevents it from re-fitting the easy majority that All-LoRA
memorizes from scratch.

\begin{table*}[t]
\centering
\setlength{\tabcolsep}{7pt}
\caption{Ablation of the proposed Modality Synergism. Image/video-level detection task}
\label{tab:2_seg}
\begin{tabular}{cccccc}
\hline
Acc. / Ratio   & Frame Acc.    & Video Acc. & Frame Ratio & Video Ratio & Ours Acc.     \\ \hline
CASIA-FASD     & 90.0          & 89.8       & 61.5\%        & 38.5\%        & \textbf{90.3} \\
DLC2021        & 99.9          & 99.7       & 95.4\%        & 4.6\%         & \textbf{99.9} \\
GenBuster++    & 72.2          & 88.4       & 23.7\%        & 76.3\%        & \textbf{88.6} \\
GenBuster ID   & 99.9          & 99.9       & 92.4\%        & 7.6\%         & \textbf{99.9} \\
GenBuster OOD  & 97.2          & 98.5       & 23.5\%        & 76.5\%        & \textbf{99.6} \\
RABench        & \textbf{98.2} & 87.4       & 44.0\%        & 56.0\%        & 96.3          \\
RRVBench        & \textbf{97.6} & 85.2       & 62.0\%        & 38.0\%        & 95.3          \\
LOKI           & 89.3          & \textbf{95.1}       & 32.6\%         & 67.4\%        & 92.9 \\
SCVD           & 71.9          & 95.1       & 5.4\%         & 94.6\%        & \textbf{97.3} \\
VideoPhy2 Test & 99.4          & 95.6       & 59.4\%        & 40.6\%        & \textbf{99.4} \\ \hline
Average        & 91.6          & 93.5      & 50.0\%        & 50.0\%        & \textbf{95.9} \\ \hline
\end{tabular}
\label{tab: det_abl_ms}
\end{table*}

\subsubsection{Ablation on Modality Synergism} 

Table~\ref{tab: det_abl_ms} isolates the inference-side component of Modality
Synergism. The first two columns force the same jointly trained network to a
single input pathway---Ada-FPS frame sequence or native clip---and therefore
characterize the two feature-acquisition pathways described in
Sec.~\ref{sec: method_ms}. The third and fourth columns report the fraction of test
samples for which the entropy criterion of Eq.~\eqref{eq:eta} selects each
pathway, and the last column gives the accuracy of the routed predictor.
Averaged over the ten benchmarks, the frame pathway attains $91.6$, the video
pathway $93.5$, and the routed predictor $95.9$, i.e.\ $+4.3$ and $+2.4$ points
over the respective fixed choices. In relative terms the error rate is nearly
halved with respect to the frame pathway ($8.4\!\rightarrow\!4.1$, a $\approx50\%$
reduction) and reduced by $37.4\%$ with respect to the video pathway
($6.5\!\rightarrow\!4.1$). Since both single-modality columns are produced by
the same set of weights, this improvement is attributable to the entropy-based
test adaptation alone, without any additional training or learned routing
module.

\noindent\textbf{The two pathways are complementary, not ordered.}
If one modality were uniformly stronger, a fixed choice would be optimal and
per-sample routing would be pointless. The table refutes this directly. The
frame pathway is decisive on RABench ($98.2$ vs.\ $87.4$), RRVBench ($97.6$
vs.\ $85.2$) and VideoPhy2 Test ($99.4$ vs.\ $95.6$), whereas the
video pathway is decisive on SCVD ($95.1$ vs.\ $71.9$), GenBuster++ ($88.4$
vs.\ $72.2$) and LOKI ($95.1$ vs.\ $89.3$). The large inter-modality gaps
reach $3.8$ points on VideoPhy2 Test and $16.2$ points on GenBuster++, and on six of the
ten benchmarks the weaker pathway falls below $90$. This is the empirical
counterpart of the modality dilemma: benchmarks whose forgeries are
frame-local---text and logo-level edits in RABench and RRVBench, per-frame
generation fingerprints---are damaged by dispersing attention over redundant
temporal tokens, while benchmarks whose forgeries are
trajectory-level---physically implausible motion in LOKI, whole-clip generative
statistics in GenBuster++---leave almost no intra-frame trace to be found. The
selection ratios confirm that the entropy criterion perceives this structure:
they span from $5.4\%$ (SCVD) to $95.4\%$ (DLC2021) of frame
selections and track the stronger pathway on nine of the ten benchmarks,
including the two extremes. Yet the \emph{average} ratio is $50.0\%/50.0\%$,
which shows that the routing is not the expression of a global preference but a
genuinely sample-adaptive decision that happens to balance out over the suite.

\noindent\textbf{Routing is super-additive on the hard benchmarks.}
On four benchmarks the routed predictor exceeds \emph{both} fixed modalities:
SCVD ($71.9/95.1\!\rightarrow\!97.3$), GenBuster OOD
($97.2/98.5\!\rightarrow\!99.6$), CASIA-FASD ($90.0/89.8\!\rightarrow\!90.3$)
and GenBuster++ ($72.2/88.4\!\rightarrow\!88.6$). Such a result cannot be
explained by selecting the better of two predictors, and it verifies the design
of Modality Synergism at the level of individual samples: within one benchmark,
the modality carrying the stronger evidence varies from sample to sample, so a
mixture of the two pathways covers a strictly larger region of the sample space
than either pathway alone. The effect is largest exactly where a single pathway
is weakest---SCVD gains $25.4$ points over the frame pathway and $2.2$ over the
video pathway, and GenBuster++ gains $16.4$ and $0.2$ respectively---which is
the intended behaviour of a router that is supposed to pay off most when the
fixed-modality error is highest. Aggregating the per-benchmark stronger pathway
gives an upper reference of $96.2$; the routed predictor reaches $95.9$ and
hence recovers $93.5\%$ of the gap between that reference and the frame pathway. The routing thus operates close to
the best static per-benchmark choice while requiring no benchmark identity at
test time.

\noindent\textbf{Interaction with joint modality training.}
The inference-side adaptation is only meaningful if both pathways are reliable
in the first place, which is the role of the dual loss in Eq.~\eqref{eq:jmt}:
optimizing the shared backbone against the same label under both input
representations forces it to internalize spatial and temporal evidence
simultaneously, so that either pathway can be invoked at test time. The fact
that the frame pathway reaches $99.9$ on DLC2021 and
GenBuster ID while the video pathway reaches $95.1$ on both SCVD and LOKI,
within one and the same set of weights, is consistent with this objective. The
Ada-FPS strategy complements it on the data side: without per-source rate
adaptation, a few abundant video sources would consume most of the frame budget,
and the frame pathway would be under-trained precisely on the scarce sources for
which it is the informative one. We note that Table~\ref{tab: det_abl_ms} varies
the inference-time modality only and thus does not separately quantify the
contribution of Eq.~\eqref{eq:jmt} and of Ada-FPS; ablating them would require
training single-modality variants of the network, which we leave for future
work. Even so, the two components are not independent in effect: a routing rule
that selects between an untrained and a trained pathway would merely recover the
trained one, and the observed super-additivity on SCVD, GenBuster-OOD,
GenBuster++ and CASIA-FASD is only attainable when both pathways are
individually competent.

\subsubsection{Ablation on Decoding Hyper-parameters} To assess the inference stability of DefakerOne2, we conduct comprehensive evaluations on decoding hyperparameters, with results presented in Table~\ref{tab: det_abl_dec}. Regarding random seed variations, we evaluate three distinct seeds (42, 1024, and 4096) and observe virtually identical performance across all configurations, with image accuracy stable at 90.8–91.0\% and video accuracy consistently ranging from 95.6\% to 95.7\%. This negligible variance demonstrates that our unified training paradigm effectively eliminates sensitivity to initialization stochasticity, ensuring reproducible predictions in deployment scenarios.
We further examine temperature scaling, a critical parameter controlling prediction diversity in generative models. Across temperatures ranging from 0.0 (highly deterministic) to 1.0 (increased stochasticity), DeFakerOne2 maintains remarkable stability with accuracy fluctuating within merely 0.1 percentage points (90.9–91.0\%) and the scores remaining stable at 95.5–95.7\%. The minimal performance degradation at elevated temperatures, where one might expect increased prediction variance, substantiates that the model has learned robust forensic decision boundaries rather than relying on brittle spurious correlations. These results collectively establish that DefakerOne2 delivers consistent, reliable outputs across diverse operational configurations, a critical prerequisite for forensic applications where verdict reproducibility carries legal and evidentiary significance.

\begin{table*}[t]
\centering
\caption{Ablation study of decoding hyperparameters. Image/video-level detection task}
\label{tab:2_seg}
\begin{tabular}{ccccccccc}
\hline
\multirow{2}{*}{Settings} &  & \multicolumn{3}{c}{temperature=1, seed=} &  & \multicolumn{3}{c}{seed=42, temperature=} \\ \cline{3-5} \cline{7-9} 
 &  & 42 & 1024 & 4096 &  & 0 & 0.5 & 1 \\ \cline{1-1} \cline{3-5} \cline{7-9} 
Image Average &  & 90.9 & 90.8 & 91.0 &  & 90.9 & 91.0 & 90.9 \\
Video Average &  & 95.7 & 95.7 & 95.6 &  & 95.7 & 95.5 & 95.7 \\ \hline
\end{tabular}
\label{tab: det_abl_dec}
\end{table*}

\subsubsection{Robustness evaluation} We perform image-level classification robustness evaluation on MFFI dataset, which has an official robustness evaluation sub-set, termed as MFFI-Distorted. The MFFI-Distorted sub-set simulates real-world degradation processes during data propagation, with randomly embedded multi-level interference factors including blur processing, noise interference, color distortion, geometric deformation, sharpening artifacts, compression distortion, and adversarial perturbations. Notably, our training pipeline does not mimic the MFFI official distortions specifically and our distortions have gap with the MFFI official distortions. Despite this, our model still shows significantly improved robustness. As shown in Table~\ref{tab: robust}, the version 1 DefakerOne model undergoes a 20.5\% AUC degradation on MFFI-Distorted (from 96.1 to 76.4), while our DefakerOne2 maintains an AUC score above 90 on MFFI-Distorted. The results indicate that our DefakerOne2 posseses strong robustness under realistic and challenging conditions, making it more reliable for practical deployment.

\begin{table*}[t]
\centering
\setlength{\tabcolsep}{7pt}
\caption{Robustness evaluation against distortions on MFFI dataset under AUC metric. Image-level detection task}
\vspace{-0.2cm}
\label{tab:2_seg}
\begin{tabular}{ccccccccc}
\hline
Model         &  & \multicolumn{3}{c}{DefakerOne} &  & \multicolumn{3}{c}{DefakerOne2} \\ \cline{1-1} \cline{3-5} \cline{7-9} 
Dataset/$\Delta$ &  & MFFI & MFFI-Distorted & $\Delta$ &  & MFFI  & MFFI-Distorted & $\Delta$ \\ \cline{1-1} \cline{3-5} \cline{7-9} 
AUC           &  & 96.1 & 76.4            & -20.5\% &  & 97.3  & 90.1            & -7.4\%  \\ \hline
\end{tabular}
\label{tab: robust}
\vspace{+0.2cm}
\end{table*}
\begin{table*}[t]
\centering
\setlength{\tabcolsep}{6.5pt}
\caption{Ablation study of the proposed Anomaly Magnifier and MLLM Calibration. Pixel-level forgery localization task under binary F1 metric. The upper block reports document benchmarks and the lower block reports nature benchmarks}
\vspace{-0.2cm}
\begin{tabular}{lcccc}
\toprule
\multicolumn{5}{c}{\textbf{Document Image Manipulation Localization Performance (Pixel-F1)}} \\
\midrule
Setting & (a) Baseline & (b) +A.M. (Conv) & (c) +A.M. (DINO) & (d) DefakerOne2 \\
\midrule
DocTamperFCD  & 59.7 & 76.1 & 86.7 & \textbf{86.9} \\
DocTamperSCD  & 34.6 & 68.4 & 83.4 & \textbf{83.5} \\
DocTamperTest & 41.9 & 73.9 & 79.3 & \textbf{79.6} \\
Tampered IC13 & 85.8 & 91.6 & \textbf{92.8} & 92.1 \\
T-SROIE       & 88.2 & \textbf{89.7} & 87.2 & 84.6 \\
RTM           & 27.2 & 46.2 & 51.7 & \textbf{56.5} \\
OSTF          & 80.4 & 84.5 & \textbf{91.7} & 91.4 \\
DanceText     & 48.5 & 53.9 & 59.6 & \textbf{61.7} \\
\midrule
Average & 58.3 & 73.0 & 79.1 & \textbf{79.5} \\
\midrule
\multicolumn{5}{c}{\textbf{Natural Image Manipulation Localization Performance (Pixel-F1)}} \\
\midrule
Setting & (a) Baseline & (b) +A.M. (Conv) & (c) +A.M. (DINO) & (d) DefakerOne2 \\
\midrule
CASIAv1    & 79.0 & 87.6 & \textbf{93.7} & 93.5 \\
COVERAGE   & 72.0 & 50.9 & \textbf{83.3} & 78.7 \\
Columbia   & \textbf{94.9} & 89.7 & 89.4 & 91.0 \\
NIST16     & 43.0 & 41.4 & \textbf{73.2} & 72.4 \\
CocoGlide  & 40.8 & 34.0 & 55.9 & \textbf{65.6} \\
AutoSplice & 41.4 & 36.8 & 59.8 & \textbf{67.4} \\
\midrule
Average & 61.9 & 56.7 & 75.9 & \textbf{78.1} \\
\bottomrule
\end{tabular}
\label{tab: loc_abl}
\vspace{-0.2cm}
\end{table*}

\subsection{Ablation Study for Forgery Localization Task}
The subsection evaluates forgery localization performance under different ablation settings. The results are shown in Table~\ref{tab: loc_abl}. Setting (a) is the baseline, where output mask is obtained from the Qwen3.5 vision encoder and the same convolutional decoder. In setting (b), we use a ConvNeXt-Large~\cite{Convnet_2022} backbone as the Anomaly Magnifier. In setting (c), we use the DINOMAC encoder~\cite{dinomac} backbone as the Anomaly Magnifier. In the full setting (d), we build an MLLM calibration loop upon setting (c).
\subsubsection{Ablation on Anomaly Magnifier}

We first examine the contribution of the auxiliary encoder by comparing
settings (a), (b) and (c). Averaged over all 14 benchmarks, the pixel-level F1 rises from 59.8 in the baseline to 66.0 with a ConvNeXt-Large Anomaly Magnifier and further to 77.7 with the DINOMAC Anomaly Magnifier, indicating that the vision encoder of the MLLM alone is insufficient for dense forensic prediction and that an auxiliary pixel-oriented branch is indispensable.

\noindent \textbf{Adding a generic auxiliary encoder is not enough.}
Setting (b) reveals a clear dichotomy between the two blocks. On document benchmarks, the ConvNeXt-Large branch brings a consistent and substantial improvement, raising the average from 58.3 to 73.0, with the largest gains on datasets whose tampering is region-scoped and boundary-driven
(DocTamper-{SCD}, $34.6\!\rightarrow\!68.4$;
DocTamper-{Test}, $41.9\!\rightarrow\!73.9$; RTM, $27.2\!\rightarrow\!46.2$). On natural images, however, the same module \emph{degrades} the average from 61.9 to 56.7, and COVERAGE, CocoGlide, AutoSplice, NIST16 and Columbia all fall below the baseline,
with COVERAGE dropping by 21.1 points. The reason is that ConvNeXt-Large is pre-trained with image-level classification supervision, so its features encode semantic saliency rather than forensic inconsistency. When such features are concatenated with the MLLM representation, the decoder gains capacity but no new evidence: it simply becomes better at segmenting whatever region looks prominent, which helps on splicing-style document tampering yet over-segments benign texture and fails on manipulations that leave no semantic cue, such as the diffusion-based inpainting of CocoGlide and AutoSplice. The only natural benchmark that benefits (CASIA{v1}, $79.0\!\rightarrow\!87.6$) is
precisely one dominated by obvious object-level splicing, which is consistent with this explanation.

\noindent \textbf{A forensically tuned encoder turns the extra capacity into gains.} Replacing the ConvNeXt-Large backbone with DINOMAC in setting (c) improves the document average by 6.1 points (to 79.1) and the natural average
by 19.2 points (to 75.9), removing the regression introduced in (b) and lifting 12 of the 14 benchmarks above the baseline. The improvements are concentrated where the traces are low-level rather than semantic: COVERAGE
($50.9\!\rightarrow\!83.3$), NIST16
($41.4\!\rightarrow\!73.2$), AutoSplice ($36.8\!\rightarrow\!59.8$) and CocoGlide ($34.0\!\rightarrow\!55.9$). This verifies our motivation for the DINOMAC backbone: the self-supervised DINOv3 representation provides dense, patch-aligned features suitable for pixel-wise prediction, while the LoRA
adapter and the forensic training strategy redirect these features towards artifacts such as inconsistent blur, resampling and boundary discontinuity, instead of object semantics. The two exceptions are informative as well. On
T-SROIE the DINO backbone is slightly behind both the baseline and the ConvNeXt variant ($89.7\!\rightarrow\!87.2$), which we attribute to the coarse
patchification of ViT-style encoders that discards the stroke-level high-frequency detail on which small receipt-text tampering depends, whereas the hierarchical stem of ConvNeXt preserves it. On Columbia, a small splicing-only benchmark where the baseline is already close to saturation (94.9), the additional branch mainly introduces redundant features. Both cases are limited in magnitude and do not offset the overall trend, and the average gain of setting (c) over the baseline is 20.8 points on documents and 14.0 points on natural images.

\subsubsection{Ablation on MLLM Calibration}

Setting (d) adds the calibration loop of Eqs.~\eqref{eq:dynamic-threshold}--\eqref{eq:logit-boundary} on top of setting (c). The loop introduces no trainable parameter, yet it improves the document average from 79.1 to 79.5 and the natural average from 75.9 to 78.1, i.e., $+1.2$ points over all 14 benchmarks, and it attains the best on 7 of them.

\noindent \textbf{The gains concentrate on subtle and diffuse manipulations.}
Eight benchmarks improve, with an average gain of $+3.3$ points, and the largest ones are CocoGlide ($55.9\!\rightarrow\!65.6$), AutoSplice
($59.8\!\rightarrow\!67.4$), RTM ($51.7\!\rightarrow\!56.5$) and DanceText ($59.6\!\rightarrow\!61.7$). These are exactly the cases in which the pixel branch alone under-segments: generative inpainting leaves no splicing boundary,
so the anomaly map $\mathbf{M}$ is broad but of low confidence, and text-level edits alter only a few strokes within a large authentic background. A fixed cut-off suppresses most of these weak responses, whereas the global branch correctly identifies such images as forged. With $p\!\rightarrow\!1$, the dynamic threshold is lowered to $\tau\!=\!0.25$, the admissible region is enlarged, and the diffuse responses that were already ranked correctly but sitting below the neutral boundary are recovered. The calibration therefore acts as a recall-oriented prior on hard samples, and the improvement is largest where the gap between (c) and the achievable upper bound is widest.

\noindent \textbf{Summary.} The two ablations are complementary. The Anomaly Magnifier supplies forensic evidence that the MLLM vision encoder cannot provide and accounts for the bulk of the improvement ($59.8\!\rightarrow\!77.7$), whereas the calibration loop exploits the global judgment to re-balance the localization decision boundary and contributes a further $+1.2$ points at zero parameter cost, mainly by recovering manipulations that are visible globally but weak locally.

\begin{figure}[t]
  \centering
  \includegraphics[width=\textwidth]{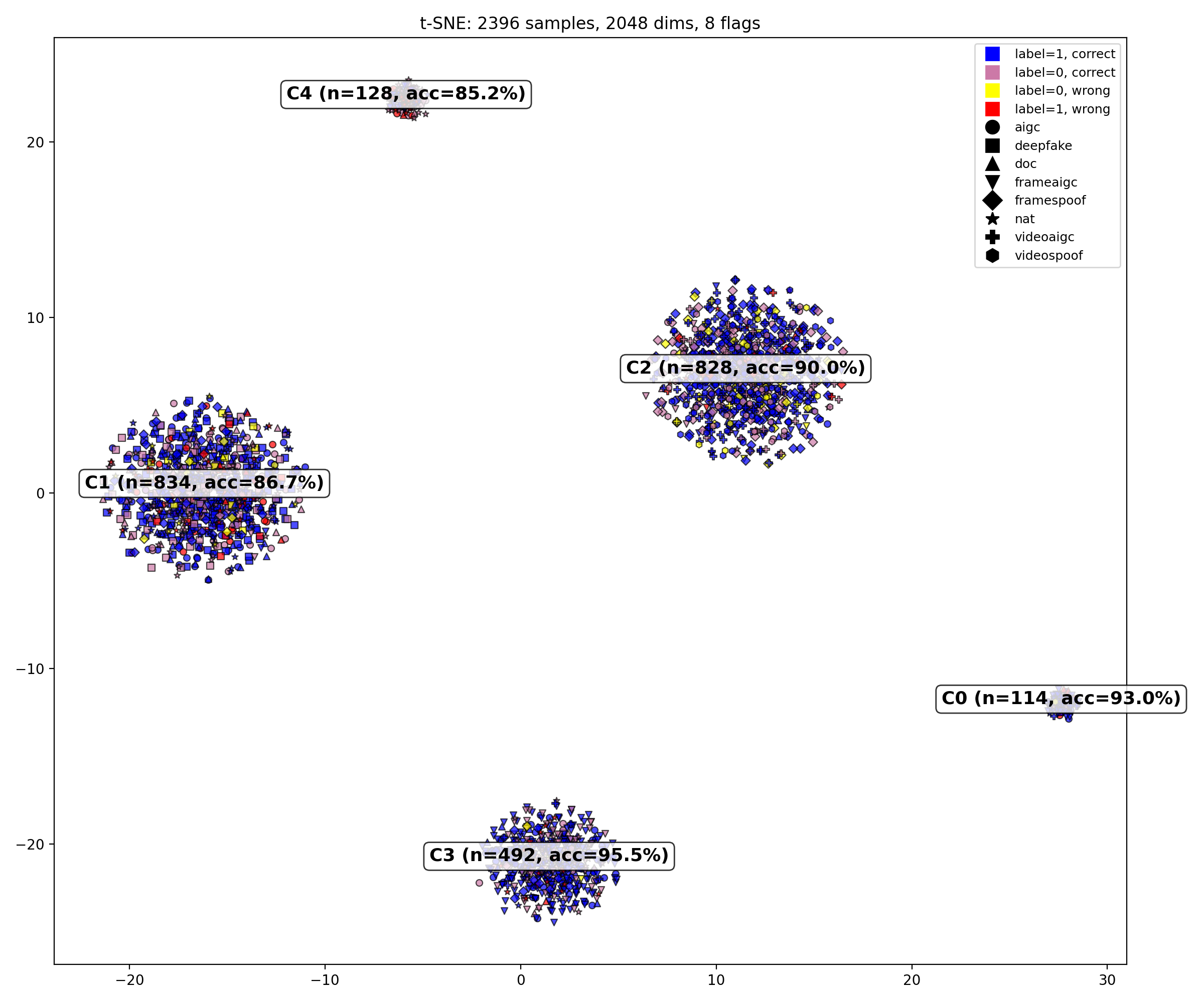}
  \caption{t-SNE projection of the last-layer hidden states preceding the LLM classification head ($2{,}396$ samples, $2{,}048$ dimensions, eight flags). All points are randomly sampled from the test sets, with no filtering by prediction. Marker shape denotes the forensic flag; marker color denotes the ground-truth label and prediction correctness (blue: fake and correct; pink: real and correct; yellow: real but wrong; red: fake but wrong). The five annotated manifolds C0--C4 are quantified in Table~\ref{tab:tsne_clusters}.}
  \label{fig:tsne}
\end{figure}

\subsection{t-SNE visualization and Analyses}
\label{subsec:tsne}

To inspect what the unified multi-domain training actually writes into the backbone, we project with t-SNE~\cite{tsne} the hidden states of the last transformer layer preceding the LLM classification head, i.e.\ the representation from which the single Real/Fake decision token is decoded. All data points are drawn by \emph{random sampling from the test sets} of the eight evaluation flags (\texttt{aigc}, \texttt{deepfake}, \texttt{doc}, \texttt{frameaigc}, \texttt{framespoof}, \texttt{nat}, \texttt{videoaigc}, \texttt{videospoof}), without any selection or filtering by model prediction, which yields $2{,}396$ test inputs of $2{,}048$ dimensions; no training sample is involved. For the sake of clarity we recall the semantics of each flag: \texttt{aigc}, \texttt{deepfake}, \texttt{doc} and \texttt{nat} denote AI-generated, face-swapped/manipulated, tampered-document and manipulated-natural still images respectively; \texttt{frameaigc} and \texttt{framespoof} are the frames cut with Ada-FPS (Sec.~2.3) from AI-generated videos and from physical presentation-attack (anti-spoofing) videos respectively, i.e.\ they are the frame-modality views of the two video families; and \texttt{videoaigc} and \texttt{videospoof} are the same two families fed through the native video modality. In Figure~\ref{fig:tsne}, marker shape encodes the forensic flag and marker color encodes the ground-truth label together with the prediction correctness; the five visually separated manifolds are annotated as clusters C0--C4 with their sizes and accuracies, and Table~\ref{tab:tsne_clusters} reports the full cluster--flag cross-tabulation, split into two sub-tables of four flags each.

\begin{table}[t]
  \centering
  \caption{Per-cluster statistics of the t-SNE projection in Figure~\ref{fig:tsne}, over $2{,}396$ inputs randomly sampled from the eight test sets (\texttt{frameaigc}/\texttt{framespoof}: frames cut with Ada-FPS from AI-generated videos / from physical presentation-attack videos; \texttt{videoaigc}/\texttt{videospoof}: the same two video families under the native video modality). The two sub-tables split the eight flags into the image-derived ones (a) and the video-derived ones (b); both share the same five clusters. Each cell reports the number of samples of that flag falling into the cluster, its share of all sampled instances of the flag (in \%), and the detection accuracy within the cell (in \%), i.e.\ \texttt{count (share\%, acc\%)}; the dominant cluster of each flag is bolded, and ``--'' denotes an empty cell. The bottom row of each sub-table aggregates over all clusters, and the \texttt{All} column reports the cluster size together with the overall cluster accuracy.}
  \label{tab:tsne_clusters}
  \footnotesize
  \setlength{\tabcolsep}{4pt}
  \renewcommand{\arraystretch}{1.15}

  \begin{center}
  \begin{tabular}{lccccc}
    \toprule
    Cluster & \texttt{aigc} & \texttt{deepfake} & \texttt{doc} & \texttt{frameaigc} & All \\
    \midrule
    C0 & 15 (4.1\%, 86.7\%)          & --                         & 11 (4.6\%, 63.6\%)          & 71 (14.1\%, 98.6\%)            & 114 (4.8\%, 93.0\%) \\
    C1 & \textbf{204 (55.4\%, 78.4\%)} & \textbf{278 (100.0\%, 91.7\%)} & \textbf{135 (56.0\%, 94.1\%)} & 36 (7.2\%, 97.2\%)            & 834 (34.8\%, 86.7\%) \\
    C2 & 24 (6.5\%, 91.7\%)          & --                         & 34 (14.1\%, 94.1\%)          & 88 (17.5\%, 88.6\%)            & 828 (34.6\%, 90.0\%) \\
    C3 & 94 (25.5\%, 100.0\%)        & --                         & 42 (17.4\%, 92.9\%)          & \textbf{277 (55.1\%, 98.2\%)}  & 492 (20.5\%, 95.5\%) \\
    C4 & 31 (8.4\%, 77.4\%)          & --                         & 19 (7.9\%, 89.5\%)           & 31 (6.2\%, 100.0\%)            & 128 (5.3\%, 85.2\%) \\
    \midrule
    All & 368 (100.0\%, 85.1\%)      & 278 (100.0\%, 91.7\%)      & 241 (100.0\%, 92.1\%)        & 503 (100.0\%, 96.6\%)          & 2{,}396 (100.0\%, 89.9\%) \\
    \bottomrule
  \end{tabular}

  (a) Image-derived flags: AI-generated imgae (aigc), deepfake, natural (nat), tampered-document (doc), physical presentation-attack frames (framespoof), and frame-sequence AI-generated samples (frameaigc).
  \end{center}

  \vspace{0.6em}

  \begin{center}
  \begin{tabular}{lccccc}
    \toprule
    Cluster & \texttt{framespoof} & \texttt{nat} & \texttt{videoaigc} & \texttt{videospoof} & All \\
    \midrule
    C0 & --                            & 10 (3.8\%, 100.0\%)          & 6 (3.1\%, 100.0\%)          & 1 (0.6\%, 0.0\%)             & 114 (4.8\%, 93.0\%) \\
    C1 & 45 (11.9\%, 84.4\%)           & \textbf{128 (48.5\%, 78.9\%)} & 8 (4.1\%, 87.5\%)           & --                           & 834 (34.8\%, 86.7\%) \\
    C2 & \textbf{312 (82.5\%, 87.8\%)} & 33 (12.5\%, 63.6\%)          & \textbf{170 (87.6\%, 93.5\%)} & \textbf{167 (98.2\%, 95.2\%)} & 828 (34.6\%, 90.0\%) \\
    C3 & 21 (5.6\%, 95.2\%)            & 46 (17.4\%, 71.7\%)          & 10 (5.2\%, 100.0\%)         & 2 (1.2\%, 100.0\%)           & 492 (20.5\%, 95.5\%) \\
    C4 & --                            & 47 (17.8\%, 78.7\%)          & --                          & --                           & 128 (5.3\%, 85.2\%) \\
    \midrule
    All & 378 (100.0\%, 87.8\%)         & 264 (100.0\%, 76.5\%)        & 194 (100.0\%, 93.8\%)       & 170 (100.0\%, 94.7\%)        & 2{,}396 (100.0\%, 89.9\%) \\
    \bottomrule
  \end{tabular}

  (b) Video-derived flags: AI-generated videos (videoaigc) and physical presentation-attack videos (videospoof).
  \end{center}
\end{table}

\noindent\textbf{The representation organizes by forensic domain, not by authenticity.}
The five manifolds recover the taxonomy of the training corpus (Sec.~3.3) rather than the binary Real/Fake label. C0 and C3 are dominated by frame-sequence AIGC samples (\texttt{frameaigc} accounts for $62\%$ and $56\%$ of their members, with per-flag accuracies of $98.6\%$ and $98.2\%$) and form the tightest and most accurate clusters ($93.0\%$ and $95.5\%$). C2 aggregates the physical-spoofing and video-forgery family (\texttt{framespoof} $38\%$, \texttt{videoaigc} $21\%$, \texttt{videospoof} $20\%$; $90.0\%$). C1, the largest manifold ($34.8\%$ of all samples), gathers the content-level forgeries---\texttt{deepfake} ($33\%$, and in fact $100\%$ of all sampled deepfake samples land here), \texttt{aigc} ($24\%$) and \texttt{doc} ($16\%$)---together with a substantial population of authentic natural images (\texttt{nat}, $15\%$ of the cluster). C4 is a small residual manifold mixing \texttt{nat} ($37\%$), \texttt{aigc} ($24\%$) and \texttt{frameaigc} ($24\%$) and attains the lowest accuracy ($85.2\%$). The geometry therefore confirms that a single backbone internalizes domain-specific forensic evidence in distinct regions of the representation space, which is precisely what makes the unified formulation of Sec.~2 feasible.

\noindent\textbf{What drives a flag into one cluster or into many.}
The degree to which each flag concentrates is not arbitrary: it is governed by two factors, the \emph{appearance homogeneity} of the content and the \emph{provenance} of the samples.

At one extreme, \texttt{deepfake} is the only flag that collapses into a \emph{single} manifold---all $278$ sampled deepfake inputs, real and fake alike, land in C1. This is expected, because a face-forgery sample is almost always a face-centric portrait: the composition, the anatomical layout, the skin texture and the illumination statistics are shared across identities and generators, so the representation is dominated by a strong ``face'' appearance prior that overrides both the manipulation type and the authenticity of the sample. \texttt{doc} behaves similarly but less absolutely: document images possess an equally idiosyncratic appearance---dense stroke-level text, tabular layout, seals, stamps and the flat, texture-poor background of a scanned page---which is visually unlike any other domain, and accordingly $56.0\%$ of them concentrate in C1 at $94.1\%$ accuracy. The residual $44\%$ nevertheless leaks into every other manifold, and the leakage is exactly where the accuracy breaks down: the $11$ document samples absorbed by the frame-AIGC manifold C0 fall to $63.6\%$, plausibly because plain, low-texture document layouts, once down-sampled or recompressed, become indistinguishable in appearance from generative frames.

At the other extreme, the video-derived flags are governed by provenance rather than by appearance. \texttt{framespoof} and \texttt{videospoof} are \emph{two views of the same source corpus}: the former are frames cut with Ada-FPS from physical presentation-attack videos, and the latter are the very same videos fed through the native video modality. They therefore share the acquisition chain end-to-end---the same cameras, the same replay devices (screens, printed photos, masks), and the same display-induced cues such as moir\'e patterns, screen borders, specular reflection and recapture blur. This shared origin is visible in the projection: $82.5\%$ of \texttt{framespoof} and $98.2\%$ of \texttt{videospoof} co-locate in C2, and the same manifold also absorbs $87.6\%$ of \texttt{videoaigc}, whose frames likewise come from a video pipeline and carry the corresponding compression and temporal-resampling statistics. In other words, C2 is best read not as a ``spoofing cluster'' but as a ``video-provenance cluster'', within which physical and digital forgery families are indistinguishable to the representation. \texttt{frameaigc}, being the frame view of AI-generated videos, is similarly split between C3 ($55.1\%$) and C0 ($14.1\%$)---two attractors that we attribute to different source-generator and frame-sampling families within the AIGC video corpus.

The most dispersed flag is \texttt{aigc}, whose samples appear in \emph{all five} clusters (C1 $55.4\%$, C3 $25.5\%$, C4 $8.4\%$, C2 $6.5\%$, C0 $4.1\%$). The reason is intrinsic to the domain: AI-generated still images are produced from open-domain prompts by a large pool of commercial and open-source generators (Sec.~3.3.1), so their semantic content spans portraits, landscapes, documents, food and objects with no shared appearance prior at all---what they share is a generation fingerprint, not a content distribution. The same argument explains why \texttt{nat} has no strong attractor either ($48.5\%$ in C1, the rest spread over C2--C4): authentic natural images cover the entire content space and act as the background population into which every other domain leaks. Tellingly, the accuracy of both flags peaks precisely where they are pulled into a coherent manifold instead of being scattered: \texttt{aigc} reaches $100.0\%$ inside the frame-AIGC manifold C3, where its generation fingerprint aligns with that of AIGC video frames, versus $78.4\%$ and $77.4\%$ in the heterogeneous C1 and C4; and \texttt{nat} attains its best scores in the small homogeneous pockets rather than in the mixed cluster. This is consistent with the overall ranking of cluster purity against cluster accuracy, and identifies content-diverse flags with no appearance prior as the population that most needs an explicit structural constraint in the embedding space.

\begin{figure}[t]
  \centering
  \includegraphics[width=\textwidth]{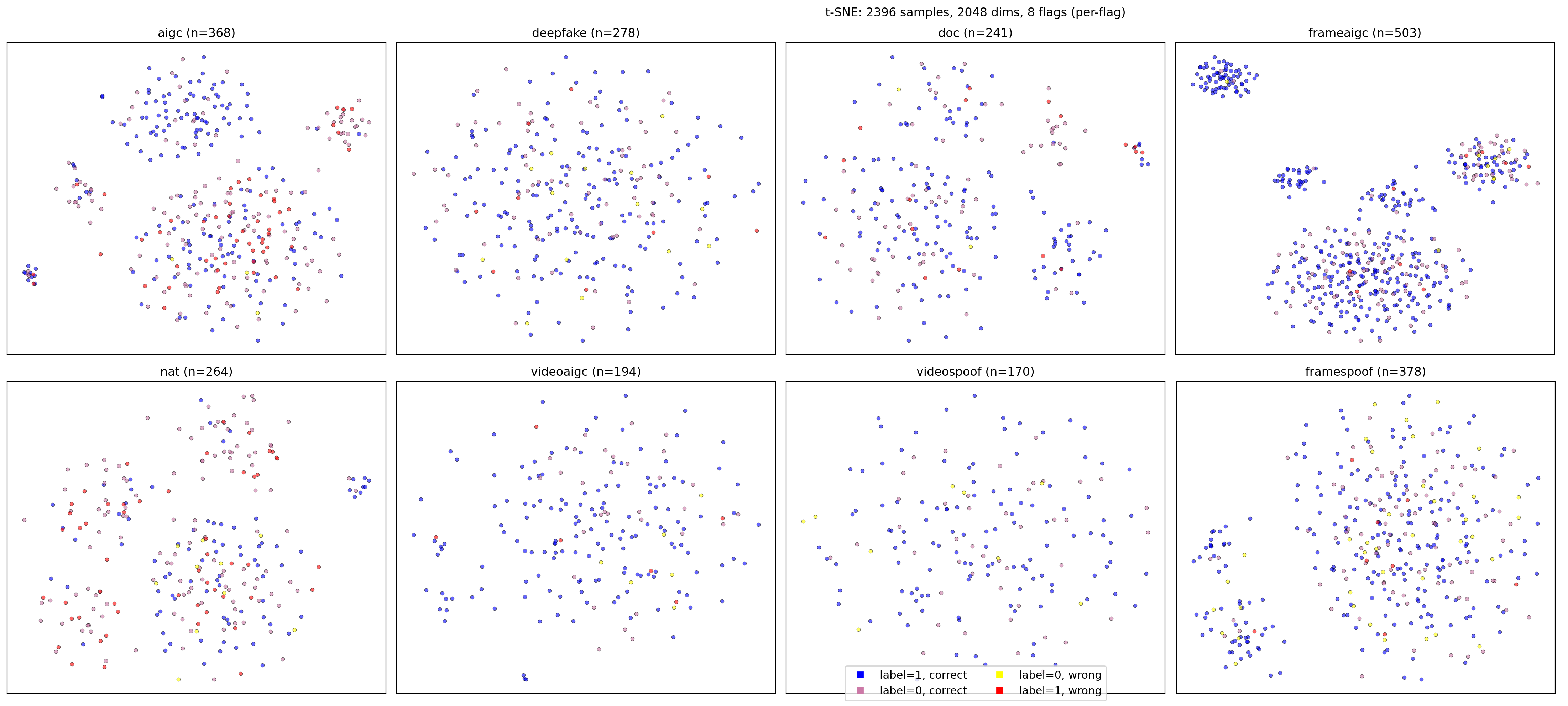}
  \caption{Per-type view of the t-SNE projection of Figure~\ref{fig:tsne}: the identical embedding ($2{,}396$ randomly sampled test inputs, $2{,}048$ dimensions) re-plotted once per forensic flag, with each panel showing only the samples of that flag ($n$ as titled; marker colors as in Figure~\ref{fig:tsne}).}
  \label{fig:tsne_perflag}
\end{figure}

To make this per-type structure explicit, Figure~\ref{fig:tsne_perflag} re-plots the \emph{same} embedding once per forensic flag, each panel isolating the samples of a single type against the identical global geometry. This per-type decomposition corroborates the conclusions drawn from Figure~\ref{fig:tsne} from the opposite direction: flags with a strong appearance or provenance prior (\texttt{frameaigc}, \texttt{deepfake} and the video-derived flags) occupy compact or contiguous regions of the shared geometry, whereas the content-diverse flags (\texttt{aigc}, \texttt{nat}, \texttt{doc}) reappear on several disjoint islands within their own panels, and in every panel the real/fake and correct/wrong markers again intermix without any separable sub-structure---reaffirming both the domain-organized clustering and the absence of an explicit authenticity boundary.

\noindent \textbf{Cluster membership, not sample difficulty alone, governs per-flag accuracy.}
No flag is uniformly accurate across the manifold: for most clusters, each flag is markedly more accurate inside its dominant cluster than outside it, indicating that part of the residual error is representation-level mis-routing rather than a lack of evidence in the input. The clearest case is \texttt{doc}: the minority of document samples projected into the frame-AIGC manifold C0 are classified at only $63.6\%$ ($n{=}11$), whereas the same flag reaches $94.1\%$ inside its dominant cluster C1 ($n{=}135$)---a gap of $30.5$ points for inputs that differ only in where the backbone places them. The pattern repeats across flags: \texttt{nat} drops to $63.6\%$ when absorbed by the spoofing/video manifold C2, against $78.7\%$ in C4 and $78.9\%$ in C1; \texttt{aigc} scores $78.4\%$ in C1 and $77.4\%$ in C4 but $100.0\%$ in C3 ($n{=}94$); \texttt{framespoof} falls to $84.4\%$ inside C1 versus $87.8\%$ in C2; and \texttt{frameaigc} degrades to $88.6\%$ in C2 versus $98.6\%$/$98.2\%$ in its own manifolds C0/C3.

Cluster-level accuracy is broadly consistent with the same logic, though the relation is a trend rather than a strict ordering. We quantify purity by the \emph{dominant-flag share}, i.e.\ the fraction of a cluster's members carrying the flag that defines it: C0 ($62.3\%$, \texttt{frameaigc}), C3 ($56.3\%$, \texttt{frameaigc}), C2 ($37.7\%$, \texttt{framespoof}), C4 ($36.7\%$, \texttt{nat}) and C1 ($33.3\%$, \texttt{deepfake}). The two clusters with the highest dominant-flag share, C0 ($93.0\%$) and C3 ($95.5\%$), are also the two most accurate, and both exceed every cluster in the lower-purity group; the rank correlation between dominant-flag share and cluster accuracy is $\rho{=}{+}0.80$ over the five clusters. The correspondence is not monotone at the bottom of the range: C1, the \emph{least} pure cluster, attains $86.7\%$ and thus edges out C4 ($85.2\%$), whose purity is marginally higher. We note two caveats before reading this as a purity effect. First, the ranking of C4 depends on the purity metric: because only four of the eight flags are populated in C4, the empty cells depress its normalized entropy ($H/\log 8 = 0.64$), which would place C4 \emph{third} in purity---between C3 and C2---rather than fourth; the dominant-flag share is therefore the more faithful measure here, since entropy conflates purity with flag sparsity. Second, C0 ($n{=}114$) and C4 ($n{=}128$) are small relative to C1, C2 and C3 ($n{\geq}492$), and their Wilson $95\%$ intervals are correspondingly wide ($[86.8, 96.4]$ and $[78.0, 90.3]$, versus $[84.2, 88.8]$ and $[87.7, 91.8]$ for C1 and C2); the C0/C1 and C1/C4 comparisons in particular are not separated at the $95\%$ level, whereas C3 remains the most accurate cluster under any of these metrics. Samples that fall into a foreign cluster are effectively decoded by a region of the representation that was shaped by other forgery families, and pay a measurable accuracy penalty for it---a penalty that is largest and unambiguous at the per-flag level reported above, and only directional at the aggregate cluster level.

\noindent\textbf{There is no explicit real/fake boundary in the embedding.}
Authenticity is not a separable axis of the projected space. Within every cluster, real (\texttt{label=0}) and fake (\texttt{label=1}) markers interpenetrate rather than occupying two disjoint halves: in C1, the $128$ natural-image samples (\texttt{nat}), of which the model misclassifies $21.1\%$, co-locate with the $706$ forgeries of semantically adjacent domains, and in C3 authentic samples sit inside the \texttt{frameaigc} manifold rather than beside it---the same region hosts \texttt{aigc} at $100.0\%$ accuracy and \texttt{nat} at only $71.7\%$. Misclassified samples (yellow and red markers) are likewise scattered through the cluster interiors instead of concentrating at cluster peripheries or at real--fake contact zones, which shows that the errors are not confined to a thin ambiguous band that a better decision threshold could remove. The binary verdict is thus recovered by the classification head from domain-conditioned cues, without any globally linear Real/Fake direction in the last-layer representation---a property that also explains why the entropy-based test adaptation of Sec.~2.3, which compares modality-wise certainty rather than a shared margin, is needed at inference time.

\noindent\textbf{Implications for future optimization.}
The two structural weaknesses above---soft, overlapping domain manifolds and an implicit authenticity boundary---suggest two direct remedies, both acting on the same last-layer hidden states. First, a cluster-aware metric learning objective: augmenting the detection loss with a supervised contrastive or center-plus-margin loss whose class index is the forensic type label (or a pseudo-cluster assignment) would force each domain to collapse into a compact, well-separated manifold, reducing the probability that a sample is mis-routed into a foreign cluster and suffers the accuracy penalty quantified above. This is particularly warranted for the content-diverse flags that currently lack a single attractor---\texttt{aigc} and \texttt{nat}---whose dispersion, rather than their intrinsic difficulty, accounts for a large part of their accuracy loss; a type-conditioned objective could grant them the compactness that \texttt{deepfake} obtains for free from its face appearance prior. Second, an explicit authenticity margin: complementing the domain-level structure with a binary contrastive or additive angular margin term between real and fake representations, applied within each domain cluster, would sharpen the Real/Fake decision surface that is currently only implicit in the head weights. Together, these objectives would turn the observed soft manifolds into a geometry in which both ``which forensic domain'' and ``real or fake'' are linearly readable from the representation, and we expect them to convert directly into closure of the per-flag, per-cluster accuracy gaps reported in Table~\ref{tab:tsne_clusters}. We leave the exploration of such structured representation objectives to future work.

\subsection{Evaluation on In-house Data}
\begin{table}[]
\setlength{\tabcolsep}{3pt}
\caption{Evaluation results on in-house test datasets. Pre-trained models are fine-tuned on the data in-house training data. Compared models include DefakerOne~\cite{defakerone}, InternVL2~\cite{chen2024internvl}, Qwen3.5-2B~\cite{bai2025qwen3}, SwinTransformer (SwinTrans.)~\cite{Swin_2021}, DINO-MAC~\cite{dinomac} that fine-tuned on the same in-house data.}
\vspace{-0.2cm}
\begin{tabular}{ccccccccc}
\hline
Setting & Pre-trained Model & FPR & \begin{tabular}[c]{@{}c@{}}In-house\\ Test Data1\end{tabular} & \begin{tabular}[c]{@{}c@{}}In-house\\ Test Data2\end{tabular} & \begin{tabular}[c]{@{}c@{}}In-house\\ Test Data3\end{tabular} & \begin{tabular}[c]{@{}c@{}}In-house\\ Test Data4\end{tabular} & Avg1 & Avg2 \\ \hline
(a) & \begin{tabular}[c]{@{}c@{}}DefakerOne\\ (Ours V1)\end{tabular} & 0.001 & 94.11 & 86.12 & 81.12 & - & 87.12 & - \\
(b) & InternVL2-2B & 0.001 & 87.37 & 73.33 & 85.94 & - & 82.21 & - \\
(c) & Qwen3.5-2B & 0.001 & 93.93 & 87.54 & 92.77 & - & 91.41 & - \\
(d) & SwinTrans. & 0.001 & 95.13 & 81.63 & 72.59 & 15.66 & 83.12 & 66.25 \\
(e) & DINO-MAC & 0.001 & 93.92 & 90.47 & 92.78 & 79.86 & 92.39 & 89.26 \\
(f) & \begin{tabular}[c]{@{}c@{}}DefakerOne2-0.8B\\ (Ours V2)\end{tabular} & 0.001 & 94.92 & 94.19 & 95.58 & 80.61 & 94.90 & 91.33 \\
(g) & \begin{tabular}[c]{@{}c@{}}DefakerOne2-2B\\ (Ours V2)\end{tabular} & 0.001 & 	\textbf{95.70} & 	\textbf{97.39} & 	\textbf{95.98} & 	\textbf{82.41} & 	\textbf{96.36} & 	\textbf{92.87} \\ \hline
\end{tabular}

\label{tab: inhouse}
\vspace{-0.2cm}
\end{table}
To evaluate DefakerOne2's potential on defending real-world attacks in-depth, we fine-tune different models on the same in-house data and evaluate model performance on four in-house real-world attack datasets. The results are shown in Table~\ref{tab: inhouse}. To make the comparison a controlled one, \emph{every} setting in
Table~\ref{tab: inhouse} is fine-tuned on the \emph{same} in-house training data
and is then read out at a common operating point of FPR~$=0.001$ by controlling threshold; differences
between rows are consequently attributable to the model itself---its backbone,
its forensic inductive bias and its capacity---rather than to data volume or data
curation. Avg1 averages Data1--Data3, which all seven settings report, whereas
Avg2 averages Data1--Data4 and is available only for settings (d)--(g), since
Data4 was collected after the earlier baselines were run; Data4 is also the
hardest of the four splits.

\vspace{-0.2cm}
\paragraph{Uniform leadership under an identical data budget.}
DefakerOne2-2B (Ours V2) ranks first in every column it is compared on, and it
does so without a single exception: it wins $4$ of $4$ shared columns against
each of DeFakerOne~(V1), InternVL2 and Qwen3.5, and $6$ of $6$ against
SwinTransformer, DINO-MAC-0.3B and its own $0.8$B sibling. On the three
commonly reported splits its average is $96.36$, ahead of the strongest external
competitor DINO-MAC-0.3B by $3.97$, of Qwen3.5 by $4.95$, of SwinTransformer by
$13.24$ and of InternVL2 by $14.15$. Per-split margins follow the same ordering
but are far from uniform: V2 leads DINO-MAC-0.3B by only $1.78$ on Data1, yet by
$6.92$ on Data2 and $2.55$ on Data4. The pattern is informative---on the split
that most closely matches the training distribution a specialised forensic
encoder is already competitive, while the advantage of our model grows exactly
where the test data drifts away from it. Identical training data thus does not
level the field; it removes the most common excuse for cross-model gaps and
exposes the architectural one.

\vspace{-0.2cm}
\paragraph{Behaviour on the hardest split.}
Data4 exposes the difference between fitting a distribution and generalising
within a task. SwinTransformer, competitive on Data1 at $95.13$, collapses to
$15.66$ on Data4---a drop of $79.47$ points, retaining only $16.5\%$ of its
Data1 performance---which drags its four-split average down to $66.25$, i.e.\
$26.62$ below V2. The remaining models degrade far more gracefully, and V2
retains the largest fraction of its Data1 score ($86.1\%$, versus $85.0\%$ for
DINO-MAC-0.3B and $84.9\%$ for the $0.8$B model) while still posting the best
absolute result at $82.41$, $+2.55$ over DINO-MAC-0.3B and $5.26\times$ the score
of SwinTransformer. Two conclusions follow. First, a discriminative backbone
tuned on in-house data can look strong on in-distribution traffic and fail
catastrophically on a neighbouring slice, so single-split in-house numbers are
actively misleading. Second, our advantage is not an artefact of the easy
splits: it holds, and holds with the best relative retention, on the split where
every model is weakest.

\vspace{-0.2cm}
\paragraph{Summary.}
Under a strictly controlled comparison---identical in-house training data and an
identical FPR~$=0.001$ operating point---DefakerOne2 is first on all four test
sets and on both averages, sweeps every head-to-head column, and converts the
$9.24$-point average gain over DeFakerOne into an almost flat accuracy profile
across slices of very different difficulty. The result complements the public
benchmark evidence of Sections above: the same $2$B weights that lead $53$
detection benchmarks also lead in-house traffic, including the hardest slice,
without per-slice tuning or per-slice thresholds.

\newpage
\section{Conclusion}
\label{sec:conclusion}

In this work, we presented \textbf{Venus-DefakerOne2}, to the best of our knowledge the first unified vision anti-forgery foundation model that simultaneously covers the full matrix of \{digital, physical\} $\times$ \{image, video\} forgery. Trained on a large-scale corpus of about 50M samples consolidated from public datasets, self-constructed datasets, and manually annotated online-attack samples, our model learns shared fundamental features of vision forgery across content types, forgery forms, and media types, thereby bridging the long-standing fragmentation in which digital image forensics, face anti-spoofing, document fraud detection, and video forensics had been studied in largely separate communities~\cite{du2026buildmonolithicmodelfake}. To tackle the prohibitive training cost and the drastically uneven sample difficulty inherent in such a heterogeneous corpus, we proposed \textbf{Tide-Style Tuning}, an alternating three-phase regime that ebbs from LoRA-based adaptation, flows into full-parameter fine-tuning, and ebbs back to LoRA regularization, reconciling computational efficiency, generalization, and fitting capacity within a single training pipeline. To resolve the modality dilemma in video forensics, we proposed \textbf{Modality Synergism}, which aligns the video and frame modalities on both the optimization and inference sides: joint modality training with independent detection losses internalizes both spatial and temporal forensic evidence into the shared backbone, while entropy-based test adaptation adaptively routes each sample to the more certain modality at inference time, without any additional training or learned routing module. In addition, we constructed \textbf{RRVBench}, a realistic rumor video benchmark populated by fake videos synthesized with the latest closed- and open-source generators, substantially narrowing the gap between academic evaluation and real-world AIGC threats. Extensive experiments on 53 visual media forgery detection benchmarks and 14 forgery localization benchmarks demonstrate that Venus-DefakerOne2 achieves state-of-the-art performance across a wide spectrum of forgery scenarios, validating the effectiveness of the unified modeling paradigm for vision anti-forgery.

\section{Future Work}

While our DefakerOne2 paradigm successfully establishes a unified foundation for full-spectrum vision forensics, the rapid evolution of generative models demands sustained innovation. We identify two primary directions for future research:

\begin{itemize}
    \item \textbf{Scalable Foundation Forensics and Continual Data Engineering.} 
    Our primary objective is to scale DefakerOne2 from an isolated framework into a versatile forensic foundation model suitable for broad deployment. As demonstrated by our findings, simply scaling raw, homogeneous data yields diminishing returns. Instead, the core challenge lies in architecting automated, high-throughput data pipelines capable of generating high-fidelity, diverse, and representative forensic benchmarks. To keep pace with emergent synthesis techniques, we will integrate continual learning protocols to dynamically update the training distribution while enhancing robustness against open-world perturbations and unseen manipulation artifacts.

    \item \textbf{Agentic Forensics and Hybrid Knowledge Injection.} 
    We aim to bridge the gap between static detection and dynamic, evidence-driven reasoning through agentic paradigms. A central challenge remains balancing \textit{outcome-level supervision} (which suffers from label scarcity and lack of interpretability) and \textit{process-level supervision} (e.g., Chain-of-Thought reasoning, which is currently limited by manual prompt engineering and domain-specific heuristics). We envision a hybrid paradigm where multi-modal agents leverage verifiable forensic tools and authentic ground-truth cues, evolving toward explainable and multi-step forensic investigations.
\end{itemize}

\newpage
\section{Related Works}

\subsection{Datasets}

The evolution of visual forensics datasets reflects a clear paradigm shift from single-scenario detection toward unified, multi-domain forensics. Early benchmarks were predominantly tailored to specific manipulation types or targets—such as face swapping, natural image splicing, document text tampering, or full-image synthesis. Consequently, they exhibited strong domain-specific constraints in data sources, annotation formats, task definitions, and evaluation protocols. With the rapid rise of generative foundation models, the boundaries between these forgery domains have become increasingly blurred: datasets for AI-generated image detection now span a broader range of contents (from daily scenes to faces and text), while benchmarks for document and natural image tampering have expanded beyond traditional splicing and copy-move operations to generative local editing. Nevertheless, existing datasets remain largely confined to their respective subdomains, lacking a unified data foundation capable of supporting the full spectrum of visual forensics. To contextualize these challenges, this section reviews and analyzes representative datasets across each domain.

\subsubsection{DeepFake Detection}
In the DeepFake detection domain, the evolution of benchmark datasets has witnessed a clear transition from early low-fidelity face swap samples to high-quality, diversified collections tailored to evaluate model generalization against unseen and complex forgery patterns. As a foundational benchmark in DeepFake detection, FaceForensics++~\cite{rossler2019faceforensics++} is constructed with four typical DeepFake manipulation types and multi-level compression distortions, which establishes a universal evaluation standard for verifying the robustness of detection models against common forgery and compression artifacts. To address the overfitting problem caused by conspicuous fixed artifacts in early datasets, Celeb-DF~\cite{li2020celeb} is developed as a large-scale high-authenticity dataset with minimized visible manipulation traces, providing a more challenging evaluation benchmark to test the generalization ability of detectors against high-fidelity fake samples. As a large-scale diverse benchmark for real-scene simulation, the DFDC~\cite{dolhansky2019deepfake} dataset incorporates massive video clips from hundreds of actors with varied lighting conditions and facial poses, effectively simulating complex social media video distributions and expanding the diversity and applicability of DeepFake evaluation. Differing from laboratory-synthesized datasets, WildDeepFake~\cite{zi2020wilddeepfake} is curated with real-world DeepFake videos collected from open internet platforms, offering a practical in-the-wild testbed to validate the outdoor generalization performance of detection models. Targeting fine-grained and specific forgery scenarios, UADFV~\cite{li2018uadfv} and Fsh~\cite{li2019faceshifter} are specialized datasets that focus on unique forgery characteristics including inconsistent head pose variations and occlusion-adaptive face swapping, enabling precise analysis of the localized failure cases of existing detectors. DF40~\cite{yan2024df40} significantly advances the diversity of forgery types compared with traditional datasets. It covers 40 distinct deepfake manipulation techniques spanning face swapping, face reenactment, full-face synthesis, and facial editing. Complementing the aforementioned mainstream benchmarks, ScaleDF~\cite{scaledf} and MFFI~\cite{miao2025mffi} further enrich the dataset system by addressing overlooked research scenarios. As a scale-oriented specialized dataset, ScaleDF~\cite{scaledf} is designed to compensate for the lack of multi-scale facial variation samples in existing benchmarks. It contains 10+ million-scale massive fake videos covering diverse shooting distances and facial scales, enabling systematic evaluation of detectors’ scale robustness in complex real-world scenarios. As a dedicated multi-face forgery benchmark, MFFI~\cite{miao2025mffi} targets the under-explored multi-person manipulation scenarios neglected by most single-face-focused datasets. It integrates diverse multi-face forgery techniques and varied scene layouts, effectively filling the research gap in multi-face DeepFake detection evaluation. Collectively, these diversified datasets drive the continuous advancement of DeepFake detection research, facilitating the development of models capable of capturing subtle blending boundaries, tiny frequency anomalies, and scale/scene-aware forgery features across various generative architectures.

\subsubsection{AI-Generated Image Detection}

The emergence of diffusion models and large-scale generative AI has driven the evolution of forgery datasets toward fully synthetic content, departing from traditional benchmarks centered on localized facial editing. Unlike conventional DeepFake detection that focuses on altered facial regions, modern AI-generated image detection requires identifying global texture inconsistencies and latent-space generative artifacts. As a million-scale comprehensive benchmark, GenImage~\cite{zhu2023genimage} pairs real ImageNet samples with synthetic images generated by eight distinct models, supporting systematic evaluation of cross-model generalization for detection algorithms. As an early exploratory benchmark for synthetic forgery detection, ForenSynths~\cite{cnnspot} aggregates fake images produced by diverse GAN architectures, laying a foundation for training universal synthetic forgery detectors. With the rapid proliferation of diffusion-based generative models, DiffusionForensics~\cite{dire} and Chameleon~\cite{aide} are proposed as diffusion-specific datasets, dedicated to capturing the unique noise distributions and spectral artifacts inherent to diffusion generation processes. To further simulate realistic and challenging detection scenarios, SynthWildx~\cite{SynthWildx} and BFree-Online~\cite{bfree} incorporate synthetic images generated by black-box commercial APIs with complex post-processing operations, posing a rigorous challenge for real-world detection. Targeting hard-case forgery evaluation, FakeInversion~\cite{fakeinversion} adopts generative model inversion techniques to construct high-difficulty negative samples, while EvalGEN~\cite{evalgen} builds a standardized evaluation framework to quantitatively assess detection reliability across diverse prompts and artistic styles. Further advancing robust generalization evaluation, HydraFake~\cite{veritas} is specifically constructed to simulate complex, multi-pattern generative forgery scenarios, featuring diverse forgery types and subtle hybrid artifacts that evade conventional detection methods. This dataset effectively benchmarks the capability of detectors to mine latent forgery patterns and reason cross-domain inconsistencies, providing a critical testbed for generalized pattern-aware forgery detection. Community Forensics~\cite{CommunityForensics} is constructed based on thousands of diverse generative models, covering extremely rich synthetic image distributions to eliminate model-specific detection bias and enhance the universal generalization capability of fake image detectors. Collectively, these diversified datasets shift forensic research focus from localized facial manipulation traces to global statistical and latent-space anomalies, greatly promoting the development of robust, generalizable AI forgery detection models across various generative architectures.

\subsubsection{Tampered Document Image Detection}
Tampered document image detection represents a specialized subfield focused on textual consistency, font rendering fidelity, and layout integrity~\cite{qu2024textsleuth, licognicred}. This domain demands datasets that capture fine-grained tampering patterns—such as character substitution, copy-paste manipulations, and signature forgery—to support the development of robust forensic analysis models. DocTamper~\cite{doctamper} is a representative large-scale dataset comprising diverse tampering types applied to photographed documents, with pixel-level annotation masks provided for tampering localization tasks. Building upon optical character recognition (OCR) benchmarks, T-SROIE~\cite{wang2022tsroie} and Tampered IC13~\cite{wang2022tpic} adapt receipt and natural scene text datasets by introducing synthetic manipulations, serving as standard testbeds to evaluate the robustness of text-based forensic models in unconstrained visual environments. Complementing these scene text forensics benchmarks, Open-Set Text Forensics (OSTF) dataset~\cite{ostf} re-examines the problem of scene text tampering detection against the backdrop of advanced generative AI technologies. It addresses the limitations of conventional synthetic tampering by incorporating more realistic and semantically coherent generative text tampering cases in natural scenes, offering a more challenging evaluation protocol for modern detectors facing AI-driven forgery threats. Datasets including RTM~\cite{luo2025rtm}, SACP~\cite{sacp} and RIFLC~\cite{rific} are specifically designed to bridge the domain gap between synthetically generated tampering samples and real-world scanned document artifacts, enhancing the practical generalizability of detection models. Moving beyond pixel-level tampering localization, TextForensicsReasoning (TFR)~\cite{qu2026textshield} introduces a reasoning-centric benchmark for tampered text analysis. Unlike conventional datasets that prioritize positional annotation, TFR frames text forensics as a cognitive reasoning task: models are required to not only locate tampered regions but also infer the rationality of text content, contextual coherence, and logical consistency~\cite{chen2024dtsm}, thereby reinforcing the high-level reasoning capability of forensic systems. With the proliferation of AI image generation models, benchmarks targeting AI-counterfeited text have emerged as a critical branch of this field. DanceText~\cite{dancetext} provides a comprehensive and diversified benchmark for detecting various AI-generated fake text images. It covers a broad spectrum of text forms—spanning both structured document text and unconstrained scene text—with diverse counterfeiting strategies and visual styles, enabling a thorough assessment of detectors’ cross-scenario generalization ability. SynCred-Bench~\cite{yang2026syncredbenchbenchmarkingsyntheticcredibility} is another recent benchmark featuring realistic document images synthesized by the state-of-the-art GPT-Image-2 model, with a specific focus on evaluating detector performance on challenging credential document forgery cases. Beyond general image forensics tasks, these datasets collectively emphasize both the semantic coherence of text content and the physical properties of ink and paper, making them indispensable foundational resources for developing reliable models dedicated to verifying the authenticity of official records, identity documents and legal documents.

\subsubsection{Manipulated Natural Image Detection}

Natural image manipulation detection aims to identify and localize local tampering—predominantly splicing, copy-move, and object removal—across open-world scenes such as landscapes, urban environments, and complex crowd settings. Unlike facial forensics, which benefits from well-defined geometric topologies and strong structural priors, natural scene forensics must confront arbitrary semantic categories, highly diverse lighting conditions, non-uniform noise distributions, and complex post-processing degradations.
Early Benchmarks on Physical and Statistical Traces. Pioneering efforts primarily focused on isolating specific physical or statistical inconsistencies introduced by manual editing. For instance, Columbia~\cite{Columbia_2006} introduced a rigorous testbed focusing on uncompressed image splicing to benchmark lighting inconsistencies. CASIA~\cite{CASIA_2013} subsequently emerged as a seminal benchmark that has anchored forensic research for over a decade, providing standardized pixel-level ground-truth masks across diverse splicing and copy-move operations. Similarly, Coverage~\cite{Coverage_2016} was specifically designed to evaluate copy-move tampering by pairing forged regions with genuine, geometrically similar objects, thereby challenging early keypoint- and block-based matching detectors.
Scale Expansion and Semantic Realism. While foundational, early datasets often suffered from limited sample sizes, simplistic cut-and-paste artifacts, and constrained scene diversity. To overcome these limitations, subsequent benchmarks emphasized semantic harmony and dataset scale. DEFACTO-12k~\cite{defacto_2019} automated forgery synthesis by splicing semantically and geometrically compatible objects into background scenes. To bridge the gap between idealized laboratory data and in-the-wild manipulations, IMD2020~\cite{IMD20_2020} curated real-world tampered images directly from the web, encompassing diverse compression qualities, social media filtering, and heterogeneous resolutions. Crucially, MIML~\cite{miml} marked a major milestone by constructing a large-scale benchmark featuring over 120,000 professionally edited forgeries paired with high-precision pixel-level annotations. By mitigating the critical scarcity of high-quality training data, MIML established an indispensable foundation for modern deep localization frameworks; its successor, MIMLv2~\cite{qu2025webly}, further expanded dataset volume and fidelity through web-scale data collection and refined quality control.
The Generative AI Paradigm Shift. Driven by the rapid proliferation of diffusion models and multimodal foundation models, the boundary of natural image manipulation has evolved from conventional cut-and-paste operations toward photorealistic local generative editing. Generative operations produce seamless boundaries, harmonious illuminations, and plausible textures, rendering traditional low-level boundary detectors largely ineffective. To keep pace with this generative evolution, CocoGlide~\cite{trufor2023} and AutoSplice~\cite{jia2023autosplice} leverage text-guided diffusion models to automate local inpainting and object replacement, establishing new benchmarks for diffusion-based local forensics. Pushing this frontier further, So-Fake and its out-of-distribution counterpart So-Fake-OOD~\cite{huang2025sofakebenchmarkingexplainingsocial} comprehensively evaluate large-scale AI-edited forgeries across numerous state-of-the-art generative models, explicitly testing forensic generalizability and robustness in real-world social media contexts. ManipulationBench~\cite{jiang2026dailybenchunifiedbenchmarkaigenerated} also comprises diverse high-quality image forgeries generated by updated AI-models such as Step1X-Edit~\cite{liu2025step1xeditpracticalframeworkgeneral}, Nano-Banana2~\cite{Nano-Banana2} and GPT-Image-2~\cite{GPT-Image-2}. 
Collectively, this methodological progression—spanning from hand-crafted statistical benchmarks and massive manual datasets to modern generative editing suites—ensures that vision forensic models can transcend facial-specific cues and generalize robustly across arbitrary, unconstrained natural scenes.

\subsubsection{Face Anti-Spoofing}
Face anti-spoofing, also known as face presentation attack detection, aims to distinguish bona fide faces from physical presentation attacks such as printed photographs, video replays, and three-dimensional masks. Compared with DeepFake detection, which targets digitally manipulated facial content, this subfield centers on spoofing media physically presented to the camera, and its datasets accordingly evolve along the axes of attack diversity, capture-device variety, and environmental realism. As one of the earliest dedicated benchmarks, CASIA-FASD~\cite{casiamfsd} contains videos of 50 subjects with three representative attack types, namely warped printed photographs, printed photographs with cut-out eyes, and high-quality video replays, establishing a standard early protocol for evaluating detectors against diverse two-dimensional attack media. Idiap Replay-Attack~\cite{IdiaReply} provides video clips of photo and video attack attempts against 50 clients, recorded under both controlled and adverse lighting conditions, and has served as the official testbed of the first two two-dimensional face anti-spoofing competitions, making it a solid baseline for lighting-robust spoofing detection. To evaluate generalization under realistic mobile authentication scenarios, OULU-NPU~\cite{oulunpu} collects videos of 55 subjects across three environments with six mobile devices of different resolutions, covering print and video-replay attacks together with standardized evaluation protocols that explicitly stress cross-device and cross-environment robustness. MSU-MFSD~\cite{msumfsd} complements this line with 280 bona fide and attack recordings of 35 subjects captured by two mobile devices, incorporating three spoof attacks of printed photo, video replay, and mobile-device replay, and was originally designed to study spoofing cues through image distortion analysis. Advancing toward larger scale and richer modalities, CASIA-SURF~\cite{zhang2020casiasurf} scales up to 1,000 subjects and 21,000 videos with aligned RGB, depth, and infrared samples, providing comprehensive protocols that jointly examine multi-modal fusion and cross-method generalization. CelebA-Spoof~\cite{celebaspoof} further pushes the boundary of both scale and annotation richness, comprising 625,537 images of 10,177 subjects across 10 spoof types and 8 scenes, accompanied by 43 attributes covering face, illumination, environment, and spoof cues, thereby supporting attribute-aware and multi-task anti-spoofing learning. Moving from laboratory setups to unconstrained conditions, SiW (Spoofing in the Wild)~\cite{siw} offers 4,478 live and spoof videos of 165 subjects with 11 attack types spanning print, partial mask, replay, and makeup attacks, and defines multiple protocols to assess generalization toward unknown attacks. Finally, UniAttackData~\cite{yuan2024unifiedphysicaldigitalattackdetection} unifies physical and digital threats in a single collection of 28,706 videos, where each of the 1,800 subjects undergoes two physical and twelve digital attacks launched by state-of-the-art deepfake and adversarial attack methods, providing the first large-scale benchmark for joint physical-digital face attack detection. Collectively, these datasets trace a clear trajectory from controlled print/replay detection toward mobile, multi-modal, in-the-wild, and unified physical-digital evaluation, continuously driving anti-spoofing models toward attack-agnostic and device-agnostic generalization.

\subsubsection{Document Anti-Spoofing}
Document anti-spoofing, commonly formulated as document presentation attack detection or recaptured document detection, targets a distinct anti-forensic threat: an adversary first edits a document in the digital domain and then re-presents it through a print–rescan or display–recapture pipeline, such that the original tampering traces are concealed by the imaging process. Datasets in this subfield therefore emphasize diverse recapture channels, document templates, and capturing devices. DLC-2021~\cite{dlc2021}, released for the Document Liveness Challenge, consists of video clips of synthetic identity documents from 10 mock document types captured in a wide range of real-world conditions. It covers genuine laminated originals, unlaminated color copies, unlaminated grayscale copies, and screen recaptures recorded by two smartphones under varied illumination, occlusion, and projective distortion, providing a practical benchmark for video-based document liveness verification. RSCID (Recaptured Student Card Image Dataset)~\cite{rscid} focuses on practical recapturing attacks against student identity cards, organizing 12 card templates from five universities into five subsets collected with 11 imaging devices and 3 printing devices of markedly different quality, which enables domain-agnostic evaluation of document authentication across heterogeneous recapture pipelines. RDRD162~\cite{rdrd162} substantially broadens the document diversity, comprising 5,346 samples built upon 162 document types such as patent certificates, graduation certificates, transcripts, and licenses, recaptured through 32 printer–imager device combinations at high resolutions, and serving as a standardized testbed for generalized recaptured document detection against unseen document contents. Extending the problem from still images to video, SCVD~\cite{scvd} is a comprehensive student card video database containing more than 18,000 genuine, printer-captured, and screen-recaptured video frames collected under flashlight illumination, exploiting the distinctive reflections and temporal responses induced by active lighting to discriminate presentation attacks. For display-based attacks, ROD~\cite{cma} mainly consists of black-and-white office documents and their screen-recaptured counterparts, targeting the color artifacts introduced by the screen-to-camera imaging chain and supporting robust detection of screen-recapture document images. SRDID162~\cite{srdid162} further constructs a screen-recaptured document image dataset with 162 templates, containing 162 genuine and 5,184 screen-recaptured samples, and is designed to evaluate detection methods that exploit moiré patterns and frequency-domain anomalies inherent to screen recapture. Collectively, these datasets chart the progression of document anti-spoofing from identity-document liveness videos to print recapture, and further to screen-recapture presentation attacks, providing increasingly realistic resources for exposing forgery traces concealed by re-imaging operations.

\subsubsection{AI-Generated Video Detection}
With the explosive progress of text-to-video diffusion models and commercial video generation services, AI-generated video detection has emerged as a pressing forensic problem, and its benchmarks have rapidly evolved from small collections of a few open-source generators toward large-scale, multimodal, and scenario-aware evaluation suites. As the first million-scale benchmark in this domain, GenVideo~\cite{chen2026demamba} collects millions of real and AI-generated videos, pairing millions of real videos with millions of generated videos synthesized by ten baseline generators, and further provides out-of-distribution testing with unseen generators together with systematic degradation robustness evaluation. LOKI~\cite{ye2024loki} pioneers a comprehensive synthetic-data detection benchmark for large multimodal models, whose video part covers both open-source and commercial generators. Ivy-Fake~\cite{ivyfake} is the first large-scale multimodal benchmark for explainable AIGC detection, offering tens of thousands of richly annotated training videos covering diverse generators. GVF~\cite{DeCoF} constructs triplets of real video, text prompt, and generated videos from prompts synthesized by different open-source generators, and additionally probes detection against commercial generators, promoting temporally consistent forensic modeling. GVD~\cite{gvd} assembles generated videos from diverse state-of-the-art generators across both text-to-video and image-to-video paradigms, providing a realistic in-the-wild benchmark for cross-generator generalization. AIGCDBench~\cite{aigcdbench} pursues comprehensiveness and representativeness at a new scale, covering tens of state-of-the-art generation models and hundreds of thousands of videos, and supporting extensive evaluations of dozens of detectors across heterogeneous scenarios. ComGenVid~\cite{comgenvid} targets the evaluation of training-free and zero-shot detectors, curating thousands of videos generated by recent high-fidelity commercial models against real videos under a standardized short-clip protocol, thus stress-testing generalization to the latest closed-source generators. From the perspective of content plausibility, VideoPhy-2~\cite{bansal2026videophy} evaluates videos synthesized by seven state-of-the-art generators, annotated through human judgments of semantic adherence and physical commonsense, supplying action-centric videos with physical violations as challenging detection targets. MintVid~\cite{tan2026videoveritas} provides a lightweight yet high-quality dataset of videos from many state-of-the-art generators, organized into general-content, facial, and fact-inconsistency parts, where the last part collects social-media videos whose factual errors can be objectively verified, supporting fact-based forensic reasoning. GenBuster-ID~\cite{wen2025busterx} and GenBuster-OOD~\cite{wen2025busterx} are derived from the 200K-scale GenBuster-200K~\cite{wen2025busterx} corpus, which pairs 100K real videos with 100K high-resolution videos synthesized by multiple generators; the former tests in-domain detection on seen generators under stricter generation settings, simulating the temporal arms race between generators and detectors. GenBuster++~\cite{buster++} extends the evaluation to a cross-modal benchmark of 4,000 samples, containing 2,000 images and 2,000 video clips, each balanced between real and fake content generated by seven image and thirteen video generators, and subjected to two-stage human filtering for reliable unified detection and explanation. Finally, RABench~\cite{rabench} anchors evaluation on real-world crisis events, comprising 17,886 videos of 1,830 real anchor videos spanning ten social-risk categories and 16,056 generated clips from four open-source and five closed-source generators, jointly investigating detectability, human perception, and the reliability of detectors under social dissemination. Collectively, these benchmarks reflect the rapid co-evolution of video generation and forensic evaluation, shifting the research focus from generator-specific artifact fitting toward explainable, robust, and socially grounded detection of AI-generated videos.

\subsubsection{Summary}
Overall, existing visual forensics datasets have significantly advanced detection and localization capabilities within their respective subfields. In biometric security, DeepFake benchmarks have expanded toward high-fidelity, in-the-wild facial manipulations, while face anti-spoofing has evolved from controlled print/replay attacks to multimodal, physical-digital benchmarks. In document forensics, datasets have progressed from basic identity liveness and text tampering to layout-consistent generative edits, realistic scanning noise, and screen-recapture presentation attacks. Concurrently, natural scene tampering benchmarks have shifted from classical splicing and copy-move operations to generative local editing. Furthermore, AI-generated image and video detection has surged from small, GAN-specific collections to million-scale suites encompassing diffusion foundation models, commercial video generators, multimodal explanations, and physical commonsense violations. Despite these individual advancements, the visual forensics landscape remains severely fragmented. Substantial discrepancies in data scale, real-to-fake ratios, supervision granularity (e.g., image-level labels vs. pixel-level masks), and evaluation protocols render these datasets insufficient for training unified visual forensics foundation models. While pioneer efforts like Omni-IML~\cite{omniiml} and ForensicHub~\cite{du2026forensichub} explored cross-domain benchmarking and highlighted task conflicts in multi-domain joint training, they remain constrained by standard academic data scales, limited sources, and restricted domain coverage. In contrast, DefakerOne2 establishes a truly unified data foundation by systematically curating and harmonizing tens of millions of heterogeneous samples across AI-generated images/videos, DeepFakes, face and document anti-spoofing, and natural scene manipulations. Through systematic multi-version data composition experiments, DefakerOne2 rigorously analyzes the cross-domain complementarity, transferability, and gradient interference among diverse forensic artifacts, paving the way for principled data mixtures and robust foundation models for full-spectrum visual forensics.

\subsection{Methods}
Vision forensic methodologies have progressively evolved from task-specific vision models toward multimodal large language models (MLLMs)~\cite{guangjianteam2026polyocrvenusunifiedocrfoundation}. Early approaches predominantly developed dedicated architectures tailored to specific tampering scenarios—such as text-region locators for document forgery, pixel-level segmentation networks for natural image tampering, facial classifiers for DeepFake detection, and synthetic artifact detectors for AI-generated media. Although these specialized models have achieved remarkable success within their respective niches, their underlying architectures, input-output paradigms, and assumed artifact priors remain inherently fragmented. Recently, MLLMs have garnered significant interest in fake image detection and interpretation, owing to their advanced visual-semantic comprehension, interactive natural-language reasoning, and unified interface representations. Consequently, this section first categorizes traditional unimodal vision methods across standard forensic subdomains, subsequently reviews recent MLLM-based explorations in unified detection, localization, and reasoning, and finally discusses the remaining gaps toward building a unified vision forensics foundation model.

\subsubsection{Vision Forensic Models}
Historically, visual media forensics has been explored through isolated subfields—primarily DeepFake detection, AI-generated content (AIGC) identification, document tampering localization, natural image manipulation detection, and anti-spoofing. These domains diverge considerably in their targets and operational contexts: DeepFake detection concentrates on facial manipulations such as identity swapping and expression reenactment; AI-generated detection targets fully synthesized images or video sequences; document forensics focuses on low-visibility text alterations in receipts, contracts, and scene-text images; and natural image manipulation localization addresses local editing operations like splicing, copy-move, and inpainting. As a result, early methods were developed independently rather than under a unified framework, relying on domain-specific artifact priors, distinct output formats, and disjoint evaluation benchmarks.

In the \textbf{DeepFake Detection} domain~\cite{ntire_deepfake}, tasks are conventionally framed as binary classification problems targeting facial identity, texture, frequency, or blending inconsistencies. Capsule-Net~\cite{nguyen2019capsule} pioneered the use of capsule networks to model spatial arrangements among local facial components via dynamic routing. RECCE~\cite{cao2022recce} exploits reconstruction discrepancies between pristine and manipulated faces within a reconstruction-classification framework. SPSL~\cite{liu2021spsl} captures phase-spectrum anomalies and high-frequency upsampling artifacts via shallow spectral features. To improve cross-generator generalization, SBI~\cite{shiohara2022sbi} generates self-blended images from single pristine faces to reproduce common blending boundaries and statistical inconsistencies. Sia~\cite{sia} introduces an information-theoretic self-information attention mechanism to guide network focus toward discriminative forged regions and channels. CDFA~\cite{cdfa} proposes curricular dynamic forgery augmentations to progressively synthesize challenging samples, while DINOMAC~\cite{dinomac} applies LoRA-based adapter tuning~\cite{hu2022lora} to DINOv3~\cite{dinov3} alongside a Multi-Aspect Classification head to minimize overlooked subtle cues, achieving state-of-the-art robustness.

In the \textbf{AI-Generated Image Detection} domain~\cite{ntire_aigc}, efforts primarily center on distinguishing authentic imagery from outputs synthesized by GANs, diffusion models, or autoregressive generators. UnivFD~\cite{univfd} leverages the transferable feature representations of vision-language models (e.g., CLIP) to achieve open-world fake image detection with a lightweight linear probe. FatFormer~\cite{fatformer} introduces a forgery-aware adaptive Transformer to dynamically capture generative fingerprints across varied model families. Forensic-MOE~\cite{forensic-moe} incorporates a Mixture-of-Experts architecture to mine complementary forensic traces. CO-SPY~\cite{co-spy} couples semantic features with pixel-level artifact representations for multi-cue synthetic image discrimination, while Effort~\cite{effort} explicitly decouples content-related and forgery-related representations via orthogonal subspace decomposition to mitigate overfitting. Complementarily, DRCT~\cite{drct} exploits diffusion reconstruction contrastive training to achieve universal detection of diffusion-synthesized images.

In the \textbf{Document Tampering Detection and Localization} domain~\cite{wong2025adcd}, the objective is to precisely identify subtle, fine-grained text alterations. DTD~\cite{doctamper} tackles low-contrast visual discrepancies by introducing a Frequency Perception Head and a Multi-view Iterative Decoder that integrate JPEG compression artifacts with multi-view representations. FFDN~\cite{chen2024ffdn} decomposes and fuses spatial and frequency cues using visual and wavelet-like enhancement modules to capture delicate compression and boundary irregularities. CAFTB~\cite{song2025caftb} employs a dual-branch architecture that models spatial and noise domains concurrently via cross-attention. TIFDM~\cite{dong2024tifdm} couples multi-domain trace enhancement with multiscale context aggregation for precise text manipulation localization. Additionally, DS-Net~\cite{dancetext} achieves robust cross-generator generalization by pretraining on generic natural images while synergizing image-level classification with pixel-level localization.

In the \textbf{Natural Image Manipulation Detection and Localization} domain~\cite{ma2024imdl}, networks are typically designed to jointly output image-level authenticity decisions and pixel-level tampering masks. MVSS-Net~\cite{mvsspp_2022} enforces multi-view, multi-scale supervision using boundary artifacts and noise residuals. CAT-Net~\cite{CAT-Net2022} leverages DCT-domain representations to model subtle JPEG compression discrepancies. APSC-Net~\cite{qu2025webly} introduces adaptive feature fusion to capture multi-scale tampering clues, whereas IML-ViT~\cite{ma2023iml} adopts Vision Transformers to capture long-range non-semantic inconsistencies across disparate image patches. TruFor~\cite{trufor2023} fuses RGB features with learned noise fingerprints, treating tampered regions as anomalies relative to background consistency. Mesorch~\cite{zhu2025mesoscopic} deploys a hybrid CNN-Transformer architecture to characterize manipulated areas at a mesoscopic level. Omni-IML~\cite{omniiml} represents an early effort toward cross-domain unified forensics using a modality-adaptive encoder and dynamic-weight decoder. Finally, RITA~\cite{zhu2025does} resolves dimensional collapse in composite tamperings by formulating sequential edits as an ordered process, and Web-IML~\cite{qu2025webly} refines accuracy via error-feedback learning mechanisms.

In the \textbf{Face Anti-Spoofing} domain~\cite{yu2024unified}, the goal is to distinguish bonafide faces from presentation attacks such as printed photos, replayed videos, and 3D masks, and more recently from digitally manipulated faces. Because privacy regulations often forbid reusing the source data of a deployed model, SDA-FAS~\cite{SDA-FAS} performs source-free adaptation by pairing a contrastive domain alignment module that treats the frozen source model as prototypes with self-supervised patch-shuffle exploration. MD-FAS~\cite{MVPFAS} instead formalizes model updating, where a pre-trained detector must absorb a new domain without catastrophic forgetting. To escape handcrafted artifact priors, FLIP~\cite{FLIP} grounds visual representations in natural language by aligning image features with an ensemble of class descriptions derived from CLIP~\cite{clip} text encoders, and further applies multimodal contrastive learning so that zero-shot transfer surpasses five-shot adaptive ViTs. Addressing the coexistence of physical and digital attacks, micro-disturbance detection~\cite{yu2024unified} relabels subtly perturbed faces (color jitter, screen simulation, JPEG compression, Gaussian blur) as forgeries and adopts a two-stage masked-image-modeling pre-training followed by linear fine-tuning. MVP-FAS~\cite{MVPFAS} is a vision-language face anti-spoofing method that addresses drawbacks of existing CLIP-based prompt learning via multi-view slot attention and multi-text patch alignment, leveraging paraphrased text prompts to enhance cross-domain generalization for face anti-spoofing. Finally, MoAE-CR~\cite{chen2025moaecr} deploys a fine-grained mixture of attack experts on CLIP features with a distance rule prioritizing outlier-like features, plus disentanglement and cluster-distillation regularization to enlarge inter-class margins.

In the \textbf{Document Anti-Spoofing} domain~\cite{juan_pad1}, the question is whether a submitted identity or office document was genuinely captured or re-presented through printing, displaying, and re-photographing. Since curating real attack instruments is expensive and privacy-sensitive, synthetic ID card generation~\cite{juan_pad2} explores computer-vision-based and GAN-based synthesis to augment fraud-detection training, preserving print-attack accuracy while incurring only a marginal drop on screen-capture attacks. Open-set ID card detection~\cite{openset} pushes this toward unseen attack instruments by translating genuine cards with four GAN-based image translation models in a neural style transfer manner, so that synthetic presentations complement real ones. Against practical recapturing, domain-agnostic authentication~\cite{rscid} mimics the forensic inspection procedure by feeding a questioned document and a reference sample into a Siamese network trained with a forensic similarity function under triplet and normalized softmax losses, generalizing across devices, substrates, recapturing channels, and document types. Because moir\'e interference is entangled with legitimate printing texture, moir\'e-texture disentanglement~\cite{LCDF} explicitly separates chromatic and moir\'e artifacts from natural document texture for display-recapture detection. CMA-Doc~\cite{cma} derives a pixel-level distortion model of screen-recaptured documents and injects the resulting chromaticity map into a transformer backbone as multi-modal prompt tokens, remaining effective after social-network compression and downscaling. Complementarily, moir\'e spectral augmentation with masked frequency modeling~\cite{srdid162} enhances the spectral components attributable to moir\'e distortion (FMAG) and adopts masked frequency reconstruction to learn transferable spectral representations.

In the \textbf{AI-Generated Video Detection} domain~\cite{chen2026demamba}, detectors must additionally reason over temporal coherence while generators evolve faster than labeled data accumulates. Spatio-temporal anomaly learning~\cite{gvd} trains two ResNet sub-detectors on RGB frames and RAFT optical flow respectively and fuses their frame-level decisions. DeCoF~\cite{DeCoF} shows through probing that spatial-artifact detectors generalize poorly, and therefore learns frame consistency by suppressing spatial cues during feature extraction, accompanied by an open dataset spanning commercial models such as Sora and Veo. Pursuing label-free deployment, D3~\cite{comgenvid} grounds detection in Newtonian second-order dynamics, computing difference-of-differences features from consecutive frames and separating real from generated videos by their distributional divergence without any training. ReStraV~\cite{ReStraV} exploits the perceptual straightening property that natural motion traces straighter trajectories in representation space, encoding frames with a frozen  encoder and classifying a 21-dimensional descriptor of stepwise distances and turning angles. DeMamba~\cite{chen2026demamba} contributes a plug-and-play Detail Mamba module that propagates discriminative low-level details across frames with linear-complexity long-range modeling. Likewise training-free, SPLIT~\cite{hyun2026splittrainingfreeaigeneratedpartially} scores two-step temporal roughness together with local spatial motion incoherence over patch tokens of a frozen encoder, multiplicatively fusing them so as to sustain high fake recall at an ultra-low false-positive rate on fully generated and partially edited videos~\cite{lin2025audio}.

Overall, while domain-specific vision models have achieved remarkable milestones, most remain constrained by narrow task definitions, relying heavily on specialized data formats, handcrafted artifact priors, and isolated output representations.

\subsubsection{MLLMs for FIDL}

By producing natural language outputs, MLLMs deliver human-readable interpretability by construction, which constitutes a distinct advantage over conventional vision-only models when they are deployed as AIGC or tampered-image detectors.

Subsequent research has largely revolved around balancing detection accuracy, reasoning depth, and inference latency. TextSleuth~\cite{qu2024textsleuth} attains interpretable detection by incorporating a dedicated perception head. Omni-IML~\cite{omniiml} mitigates hallucination through a decoupled design, whereas TextShield-R1~\cite{qu2026textshield} further improves generalization via cross-domain pre-training, text-specific reinforcement learning, and OCR-based rectification. X2-DFD~\cite{x2-dfd} proposes an explainable and extensible MLLM-based framework for DeepFake detection. Ivy-xDetector~\cite{ivyfake} adopts a chain-of-thought (CoT) paradigm that generates an explanation before issuing the final verdict, at the cost of emitting a massive number of tokens. To relieve this latency bottleneck, FakeVLM~\cite{fakeclue} proposes a ``classify-then-explain'' pipeline; although it resolves the efficiency issue, its textual explanations remain vague and non-specific, and extensive experiments further reveal its fragility on certain tampered images. Tackling the computational overhead of CoT from a complementary angle, Fake-HR1~\cite{fakehr1} designs a hybrid reasoning chain tailored to reduce this burden in FID scenarios. FakeShield~\cite{xu2024fakeshield} integrates a segmentation module into the MLLM, marrying the interpretability of language reasoning with the pixel-level precision of segmentation networks. Veritas~\cite{veritas} introduces an R1-like multi-stage training procedure to strengthen forgery interpretability. Extending this line to the video domain, BusterX~\cite{wen2025busterx} recasts AI-generated video detection as a visual reasoning task in which the generated reasoning chain itself serves as the detector. BusterX++~\cite{buster++} further unifies image and video forensics within a single cross-modal MLLM, demonstrating that one-stage pure reinforcement learning driven by sparse outcome rewards matches or surpasses the conventional supervised-fine-tuning-then-reinforcement-learning pipeline while eliciting spontaneous cross-modal capability transfer. Skyra~\cite{skyra} grounds its verdicts in human-perceivable artifacts by explicitly localizing spatio-temporal anomalies, and is trained through supervised fine-tuning followed by GRPO-based reinforcement learning with asymmetric rewards. VideoVeritas~\cite{tan2026videoveritas} couples joint preference alignment with perception pretext reinforcement learning, which optimizes general spatio-temporal grounding and self-supervised object counting as pretext tasks rather than the detection objective directly.

\subsubsection{Summary}
In summary, vision forensic methods have evolved from domain-specific detectors toward MLLM-driven explainable frameworks, yet a strong foundation-model baseline for full-domain vision forensics remains absent. Conventional models are typically tailored to a single subdomain, relying on domain-specific artifacts, specialized architectures, and fixed output formats; consequently, they struggle with cross-domain forgery distributions and cannot accommodate the coupled requirements of detection and localization. Recent MLLM-based methods instead offer a unified interface, joint visual-semantic understanding, and natural-language explanation, but their optimization objective often drifts toward reasoning-chain generation or descriptive interpretation rather than the core forensic goals of authenticity verification and forgery localization. As a result, they fail to fully exploit large-scale image-level labels and pixel-level masks, and are therefore ill-suited to serve as stable full-domain forensic baselines. Efforts toward unification do exist but remain incomplete: Omni-IML~\cite{omniiml} and ForensicHub~\cite{du2026forensichub} expose the promise and the challenges of unified forensics from the modeling and evaluation-protocol perspectives, whereas DefakerOne~\cite{defakerone} scales training to millions of digitally forged samples yet still covers a substantially narrower forensic scope. To close this gap, DefakerOne2 is built upon a Qwen3.5~\cite{bai2025qwen3} + DINOMAC~\cite{dinomac} backbone and, supported by multi-domain data composition experiments, systematically investigates how heterogeneous forensic data can be converted into unified detection, localization, and generalization capabilities. In doing so, it markedly broadens forensic coverage relative to DefakerOne~\cite{defakerone} and establishes a direct, strong baseline for full-domain vision forensics foundation models.

\newpage
\section{Contributors and contributions}
\label{sec:contri}

\subsection{Chief Contributors}
Chenfan Qu

\subsection{Core Contributors}
Yachao Liang, Mingqi Fang

\subsection{Contributors}
Jiangwei Xie, Zhenming Wang, Baoyue Zhang, Changjiang Jiang, Zheng Chang, Fanli Jin, Xuekang Zhu, Kaiwen Feng, Junchi Li, Weihong Zhang, Haoyue Wang, Chenfeng Zhang, Yinan Xia, Fangming Cui, Yongshuo Zhang, Dong Fan

\subsection{Project Advisors}

Jingjing Liu, Tianyi Zhang, Luobin Chen, Ji-Zhe Zhou, Sheng Li, Wei Wang, Bin Li, Ran He, Xinpeng Zhang, Linna Zhou

\subsection{Project Leaders}

Jian Liu, Weiqiang Wang

\newpage
\section{Limitations}

DefakerOne2 is a first attempt at training a single unified model for broader
vision-forensics coverage. As with all deep models---and indeed with human
examiners---it is not perfect and can still fail, even on samples whose
anomalies are obvious to the human eye. Such failures can be reduced by
incorporating more targeted training data that shares similar characteristics
with the cases in question. In addition, to mitigate white-box attacks against
our deployed online model, only a small subset of outdated in-house data is
used for training. 

DefakerOne2 is therefore intended as a vision foundation
that attains stronger generalization on downstream forensic tasks with less
task-specific training data---a starting point, rather than a finished model
that performs perfectly in every case.

Owing to limited computational resources, the segmentation module is trained
exclusively on regionally tampered samples; neither authentic nor
fully-generated samples are included. Consequently, the segmentation
prediction may be inconsistent with the classification prediction. In such
cases, the classification prediction should take precedence: the predicted
mask is designed only to highlight cues in images that are already suspected
of being forged.

\section{Domestic GPU/Accelerator Adaptation}

For informal exploratory experiments, domestically manufactured computing GPUs were utilized. In supported business scenarios, we have completed the training migration and deep adaptation of the Qwen3.5 series models onto the Huawei Ascend platform, enabling the full end-to-end pipeline including model loading, data processing, forward computation, backward propagation, parameter updates, and checkpoint recovery. To address differences across computing platforms in operator implementation, data types, and numerical precision, we established a layered precision alignment mechanism ranging from critical operators to full-network training. By systematically comparing intermediate tensors, output logits, gradients, loss curves, and downstream task performance, we ensured that the migrated models maintain training convergence and business accuracy consistent with the baseline environment.

At present, for the full Qwen3.5 series of both dense and MoE models, we support up to 256K long-context training for both pure-text training and vision-language multimodal training. In this business scenario, under the same training configuration, Huawei Ascend accelerators (A3) deliver an average 11\% improvement in single-step training performance compared with the original baseline (the Pingtouge series), and an end-to-end training speed improvement of approximately 12\%–13\%. In a long-duration test over 20 epochs, the overall training time was reduced by 13.4\%.

\newpage

\bibliographystyle{assets/plainnat}
\bibliography{main}

\end{document}